\documentclass[letterpaper,11pt]{scrartcl}

\usepackage[T1]{fontenc}
\usepackage[utf8]{inputenc}
\usepackage{lmodern}
\usepackage{amsfonts,amsmath,amssymb,amsthm,mathtools}
\usepackage[letterpaper,margin=1.00in]{geometry}
\usepackage[english]{babel}
\usepackage[dvipsnames]{xcolor}
\usepackage[round,sort]{natbib}
\usepackage{bm}
\usepackage{graphicx}
\usepackage{booktabs}
\usepackage{tabularx}
\usepackage{colortbl}
\usepackage{enumitem}
\usepackage[normalem]{ulem}
\usepackage[most]{tcolorbox}
\usepackage{environ}
\usepackage[ruled,vlined,linesnumbered]{algorithm2e}
\SetAlgoSkip{smallskip}
\SetNlSty{textrm}{}{}
\usepackage{tikz}
\usepackage{wrapfig}
\usepackage{placeins}
\usepackage{needspace}
\usetikzlibrary{arrows.meta}
\usepackage[caption=false]{subfig}
\usepackage{clipboard}

\usepackage[hidelinks]{hyperref}

\theoremstyle{definition}
\newtheorem{theorem}{Theorem}
\newtheorem{proposition}{Proposition}
\newtheorem{corollary}{Corollary}[theorem]
\newtheorem{lemma}{Lemma}
\newtheorem{definition}{Definition}
\newtheorem{assumption}{Assumption}
\newtheorem{remark}{Remark}

\tcbset{dfppibox/.style={fonttitle=\bfseries,boxrule=0.5pt,arc=1pt,breakable,left=4pt,right=4pt,top=3pt,bottom=3pt,before skip=8pt,after skip=8pt}}
\tcolorboxenvironment{definition}{colback=gray!5!white,colframe=black,dfppibox}
\tcolorboxenvironment{theorem}{colback=gray!5!white,colframe=black,dfppibox}
\tcolorboxenvironment{corollary}{colback=gray!5!white,colframe=black,dfppibox}
\tcolorboxenvironment{lemma}{colback=gray!5!white,colframe=black,dfppibox}
\tcolorboxenvironment{proposition}{colback=gray!5!white,colframe=black,dfppibox}
\tcolorboxenvironment{remark}{colback=gray!5!white,colframe=black,dfppibox}
\tcolorboxenvironment{assumption}{colback=white,colframe=black,dfppibox}

\newcolumntype{N}[1]{>{\raggedright\arraybackslash}p{#1}}
\newcolumntype{M}{>{\raggedright\arraybackslash}X}
\newcommand{\nrule}{\arrayrulecolor{black!22}\hline\arrayrulecolor{black!60}}
\newcommand{\grule}{\arrayrulecolor{black!60}\hline}
\newenvironment{notationtable}{\arrayrulecolor{black!60}%
  \tabularx{\textwidth}{|N{0.46\textwidth}|M|}\grule}{\grule\endtabularx\arrayrulecolor{black}}

\newcommand{\mdl}{\mathtt{mdl}}
\newcommand{\E}{\mathbb{E}}
\newcommand{\Var}{\mathrm{Var}}
\newcommand{\Cov}{\mathrm{Cov}}
\newcommand{\hmu}{\widehat{\mu}}
\newcommand{\hg}{\widehat{g}}
\newcommand{\hr}{\widehat{r}}
\newcommand{\hlam}{\widehat{\lambda}}
\newcommand{\homg}{\widehat{\omega}}
\newcommand{\htheta}{\widehat{\theta}}
\newcommand{\hV}{\widehat{V}}
\newcommand{\hzeta}{\widehat{\tau}}
\newcommand{\hA}{\widehat{A}}
\newcommand{\hB}{\widehat{B}}
\newcommand{\hC}{\widehat{C}}
\newcommand{\hD}{\widehat{D}}

\graphicspath{{figures/}}
\title{\Copy{title}{Prediction-Powered Data Fusion for Treatment Effect Estimation}}

\author{
Yonghan Jung$^{1}$ and Shu Yang$^{2}$ \\[2pt]
\small $^{1}$ University of Illinois Urbana-Champaign, \texttt{yonghan@illinois.edu} \\
\small $^{2}$ North Carolina State University, \texttt{syang24@ncsu.edu}
}

\date{\today}
\begin{document}
\maketitle
\allowdisplaybreaks

%%%%%%%%%%%%%%%%%%%%%%%%%%%%%%%%%%%%%%%%%%%%%%%%%%%%%%%
% Abstract
%%%%%%%%%%%%%%%%%%%%%%%%%%%%%%%%%%%%%%%%%%%%%%%%%%%%%%%
\begin{abstract}
Randomized controlled trials (RCTs) identify treatment effects without confounding but are often small, whereas observational studies (OBS) are large but may be confounded. Many estimators combining a small RCT with a large OBS have been developed for the average treatment effect (ATE) and the conditional ATE (CATE). However, existing ATE estimators either make assumptions on the OBS or do not borrow enough power from them. The CATE has been studied less than the ATE. Existing CATE methods either assume the OBS are unconfounded, rely on a model of the confounding function, or accept bias in exchange for lower variance. We therefore propose a framework that, without special assumptions on the OBS, fuses the OBS and the RCT by preserving the unbiasedness of RCT-based estimation while borrowing power from the large OBS to boost precision. Applying this principle, we build an ATE estimator, AIPW-Fusion, with closed-form weights and confidence intervals, and two CATE learners, DR-Fusion and R-Fusion. Experiments corroborate our findings.

\end{abstract}

%%%%%%%%%%%%%%%%%%%%%%%%%%%%%%%%%%%%%%%%%%%%%%%%%%%%%%%
% Table of contents
%%%%%%%%%%%%%%%%%%%%%%%%%%%%%%%%%%%%%%%%%%%%%%%%%%%%%%%
\tableofcontents

%%%%%%%%%%%%%%%%%%%%%%%%%%%%%%%%%%%%%%%%%%%%%%%%%%%%%%%
% Main text
%%%%%%%%%%%%%%%%%%%%%%%%%%%%%%%%%%%%%%%%%%%%%%%%%%%%%%%
\begingroup
\allowdisplaybreaks
\sloppy
\section{Introduction}\label{sec:intro}

Randomized controlled trials (RCTs) are the gold standard for identifying treatment effects but are often small, whereas observational studies (OBS) are large but may be confounded. This paper fuses the two to estimate the trial-population conditional and average treatment effects.

Transportability and data-fusion methods \citep{bareinboim2016fusion,colnet2024review,lee2022survival,shi2023integration,jung2023joint,jung2023gid,lee2023generalizability,jung2024uca} identify and estimate effects that no single source identifies, such as effects in a new target population or of joint treatments. Most of them do not address cases where the RCT already identifies the target but is too small to estimate it precisely. Our work is closer to robust borrowing methods. For the ATE, they improve a trial estimate by combining it with OBS estimates or OBS-based predictions \citep{yang2020combining,rosenman2023shrinkage,gao2025borrowing,gagnonbartsch2023precise,debartolomeis2025efficient}, or by prediction-powered inference \citep[PPI;][]{angelopoulos2023ppi,angelopoulos2026ppipp,demirel2024ppgen}. Compared with ATE fusion, only a few fusion methods have been developed for CATE estimation, which use the trial to correct the bias of OBS-based estimates or the OBS to reduce the variance of trial-based ones \citep{kallus2018removing,cheng2021adaptive,wu2022integrative,li2022robust,yang2023elastic,yang2025hte,hatt2022combining,gu2023fast,asiaee2025improving}. Section~\ref{sec:related} compares them with ours.

In this paper, we enhance treatment effect estimators from a small RCT with large, potentially heavily confounded OBS data while keeping them unbiased. Specifically, our contributions are:
\begin{enumerate}[label=(\roman*),leftmargin=*]
\item \textbf{An ATE estimator}, AIPW-Fusion (AIPWF), that stays unbiased under any OBS confounding and combines two approaches, leveraging the OBS regressions plugged into the RCT's AIPW score and a PPI correction, with a valid interval under an estimated ratio (Sec.~\ref{sec:ate}).
\item \textbf{Two CATE learners}, DR-Fusion (DRF) and R-Fusion (RF), that keep the trial's CATE target under any OBS confounding, borrow without modeling the confounding function, and whose data-driven borrowing provably does almost no worse than the trial alone (Sec.~\ref{sec:cate}).
\item \textbf{Experiments} on synthetic data and the IHDP and ACIC~2016 benchmarks, where our estimators outperform trial-only estimation, PPI, and naive pooling (Sec.~\ref{sec:experiments}).
\end{enumerate}

Code is available at \url{https://github.com/CausalDataScience/DataFusionPPI}.

Section~\ref{sec:setup} sets up the problem. Sections~\ref{sec:ate} and~\ref{sec:cate} develop the ATE estimator and the CATE learners. Section~\ref{sec:experiments} reports the experiments. The appendix contains the proofs and additional results.

\subsection{Related Work}
\label{sec:related}

\paragraph{Transportability.} Transportability and data-fusion methods \citep{pearl2011transportability,bareinboim2012completeness,pearl2014external,bareinboim2016fusion,lee2023generalizability,lee2022survival,jung2023joint,jung2023gid,jung2024uca} combine sources to identify effects that no single source identifies, as reviewed by \citet{colnet2024review} and \citet{shi2023integration}. We instead borrow from the OBS to estimate more efficiently an effect that the trial already identifies.

\paragraph{Causal online learning.}\citet{zhang2017transfer} bound the expected reward of each arm from confounded observational data and clip the arm's upper confidence bound at these bounds. \citet[Sec.~9.2]{bareinboim2025causalai} develops this as causal offline-to-online learning (COOL) \citep[see also][Sec.~5]{bareinboim2025crl}. The clipping helps only while the experimental bound is looser than the causal bounds, and it assumes one population; we let the OBS differ from the trial in both the covariate law and the outcome regression (Assumption~\ref{ass:exact-transport}).

\paragraph{ATE fusion.}
Table~\ref{tab:lit-ate} compares the ATE fusion methods. Methods that put OBS outcomes directly into the estimate are not unbiased under every OBS confounding (\citealp{yang2020combining,rosenman2023shrinkage,gao2025borrowing}; \citealp{dang2025escvtmle}). \citet{vanderlaan2026atmle} also pool the OBS outcomes but subtract an estimate of the OBS bias, so their estimator is consistent, though not unbiased, under any OBS confounding. Other methods add a prediction from external data as a covariate in the trial's estimator \citep{gagnonbartsch2023precise,debartolomeis2025efficient}. A common choice is the prognostic score, each participant's predicted outcome under control from a model trained on historical data \citep{schuler2022prognostic,liao2025prognostic,hojbjerre2026prognostic}. As in standard covariate adjustment in trials \citep{moore2009covariate,rosenblum2010simple,moore2011robust}, randomization keeps these estimators unbiased under any OBS confounding. Two methods instead generalize the trial ATE to a different target population: \citet{lee2023generalizability} weight the trial to match the target's covariates, and \citet{demirel2024ppgen} apply PPI \citep{angelopoulos2023ppi} by averaging predictions over the target covariates. \citet{vanderlaan2026calibeating} show that PPI and PPI++ are AIPW estimators, the latter with empirical efficiency maximization \citep{rubin2008efficiency}. AIPWF combines the two channels of the methods above, putting OBS regressions into the trial's AIPW score and a PPI correction over the OBS covariates, with coefficients chosen jointly in closed form (Sec.~\ref{sec:ate}); Appendix~\ref{app:prog} compares it with prognostic-score adjustment.

\begin{table}[tp]
\centering\small
\caption{ATE fusion methods. A dash means that the property does not apply. [DB25] and [H26] borrow from foundation models or historical controls rather than from an OBS. [vdL26] is consistent rather than unbiased.}
\label{tab:lit-ate}
\centering\setlength{\tabcolsep}{3pt}
\begin{tabular*}{\textwidth}{@{\extracolsep{\fill}}lcccc@{}}
\toprule
Method & \begin{tabular}[b]{@{}c@{}}Borrows\\OBS outcomes\end{tabular} & \begin{tabular}[b]{@{}c@{}}Unbiased under\\any OBS\\confounding\end{tabular} & \begin{tabular}[b]{@{}c@{}}OBS regressions\\in the trial's\\AIPW score\end{tabular} & \begin{tabular}[b]{@{}c@{}}OBS covariates\\used to reduce\\the trial variance\end{tabular}\\
\midrule
{[YD20]}~\citet{yang2020combining} & $\checkmark$ & $\times$ & $\times$ & $\times$\\
{[R23]}~\citet{rosenman2023shrinkage} & $\checkmark$ & $\times$ & $\times$ & $\times$\\
{[G25]}~\citet{gao2025borrowing} & $\checkmark$ & $\times$ & $\times$ & $\times$\\
{[Da25]}~\citet{dang2025escvtmle} & $\checkmark$ & $\times$ & $\times$ & $\times$\\
{[vdL26]}~\citet{vanderlaan2026atmle} & $\checkmark$ & $\checkmark$ & $\times$ & $\times$\\
{[L23]}~\citet{lee2023generalizability} & $\times$ & -- & $\times$ & $\times$\\
{[GB23]}~\citet{gagnonbartsch2023precise} & $\checkmark$ & $\checkmark$ & $\checkmark$ & $\times$\\
{[DB25]}~\citet{debartolomeis2025efficient} & $\checkmark$ & $\checkmark$ & $\checkmark$ & $\times$\\
{[H26]}~\citet{hojbjerre2026prognostic} & $\checkmark$ & $\checkmark$ & $\checkmark$ & $\times$\\
{[D24]}~\citet{demirel2024ppgen} & $\checkmark$ & $\checkmark$ & $\times$ & $\checkmark$\\
\midrule
AIPWF (ours) & $\checkmark$ & $\checkmark$ & $\checkmark$ & $\checkmark$\\
\bottomrule
\end{tabular*}
\end{table}

\paragraph{CATE fusion.}
Table~\ref{tab:lit-cate} compares the CATE fusion methods. Existing methods keep the trial's target only if the OBS are unconfounded \citep{li2022robust}, only if the OBS bias follows a parametric, confounding-function, or representation model \citep{kallus2018removing,wu2022integrative,yang2025hte,hatt2022combining}, or only up to a bias accepted for lower variance \citep{cheng2021adaptive,yang2023elastic,gu2023fast}. \citet{asiaee2025improving} keep this target under any OBS confounding by modeling the discrepancy between OBS predictions and trial outcome means. DRF and RF need no such model: they borrow through the two AIPWF channels inside orthogonal DR- and $R$-learner losses (Sec.~\ref{sec:cate}). Related work learns treatment rules for an observational target population \citep{wu2023itr} or estimates effects within strata of predicted risk \citep{kent2020path}.

\begin{table}[tp]
\centering\small
\caption{CATE fusion methods. [A25] is marked orthogonal because the conditional mean of its pseudo-outcome is the trial's CATE for any augmentation function.}
\label{tab:lit-cate}
\centering
\setlength{\tabcolsep}{3pt}
\begin{tabular*}{\textwidth}{@{\extracolsep{\fill}}lccccc@{}}
\toprule
Method & \begin{tabular}[b]{@{}c@{}}Borrows when\\the OBS is\\confounded\end{tabular} & \begin{tabular}[b]{@{}c@{}}No model of\\the OBS bias\end{tabular} & \begin{tabular}[b]{@{}c@{}}Flexible\\CATE class\end{tabular} & \begin{tabular}[b]{@{}c@{}}Orthogonal\\loss or score\end{tabular} & \begin{tabular}[b]{@{}c@{}}Keeps the trial's\\target under any\\OBS confounding\end{tabular}\\
\midrule
{[L22]}~\citet{li2022robust} & $\times$ & $\checkmark$ & $\checkmark$ & $\checkmark$ & $\times$\\
{[K18]}~\citet{kallus2018removing} & $\checkmark$ & $\times$ & $\checkmark$ & $\times$ & $\times$\\
{[WY22]}~\citet{wu2022integrative} & $\checkmark$ & $\times$ & $\checkmark$ & $\checkmark$ & $\times$\\
{[Y25]}~\citet{yang2025hte} & $\checkmark$ & $\times$ & $\times$ & $\checkmark$ & $\times$\\
{[H22]}~\citet{hatt2022combining} & $\checkmark$ & $\times$ & $\checkmark$ & $\times$ & $\times$\\
{[CC21]}~\citet{cheng2021adaptive} & $\times$ & $\checkmark$ & $\checkmark$ & $\checkmark$ & $\times$\\
{[Y23]}~\citet{yang2023elastic} & $\times$ & $\checkmark$ & $\times$ & $\checkmark$ & $\times$\\
{[G23]}~\citet{gu2023fast} & $\checkmark$ & $\checkmark$ & $\checkmark$ & $\times$ & $\times$\\
{[A25]}~\citet{asiaee2025improving} & $\checkmark$ & $\times$ & $\checkmark$ & $\checkmark$ & $\checkmark$\\
\midrule
DRF/RF (ours) & $\checkmark$ & $\checkmark$ & $\checkmark$ & $\checkmark$ & $\checkmark$\\
\bottomrule
\end{tabular*}
\end{table}

% Fig. 1 wraps the first paragraph; start the section on a new page when fewer than 14 lines remain.
\needspace{14\baselineskip}
\section{Background and Setting}
\label{sec:setup}

\begin{wrapfigure}{r}{0.36\textwidth}
	\vspace{-1.2\baselineskip}
	\centering
	\definecolor{coltreat}{HTML}{0066CC}
	\definecolor{colout}{HTML}{D62828}
	\definecolor{collatent}{HTML}{A6A6A6}
	\definecolor{filllatent}{HTML}{F2F2F2}
	\tikzset{
 dag/.style={>={Latex[width=1.1mm,length=1.4mm]}},
 varnode/.style={draw,circle,inner sep=0mm,minimum size=5mm,font=\footnotesize,line width=0.5pt},
 observed/.style={varnode,draw=black,text=black,fill=white},
 latent/.style={varnode,draw=collatent,text=collatent,fill=filllatent},
 treatment/.style={varnode,draw=coltreat,text=coltreat,fill=white},
 outcome/.style={varnode,draw=colout,text=colout,fill=white},
 observed edge/.style={->,draw=black,line width=0.48pt},
 latent edge/.style={->,draw=collatent,opacity=0.72,line width=0.48pt},
	}
	\subfloat[RCT]{%
 \begin{tikzpicture}[dag,scale=0.55]
 \node[observed] (X) at (0,2.598) {$X$};
 \node[treatment] (A) at (-1.5,0) {$A$};
 \node[outcome] (Y) at (1.5,0) {$Y$};
 \draw[observed edge] (X) -- (A);
 \draw[observed edge] (X) -- (Y);
 \draw[observed edge] (A) -- (Y);
 \end{tikzpicture}}%
	\hspace{0.6cm}%
	\subfloat[OBS]{%
 \begin{tikzpicture}[dag,scale=0.55]
 \node[observed] (X) at (0,2.598) {$X$};
 \node[treatment] (A) at (-1.5,0) {$A$};
 \node[outcome] (Y) at (1.5,0) {$Y$};
 \node[latent] (U) at (0,0.866) {$U$};
 \draw[observed edge] (X) -- (A);
 \draw[observed edge] (X) -- (Y);
 \draw[observed edge] (A) -- (Y);
 \draw[latent edge] (U) -- (A);
 \draw[latent edge] (U) -- (Y);
 \draw[latent edge] (U) -- (X);
 \end{tikzpicture}}
	\caption{Causal diagrams; $U$ is unmeasured.}
	\label{fig:rct-obs-dags}
	\vspace{-0.6\baselineskip}
\end{wrapfigure}
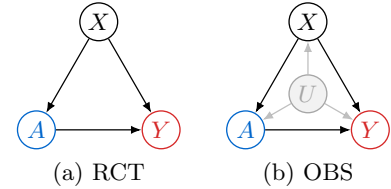

Fig.~\ref{fig:rct-obs-dags} describes the data-generating processes for RCT and OBS. Each unit has covariates $X\in\mathcal X$, a binary treatment $A\in\{0,1\}$, and an outcome $Y$; $P_R$ and $P_O$ denote the RCT and OBS distributions of $V=(X,A,Y)$. In the RCT, treatment $A$ depends only on the covariates through a known propensity $e(x)$. In the OBS, an unmeasured confounder $U$ affects both treatment $A$ and outcome $Y$. With potential outcomes $Y(1)$ and $Y(0)$, the targets are the RCT-population conditional and average treatment effects $\tau_0(x)$ and $\theta_0$. Tables~\ref{tab:notation} and~\ref{tab:notation-cate} collect the notation.

\begin{assumption}[Trial identification]
\label{ass:trial}
	(i)~$Y=Y(A)$ a.s.; (ii)~$\{Y(0),Y(1)\}\perp A\mid X$ under $P_R$; (iii)~$e$ is known and bounded away from $0$ and $1$.
\end{assumption}

Each source $S\in\{R,O\}$ is split into three i.i.d.\ samples from $P_S$: nuisance, tuning, and evaluation samples $D_S^{\mathrm{nuis}}$, $D_S^{\mathrm{tune}}$, and $D_S^{\mathrm{eval}}$ (Table~\ref{tab:notation}). The three roles can be rotated across the samples and the resulting estimates averaged (cross-fitting), as in Sec.~\ref{sec:experiments}; our exact results are for a single split.

\begin{table}[tp]
\centering\small
\caption{Notation for the ATE (Secs.~\ref{sec:setup} and~\ref{sec:ate}).}
\label{tab:notation}
\begin{notationtable}
$\tau_0(x)\triangleq\E_R[Y(1)\!-\!Y(0)\!\mid\! X\!=\!x]$ & CATE (target)\\ \nrule
$\theta_0\triangleq\E_R[\tau_0(X)]$ & ATE (target)\\ \nrule
$e(x)\triangleq P_R(A=1\mid X=x)$ & RCT propensity\\ \nrule
$r_0\triangleq dP_R^X/dP_O^X$; $\hr$ & density ratio; estimate\\ \nrule
$D_S^{\mathrm{nuis}}$, $D_S^{\mathrm{tune}}$, $D_S^{\mathrm{eval}}$ & samples by role\\ \nrule
$n$, $N$ & RCT, OBS eval.\ sizes\\ \nrule
$\mathbb P_{R,n}$, $\mathbb P_{O,N}$ & eval.-sample means\\ \nrule
$\widehat r_{\mathrm{RZ}}$, $\widehat r_{\mathrm{CLS}}$ & base ratios (Sec.~\ref{sec:ratio})\\ \nrule
$\mathtt{imb}_f(\widetilde r)\triangleq\E_O[\widetilde rf]-\E_R[f]$ & imbalance along $f$ (Prop.~\ref{prop:drift})\\ \grule
$\mu_S(x,a)\triangleq\E_S[Y\!\mid\! X\!=\!x,A\!=\!a]$ & outcome regression\\ \nrule
$\widehat\mu_S$ & fit of $\mu_S$ on $D_S^{\mathrm{nuis}}$\\ \nrule
$\hg\triangleq\widehat\mu_O(\cdot,1)-\widehat\mu_O(\cdot,0)$ & OBS effect prediction\\ \nrule
$\lambda$ & OBS weight in score\\ \nrule
$\omega$ & PPI-correction weight\\ \nrule
$\widehat\mu_\lambda\triangleq(1-\lambda)\widehat\mu_R+\lambda\widehat\mu_O$ & blended regressor\\ \nrule
$\varphi(V;q)$ & AIPW score (Eq.~\eqref{eq:aipw})\\ \nrule
$Z^\lambda\triangleq\varphi(V;\widehat\mu_\lambda)$; $Z^0$ & fused score; RCT score\\ \nrule
$\Delta\triangleq Z^1-Z^0$ & score difference\\
\end{notationtable}
\end{table}

\begin{assumption}[Sample separation]
\label{ass:honest-separation}
	The six samples $D_R^{\mathrm{nuis}}$, $D_R^{\mathrm{tune}}$, $D_R^{\mathrm{eval}}$, $D_O^{\mathrm{nuis}}$, $D_O^{\mathrm{tune}}$, and $D_O^{\mathrm{eval}}$ are mutually independent.
\end{assumption}

 All displayed moments are finite; stronger moment conditions are stated where needed.

We consider two settings for the covariate distributions $P_R^X$ and $P_O^X$: the RCT and OBS share the same domain, or they come from heterogeneous domains. In both, the OBS may be confounded. Whenever $P_R^X$ is absolutely continuous with respect to $P_O^X$, written $P_R^X\ll P_O^X$ (every covariate region with positive probability in the RCT also has positive probability in the OBS), the density ratio $r_0$ exists.

\begin{assumption}[Covariate domains]
\label{ass:domain}
One of the following holds.
\begin{enumerate}[label=(\roman*),ref=\theassumption(\roman*),leftmargin=*,itemsep=0pt,topsep=1pt,parsep=0pt]
\item (Same domain.) $P_R^X=P_O^X$, so $r_0\equiv1$.\label{ass:common-marginal}
\item (Heterogeneous domains.) $P_R^X\ll P_O^X$, so $r_0$ exists; it need not be known.\label{ass:exact-transport}
\end{enumerate}
\end{assumption}

\begin{lemma}[RCT-valid AIPW score]
\label{lem:aipw}
For any fixed measurable $q:\mathcal X\times\{0,1\}\to\mathbb R$, the AIPW score, with $e$ evaluated at $X$,
\begin{equation}
\varphi(V;q)\triangleq q(X,1)-q(X,0)+\tfrac{A}{e}\{Y-q(X,1)\}-\tfrac{1-A}{1-e}\{Y-q(X,0)\}
\label{eq:aipw}
\end{equation}
satisfies $\E_R[\varphi(V;q)\mid X]=\tau_0(X)$ under Assumption~\ref{ass:trial} (Appendix~\ref{app:proofs}).
\end{lemma}

Hence the RCT-only AIPW estimator $\mathbb P_{R,n}Z^0$ is conditionally unbiased for $\theta_0$ given $D_R^{\mathrm{nuis}}$.

\section{Fusion for the ATE}
\label{sec:ate}

In this section, we build the ATE estimator, called \textit{AIPWF}, that uses a small RCT sample and borrows power from possibly confounded OBS.

\subsection{Estimator, Variance, and Oracles}
\label{sec:ate-est}

Our estimator starts from Lemma~\ref{lem:aipw}: $\mathbb P_{R,n}Z^0$ is the RCT-only AIPW estimator, and $\mathbb P_{R,n}Z^\lambda$ is also unbiased for any $\lambda\in\mathbb R$; that is, we can use the blended regressor $\widehat\mu_\lambda$ (Table~\ref{tab:notation}) rather than the RCT-only regressor $\widehat\mu_R$ to borrow power from the OBS \citep{gagnonbartsch2023precise,debartolomeis2025efficient}. Augmenting $\mathbb P_{R,n}Z^\lambda$ with the PPI correction \citep{angelopoulos2023ppi,angelopoulos2026ppipp}, we define the \textit{AIPWF} estimator:

\begin{definition}[AIPWF estimator]
\label{def:ate-estimator}
For fixed $r:\mathcal X\to[0,\infty)$ and $(\lambda,\omega)$, the AIPWF estimator of $\theta_0$ is
\begin{equation}
 \label{eq:ate}
 \htheta_r(\lambda,\omega)
 \triangleq \mathbb P_{R,n}Z^\lambda
 +\omega\{\mathbb P_{O,N}(r\hg)-\mathbb P_{R,n}\hg\},
\end{equation}
with $r$ and $\hg$ evaluated at $X$. Equivalently, $\htheta_r(\lambda,\omega)$ is a sum of two independent sample means, an RCT term and an OBS term:
\begin{equation}
 \label{eq:ate-split}
 \htheta_r(\lambda,\omega)=\mathbb P_{R,n}(Z^\lambda-\omega\hg)+\mathbb P_{O,N}(\omega r\hg).
\end{equation}
\end{definition}
At $(\lambda,\omega)=(0,0)$ this is the RCT-only AIPW estimator $\mathbb P_{R,n}Z^0$ \citep{robins1994estimation}. The coefficient $\lambda$ sets the blended regressor $\widehat\mu_\lambda$ in the AIPW score, and $\omega$ scales a transported control variate built from $\hg$ (Fig.~\ref{fig:geometry}(a)).
In this subsection, the density ratio is exact, $r=r_0$; Sec.~\ref{sec:ratio} estimates it.

\begin{theorem}[Mean and variance]
\label{thm:mean-var}
For every fixed $(\lambda,\omega)$ and given the fitted regressions $\hmu_R$ and $\hmu_O$,
\begin{align}
\E[\htheta_{r_0}(\lambda,\omega)]=\theta_0,
\label{eq:ate-unbiased}
\end{align}
where $\E$ and $\Var$ are over $D_R^{\mathrm{eval}}$ and $D_O^{\mathrm{eval}}$. Let
\begin{align*}
A&\triangleq\Var_R(\Delta), & C&\triangleq\Cov_R(Z^0,\Delta),\\
B&\triangleq\Var_R(\hg)+\tfrac nN\Var_O(r_0\hg), & D&\triangleq\Cov_R(Z^0,\hg).
\end{align*}
If these moments are finite, then
\begin{equation}
\label{eq:j}
n\Var\{\htheta_{r_0}(\lambda,\omega)\}=\Var_R(Z^0)+(A\lambda^2+2C\lambda)+(B\omega^2-2D\omega).
\end{equation}
If $A,B>0$, its unique minimizer is
\begin{align}
\lambda^\star=-C/A\quad\text{and}\quad\omega^\star=D/B.
\label{eq:oracle}
\end{align}
\end{theorem}
\noindent

\begin{figure*}[tp]
	\centering
	\subfloat[Where $\lambda$ and $\omega$ act (schematic)]{%
 \begin{tikzpicture}[font=\footnotesize,>=stealth,
 bx/.style={draw,rounded corners=1pt,inner sep=2pt,align=center,fill=white}]
 \useasboundingbox (0,0) rectangle (5.35,3.7);
 \node[blue!70!black,inner sep=0pt] at (2.65,3.5) {in the AIPW score};
 \node[bx] (muR) at (0.38,2.95) {$\hmu_R$};
 \node[bx] (muO) at (0.38,2.15) {$\hmu_O$};
 \node[bx,draw=blue!70!black,text=blue!70!black] (bl) at (1.97,2.55) {blend by $\lambda$};
 \node[bx] (Z) at (4.3,2.55) {AIPW score\\ $Z^\lambda$ on RCT};
 \node[bx] (g) at (0.38,1.0) {$\hg$};
 \node[bx] (c) at (1.95,1.0) {OBS mean $-$\\ RCT mean\\ of $\hg$};
 \node[draw,circle,inner sep=1pt] (p) at (4.3,1.0) {$+$};
 \node (th) at (5.0,1.0) {$\htheta$};
 \draw[->] (muR) -- (bl); \draw[->] (muO) -- (bl); \draw[->] (bl) -- (Z);
 \draw[->] (muO) -- (g);
 \draw[->] (g) -- (c);
 \draw[->,blue!70!black] (Z) -- (p);
 \draw[->,orange!85!black] (c) -- node[above]{$\times\,\omega$} (p);
 \draw[->] (p) -- (th);
 \node[orange!85!black,inner sep=0pt] at (1.95,0.17) {PPI correction};
 \end{tikzpicture}}\hfill
	\subfloat[Quadratic in $(\lambda,\omega)$ (schematic)]{%
 \begin{tikzpicture}[x=3.15cm,y=2.5cm,font=\footnotesize]
 \useasboundingbox[x=1cm,y=1cm] (-1.3,-0.55) rectangle (4.05,3.15);
 \begin{scope}
 \clip (0,0) rectangle (1.08,1.08);
 \foreach \s in {0.55,1.1,1.7,2.35}{\draw[gray!55] (0.6,0.55) ellipse [x radius=\s cm, y radius=0.62*\s cm];}
 \end{scope}
 \draw[->] (0,0) -- (1.1,0) node[right]{$\lambda$};
 \draw[->] (0,0) -- (0,1.1) node[above]{$\omega$};
 \draw[->,thick,orange!85!black] (0.03,0.03) -- (0.03,0.52) node[midway,right]{$-D^2/B$};
 \draw[->,thick,blue!70!black] (0.03,0.55) -- (0.57,0.55) node[midway,above]{$-C^2/A$};
 \fill (0,0) circle (1.8pt) node[below]{RCT only};
 \fill (0,0.55) circle (1.8pt) node[left]{PPI$^{++}$};
 \fill (0.6,0) circle (1.8pt) node[below]{$\lambda$ only};
 \fill[red!80!black] (0.6,0.55) circle (2.2pt) node[right,align=left,text=red!80!black]{AIPWF\\ $(\lambda^\star,\omega^\star)$};
 \end{tikzpicture}}\hfill
	\subfloat[Ratio calibration (schematic)]{%
 \begin{tikzpicture}[font=\footnotesize,>=stealth,
 bx/.style={draw,rounded corners=1pt,inner sep=2pt,align=center,fill=white}]
 \useasboundingbox (0,0) rectangle (5.35,3.7);
 \node[bx] (rz) at (0.45,3.33) {$\hr_{\mathrm{RZ}}$};
 \node[bx] (cl) at (0.45,2.6) {$\hr_{\mathrm{CLS}}$};
 \node (base) at (1.62,2.97) {$\hr_{\mathrm{base}}$};
 \node[bx,draw=teal!60!black,text=teal!60!black] (cal) at (3.05,2.97) {calibration};
 \node (bal) at (4.7,2.97) {$\hr_{\mathrm{BAL}}$};
 \draw[->] (rz.east) -- (base.west); \draw[->] (cl.east) -- (base.west);
 \draw[->] (base) -- (cal); \draw[->,teal!60!black] (cal) -- (bal);
 \draw[gray,densely dashed] (cal.south west) -- (0.12,2.15); \draw[gray,densely dashed] (cal.south east) -- (5.3,2.15);
 \draw[gray,rounded corners=1pt] (0.12,0.05) rectangle (5.3,2.15);
 \fill[gray!55] (0.32,1.79) rectangle ++(0.17,0.14); \node[anchor=west] at (0.49,1.86) {$\hr_{\mathrm{base}}$};
 \fill[teal!60!black] (1.72,1.79) rectangle ++(0.17,0.14); \node[anchor=west] at (1.89,1.86) {$\hr_{\mathrm{BAL}}$};
 \draw[teal!60!black,densely dashed] (3.17,1.86) -- (3.42,1.86); \node[anchor=west] at (3.42,1.86) {$\pm\widehat\delta_k$};
 \draw (0.8,0.99) -- (5.12,0.99); \node[left] at (0.8,0.99) {$0$};
 \draw[teal!60!black,densely dashed] (0.8,1.11) -- (5.12,1.11); \draw[teal!60!black,densely dashed] (0.8,0.87) -- (5.12,0.87);
 \foreach \x/\h/\c in {1.15/0.55/0.04,1.9/-0.45/-0.03,2.65/0.32/0.05,3.4/0.6/-0.04,4.12/-0.5/0.03,4.82/0.27/-0.05}{
 \fill[gray!55] (\x-0.18,0.99) rectangle (\x-0.01,0.99+\h);
 \fill[teal!60!black] (\x+0.01,0.99) rectangle (\x+0.18,0.99+\c);}
 \node at (1.15,0.27) {$f_1$}; \node at (1.9,0.27) {$f_2$}; \node at (3.4,0.27) {$\cdots$}; \node at (4.82,0.27) {$f_K$};
 \end{tikzpicture}}

	\caption{ATE fusion with AIPWF (Def.~\ref{def:ate-estimator}). (a)~Where $\lambda$ and $\omega$ enter $\htheta$. (b)~Contours of Eq.~\eqref{eq:j}.  (c)~Ratio calibration (Def.~\ref{def:bal}): bars are the imbalances $\mathbb P_O^{\mathrm{nuis}}[\hr f_k]-\mathbb P_R^{\mathrm{nuis}}[f_k]$.}
	\label{fig:geometry}
\end{figure*}
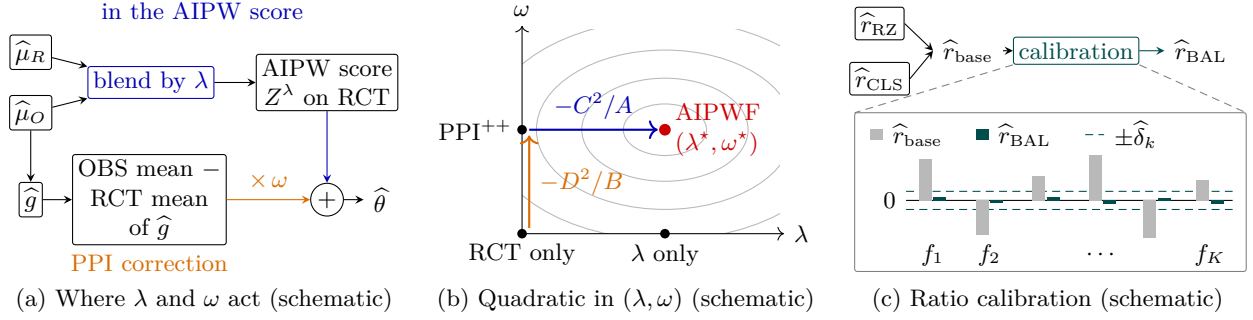
\begin{corollary}[Population oracles]
\label{cor:oracle}
 Under the conditions of Thm.~\ref{thm:mean-var} with $A,B>0$, let $\widetilde V(\lambda,\omega)\triangleq n\Var\{\htheta_{r_0}(\lambda,\omega)\}$, with RCT-only, PPI\textsuperscript{++}, and AIPWF oracle values $\widetilde V_{\mathrm{RCT}}\triangleq\widetilde V(0,0)$, $\widetilde V_{\mathrm{PPI}^{++}}\triangleq\widetilde V(0,\omega^\star)$, and $\widetilde V_{\mathrm{AIPWF}}\triangleq\widetilde V(\lambda^\star,\omega^\star)$. Then
\begin{equation*}
\widetilde V_{\mathrm{AIPWF}}=\widetilde V_{\mathrm{RCT}}-C^2/A-D^2/B\le\widetilde V_{\mathrm{PPI}^{++}}=\widetilde V_{\mathrm{RCT}}-D^2/B\le\widetilde V_{\mathrm{RCT}}.
\end{equation*}
\end{corollary}
\noindent In Fig.~\ref{fig:geometry}(b), PPI\textsuperscript{++} moves from the RCT-only origin along the $\omega$ axis, and AIPWF moves along both axes. Remark~\ref{rem:channels} shows when each axis helps.

\begin{remark}[When each channel helps]
\label{rem:channels}
 Since $D=\Cov_R(\tau_0(X),\hg(X))$, the Cauchy--Schwarz inequality gives $D^2\le\Var_R\{\tau_0(X)\}\Var_R(\hg)\le\Var_R\{\tau_0(X)\}\,B$. Hence the reduction $D^2/B$ that $\omega$ brings in Cor.~\ref{cor:oracle} is at most $\Var_R\{\tau_0(X)\}$, with or without $\lambda$, because the variance is separable in $(\lambda,\omega)$. Since $\E_R[Z^\lambda\mid X]=\tau_0(X)$ for every $\lambda$ (Lemma~\ref{lem:aipw}), $\Var_R(Z^\lambda)=\E_R\{\Var_R(Z^\lambda\mid X)\}+\Var_R\{\tau_0(X)\}$, so this cap is the effect-heterogeneity part of the score variance. It is the fraction $\Var_R\{\tau_0(X)\}/\Var_R(Z^0)$ of $\widetilde V_{\mathrm{RCT}}=\Var_R(Z^0)$, and the larger fraction $\Var_R\{\tau_0(X)\}/\Var_R(Z^{\lambda^\star})$ of the variance $\Var_R(Z^{\lambda^\star})=\widetilde V_{\mathrm{RCT}}-C^2/A$ that remains after $\lambda$. Conversely, $\Var_R(Z^{\lambda^\star})\ge\Var_R\{\tau_0(X)\}$ gives $C^2/A\le\E_R\{\Var_R(Z^0\mid X)\}$: $\lambda$ acts only on the within-covariate part of the score variance, and its reduction is zero when $\hmu_R$ equals the trial regression (Appendix~\ref{app:proofs}). Thus $\omega$ can help much only when effect heterogeneity is a large part of the score variance, and $\lambda$ helps only when $\hmu_R$ differs from the trial regression, for example when it is fitted on a small trial.
\end{remark}

\subsection{Estimating the Density Ratio}
\label{sec:ratio}
So far we have assumed $r=r_0$. This subsection gives the density-ratio estimator used in AIPWF. AIPWF is unbiased under the correct ratio (Thm.~\ref{thm:mean-var}), but a misspecified ratio introduces bias.

\begin{proposition}[Drift under a candidate ratio]
\label{prop:drift}
For a candidate ratio $\widetilde r$, let
\begin{align}
\mathtt{imb}_f(\widetilde r)\triangleq\E_O[\widetilde r f]-\E_R[f]=\E_O[(\widetilde r-r_0)f]
\label{eq:imb}
\end{align}
be its imbalance along a feature $f$. Then, for fixed $(\lambda,\omega)$ and given the fitted regressions $\hmu_R$ and $\hmu_O$,
\begin{equation}
 \label{eq:drift}
 \E[\htheta_{\widetilde r}(\lambda,\omega)]-\theta_0=\omega\, \mathtt{imb}_{\hg}(\widetilde r).
\end{equation}
\end{proposition}

Under Assumption~\ref{ass:common-marginal} we set $\hr\equiv1$. Under Assumption~\ref{ass:exact-transport} the ratio must be estimated, and two characterizations of $r_0$ give the estimators.
\begin{proposition}[Characterizations of $r_0$]
\label{prop:ratio-char}
Under Assumption~\ref{ass:exact-transport}:
\begin{enumerate}[label=(\roman*),leftmargin=*,itemsep=0pt,topsep=1pt,parsep=0pt]
\item (Reweighting.) $\E_O[r_0h]=\E_R[h]$ for every $h\in L^1(P_R^X)$, and no other weight does so for all bounded $h$.
\item (Riesz risk.) If $r_0\in L^2(P_O^X)$, then $r_0=\arg\min_{a\in L^2(P_O^X)}\{\E_O[a^2]-2\E_R[a]\}$.
\end{enumerate}
\end{proposition}
\noindent Item~(ii) follows from item~(i), since $\E_O[a^2]-2\E_R[a]=\E_O[(a-r_0)^2]-\E_O[r_0^2]$.
Two base ratios are standard. The Riesz estimator $\widehat r_{\mathrm{RZ}}$ minimizes the empirical Riesz risk $\mathbb P_O^{\mathrm{nuis}}(a^2)-2\mathbb P_R^{\mathrm{nuis}}(a)$ of Prop.~\ref{prop:ratio-char}(ii) over a linear class \citep{chernozhukov2022autodml}, with a ridge penalty. The classifier estimator is $\widehat r_{\mathrm{CLS}}\triangleq\{(1-\widehat\pi)/\widehat\pi\}\,\widehat s/(1-\widehat s)$, where $\widehat s(x)$ is a fitted probability of trial membership given covariates $x$, trained on the pooled nuisance covariates, and $\widehat\pi$ is the trial share.

By Prop.~\ref{prop:drift}, an estimated ratio biases AIPWF only through one moment, $\mathtt{imb}_{\hg}(\hr)=\E_O[\hr\hg]-\E_R[\hg]$. A base ratio, such as $\widehat r_{\mathrm{RZ}}$ or $\widehat r_{\mathrm{CLS}}$, approximates $r_0$ as a whole and leaves this moment nonzero in general. We therefore use calibration to achieve $\E_O[\hr\hg]-\E_R[\hg]=0$.
\begin{definition}[Calibrated balancing ratio]
\label{def:bal}
Given a base ratio $\hr_{\mathrm{base}}>0$ and features $f(X)=(f_1(X),\dots,f_K(X))$, the calibrated ratio is
\begin{equation*}
\hr_{\mathrm{BAL}}(x)\triangleq\frac{\hr_{\mathrm{base}}(x)\,e^{\widehat\xi^\top f(x)}}{\mathbb P_O^{\mathrm{nuis}}\{\hr_{\mathrm{base}}(X)e^{\widehat\xi^\top f(X)}\}},
\end{equation*}
where $\widehat\xi$ solves the balance equations, for all $k\le K$,
\begin{equation}
\mathbb P_O^{\mathrm{nuis}}\{\hr_{\mathrm{BAL}}(X)f_k(X)\}=\mathbb P_R^{\mathrm{nuis}}\{f_k(X)\}.
\label{eq:balance}
\end{equation}
\end{definition}
\noindent Among ratios that average to one on the OBS and satisfy Eq.~\eqref{eq:balance}, $\hr_{\mathrm{BAL}}$ is the closest to $\hr_{\mathrm{base}}$ in Kullback--Leibler divergence \citep{hainmueller2012entropy}. For AIPWF, $f=\hg$ suffices and is balanced exactly.  Fig.~\ref{fig:geometry}(c) illustrates it: calibration drives each feature's imbalance to zero, or within its tolerance $\widehat\delta_k$.

\paragraph{Computing the calibrated ratio.} Eq.~\eqref{eq:balance} is the stationarity condition of the convex objective $\log\mathbb P_O^{\mathrm{nuis}}\{\hr_{\mathrm{base}}e^{\xi^\top f}\}-\xi^\top\mathbb P_R^{\mathrm{nuis}}(f)$, whose minimizer gives the Kullback--Leibler projection of $\hr_{\mathrm{base}}$ onto the ratios that satisfy Eq.~\eqref{eq:balance} and average to one. For one feature, the stationarity condition says that the calibrated OBS mean of $f$ equals $\mathbb P_R^{\mathrm{nuis}}(f)$; its derivative in $\xi$ is the calibrated variance, which is nonnegative, so the root is unique when $\mathbb P_R^{\mathrm{nuis}}(f)$ lies strictly inside the range of $f$ on $D_O^{\mathrm{nuis}}$. With many features (Secs.~\ref{sec:cate-sieve} and~\ref{sec:experiments}), we add the penalty $\sum_k\widehat\delta_k|\xi_k|$ (approximate balance). The features $f$ have no intercept, $\widehat\delta_{k}$ is the standard error of the $k$th balance residual, estimated by $\widehat\delta_{k}^2=\widehat\Var_O^{\mathrm{nuis}}(\hr_{\mathrm{base}}f_{k})/N_{\mathrm{nuis}}+\widehat\Var_R^{\mathrm{nuis}}(f_{k})/n_{\mathrm{nuis}}$ with sample variances on $D_S^{\mathrm{nuis}}$ of sizes $n_{\mathrm{nuis}}$ and $N_{\mathrm{nuis}}$, and $\widehat\xi$ minimizes the objective below. The penalized dual objective
\begin{align}
\log \mathbb P_O^{\mathrm{nuis}}\{\hr_{\mathrm{base}}e^{\xi^\top f}\}-\xi^\top \mathbb P_R^{\mathrm{nuis}}(f)+\sum_{k}\widehat\delta_{k}|\xi_{k}|
\label{eq:dual}
\end{align}
is convex in $\xi$. At a minimizer $\widehat\xi$, the subgradient condition is $\mathbb P_O^{\mathrm{nuis}}(\hr_{\mathrm{BAL}}f_{k})-\mathbb P_R^{\mathrm{nuis}}(f_{k})+\widehat\delta_{k}\varsigma_{k}=0$ with $\varsigma_{k}\in\partial|\widehat\xi_{k}|$. Hence the $k$th residual is at most $\widehat\delta_{k}$ in absolute value, with equality when $\widehat\xi_{k}\neq0$. Normalization gives $\mathbb P_O^{\mathrm{nuis}}(\hr_{\mathrm{BAL}})=1$.

\subsection{Algorithm and Confidence Intervals}
\label{sec:ate-alg}

This subsection gives an algorithm for the AIPWF estimate and its confidence interval.

We restrict $(\lambda,\omega)$ to $[0,1]^2$, which contains the RCT-only estimator at $(0,0)$ and caps the PPI correction at its full weight $\omega=1$; $\lambda\in[0,1]$ keeps $\hmu_\lambda$ a convex blend of $\hmu_R$ and $\hmu_O$. Since Thm.~\ref{thm:mean-var} needs $A,B>0$ for the oracle coefficients $-C/A$ and $D/B$, we also prespecify thresholds $\epsilon_A,\epsilon_B>0$ and require $\hA>\epsilon_A$ and $\hB>\epsilon_B$.

With $\mathrm{Proj}_{[0,1]}$ the projection onto $[0,1]$, we select
\begin{align}
\hlam&{}\triangleq\mathrm{Proj}_{[0,1]}(-\hC/\hA)\,\mathbf 1\{\hA>\epsilon_A\},\notag\\
\homg&{}\triangleq\mathrm{Proj}_{[0,1]}(\hD/(\hB+\hB_{\mathrm{cal}}))\,\mathbf 1\{\hB>\epsilon_B\}.
\label{eq:coef-select}
\end{align}
Here $\hB_{\mathrm{cal}}\triangleq n\{\widehat\Var_R^{\mathrm{nuis}}(\hg)/n_{\mathrm{nuis}}+\widehat\Var_O^{\mathrm{nuis}}(\hr\hg)/N_{\mathrm{nuis}}\}$, or $0$ if $\hr\equiv1$, is the calibration noise of Thm.~\ref{thm:wald}(ii).

Algorithm~\ref{alg:honest-ate} uses the three samples of Assumption~\ref{ass:honest-separation} in turn: it fits the nuisances, with $\hr$ calibrated by Def.~\ref{def:bal} with $\hg$ among the features; selects $(\hlam,\homg)$ by Eq.~\eqref{eq:coef-select} from tuning-sample analogues of $A,B,C,D$ in Thm.~\ref{thm:mean-var}; and evaluates Eq.~\eqref{eq:ate}, adding $\widehat V_{\mathrm{cal}}$ (Thm.~\ref{thm:wald}) to $\hV$.

\begin{algorithm}[tp]
\small
\caption{AIPWF}
\label{alg:honest-ate}
\DontPrintSemicolon
\LinesNumbered\SetNlSty{}{}{:}
\KwIn{$D_S^{\mathrm{nuis}},D_S^{\mathrm{tune}},D_S^{\mathrm{eval}}$ for $S\in\{R,O\}$; $\epsilon_A,\epsilon_B>0$.}
\KwOut{$\htheta\triangleq\htheta_{\hr}(\hlam,\homg)$, $\hV$, and $\widehat V_{\mathrm{cal}}$.}
Fit $\hmu_R,\hmu_O,\hg,\hr$ and form $\hB_{\mathrm{cal}}$ on $D_S^{\mathrm{nuis}}$ ($\hr\equiv1$ under Assumption~\ref{ass:common-marginal})\;
Form $\hA,\hB,\hC,\hD$ on $D_S^{\mathrm{tune}}$; select $(\hlam,\homg)$ by Eq.~\eqref{eq:coef-select}\;
On $D_S^{\mathrm{eval}}$, compute $\htheta$ by Eq.~\eqref{eq:ate} with $r=\hr$\;
Return $\htheta=\htheta_{\hr}(\hlam,\homg)$ and $\hV\triangleq\widehat{\Var}_{R,\mathrm{eval}}(Z^{\hlam}-\homg\hg)/n+\widehat{\Var}_{O,\mathrm{eval}}(\homg\hr\hg)/N$, and $\widehat V_{\mathrm{cal}}$ (Thm.~\ref{thm:wald})\;
\end{algorithm}

\begin{remark}[Adaptive reuse]
\label{rem:reuse}
 Algorithm~\ref{alg:honest-ate} selects $(\hlam,\homg)$ on the tuning samples and evaluates $\htheta$ on separate samples, because selecting and evaluating on the same samples can bias the estimate. If $X$ is degenerate, $Z^0=0$, and $\Delta_i$ are i.i.d.\ Rademacher, every fixed $\lambda$ centers $\lambda \mathbb P_{R,n}\Delta$, but $\hlam=\mathrm{sign}(\mathbb P_{R,n}\Delta)$ yields $|\mathbb P_{R,n}\Delta|$ with strictly positive mean. An analogous construction applies to $\omega$ under a common marginal. Bias comes only from reuse.
\end{remark}

Thm.~\ref{thm:wald}(ii) uses the following high-level conditions. 
\begin{assumption}[Conditions for Thm.~\ref{thm:wald}(ii)]
\label{ass:wald}
$n,N\to\infty$ with $n/N$, $n_{\mathrm{nuis}}/n$, and $N_{\mathrm{nuis}}/N$ converging; $\homg\to_p\bar\omega$; $(\mathbb P_O^{\mathrm{nuis}}-\E_O)(\hr\hg)=(\mathbb P_O^{\mathrm{nuis}}-\E_O)(\bar r\bar g)+o_p(N_{\mathrm{nuis}}^{-1/2})$ for fixed $\bar r,\bar g$ with $\hg\to\bar g$ in $L^2(P_R^X)$; for some $\eta>0$, $(2+\eta)$-th moments of $Z^{\hlam}-\homg\hg$, $\homg\hr\hg$, $\bar g$, and $\bar r\bar g$ bounded uniformly in $n$; $n(V_n+V_{\mathrm{cal}})$ converging to a positive limit with $V_{\mathrm{cal}}\triangleq\bar\omega^2\{\Var_R(\bar g)/n_{\mathrm{nuis}}+\Var_O(\bar r\bar g)/N_{\mathrm{nuis}}\}$; and $(\hV+\widehat V_{\mathrm{cal}})/(V_n+V_{\mathrm{cal}})\to_p1$.
\end{assumption}

\begin{theorem}[Confidence interval of the ATE estimator]
\label{thm:wald}
Let $\htheta=\htheta_{\hr}(\hlam,\homg)$, let $\mathtt{imb}_{\hg}$ be as in Prop.~\ref{prop:drift}, and let
\begin{align}
V_n\triangleq\Var_R(Z^{\hlam}-\homg\hg)/n+\homg^2\Var_O(\hr\hg)/N
\label{eq:vn}
\end{align}
be the variance of $\htheta$ given the nuisance and tuning samples, i.e., the sum of the variances of the two terms of Eq.~\eqref{eq:ate-split}.
\begin{enumerate}[label=(\roman*),leftmargin=*,itemsep=0pt,topsep=1pt,parsep=0pt]
\item (Mean.) Given the nuisance and tuning samples,
\begin{align}
\E[\htheta]=\theta_0+\homg\,\mathtt{imb}_{\hg}(\hr)\quad\text{and}\quad\E[\hV]=V_n;
\label{eq:wald-mean}
\end{align}
in particular, $\htheta$ is unbiased when $\mathtt{imb}_{\hg}(\hr)=0$ (e.g., $\hr=r_0$, including $\hr\equiv r_0\equiv1$ under Assumption~\ref{ass:common-marginal}, or $\hr$ perfectly calibrated, $\E_O[\hr\hg]=\E_R[\hg]$).
\item (Coverage.) Let $\hr=\hr_{\mathrm{BAL}}$ with $\hg$ among its features, and let $\widehat V_{\mathrm{cal}}\triangleq\homg^2\hB_{\mathrm{cal}}/n$ with $\hB_{\mathrm{cal}}$ from Eq.~\eqref{eq:coef-select}. Under Assumption~\ref{ass:wald},
\begin{align}
\lim_{n\to\infty}P\big(\theta_0\in[\htheta\pm z_{1-\alpha/2}(\hV+\widehat V_{\mathrm{cal}})^{1/2}]\big)=1-\alpha.
\label{eq:wald-ci}
\end{align}
\end{enumerate}
\end{theorem}
Item (i) shows that $\hr$ moves the mean only through the drift $\homg\,\mathtt{imb}_{\hg}(\hr)$, while $\hV$ stays unbiased for $V_n$. Calibration does not make the drift zero. Eq.~\eqref{eq:balance} holds on the nuisance samples, so $\mathtt{imb}_{\hg}(\hr)=(\mathbb P_R^{\mathrm{nuis}}-\E_R)\hg-(\mathbb P_O^{\mathrm{nuis}}-\E_O)(\hr\hg)$, a difference of nuisance-sample averages. It has mean approximately zero and standard deviation of order $n_{\mathrm{nuis}}^{-1/2}$, the order of the standard error. Thus, with $\hr_{\mathrm{BAL}}$, item (i) gives a nonzero bias $\homg\,\mathtt{imb}_{\hg}(\hr)$ given the nuisance samples. Averaged over the nuisance samples, this bias is zero to first order, so $\htheta$ is unbiased to first order. Item (ii) therefore treats it as noise and adds its variance $\widehat V_{\mathrm{cal}}$. It follows from the Lyapunov central limit theorem \citep[Prop.~2.27]{vandervaart1998asymptotic} for the four independent nuisance and evaluation samples.

\section{Fusion for the CATE}
\label{sec:cate}

In this section, we develop two CATE learners that borrow from the OBS in the same two ways as AIPWF: DRF, built on the AIPW pseudo-outcome of the DR-learner \citep{kennedy2023dr}, and RF, built on the $R$-learner loss \citep{nie2021rlearner}. Throughout this section, Assumptions~\ref{ass:trial} and ~\ref{ass:honest-separation} hold, either Assumption~\ref{ass:common-marginal} ($r_0\equiv1$) or Assumption~\ref{ass:exact-transport} holds, and $\lambda\in[0,1]$, so the blended regressor $\hmu_\lambda$ is a convex combination of $\hmu_R$ and $\hmu_O$ and $\E_R[Z^\lambda\mid X]=\tau_0(X)$ (Lemma~\ref{lem:aipw}). Table~\ref{tab:notation-cate} collects the CATE notation.
\begin{table}[tp]
\centering\small
\caption{Notation for the CATE (Sec.~\ref{sec:cate}); the notation of Table~\ref{tab:notation} carries over.}
\label{tab:notation-cate}
\begin{notationtable}
$\mathcal R_0(t)\triangleq\E_R[\{t(X)\!-\!\tau_0(X)\}^2]$ & CATE risk (Sec.~\ref{sec:cate-loss})\\ \nrule
$\mdl\in\{\mathrm{DRF},\mathrm{RF}\}$ & learner\\ \nrule
$m_\lambda\triangleq e\widehat\mu_\lambda(\cdot,1)+(1-e)\widehat\mu_\lambda(\cdot,0)$ & marginal regression\\ \nrule
$\kappa_{\mdl}$ & unit weight (Def.~\ref{def:fusion-losses})\\ \nrule
$Z^\lambda_{\mdl}$ & pseudo-outcome (Def.~\ref{def:fusion-losses})\\ \nrule
$\mathbb P_S^{\mathrm{tune}}$ & average over $D_S^{\mathrm{tune}}$\\ \nrule
$\widehat{\mathcal R}_{\mdl}(t;\lambda,\omega,r)$ & proxy risk (Eq.~\eqref{eq:fusion-loss})\\ \nrule
$\hzeta_{\mdl}(\cdot;\lambda,\omega)$ & fitted learner (Sec.~\ref{sec:cate-select})\\ \nrule
$\mathcal G=\{(\lambda_m,\omega_m)\}_{m=1}^M$ & candidate grid (Def.~\ref{def:validation})\\ \nrule
$\widehat{\mathcal R}_{\mathrm{val}}(\hzeta_m;\omega_m)$ & validation score (Def.~\ref{def:validation})\\ \nrule
$\varrho_{\mdl}(\lambda,\omega)$ & transport error (Thm.~\ref{thm:validate})\\ \grule
$\zeta:\mathcal X\to\mathbb R^p$ & sieve basis (Def.~\ref{def:sieve})\\ \nrule
$\mathcal T_\zeta\triangleq\{\zeta^\top\beta\}$ & sieve class (Def.~\ref{def:sieve})\\ \nrule
$\widehat\beta_{\mdl}(\lambda,\omega)$ & coefficients (Def.~\ref{def:sieve})\\ \nrule
$\rho\ge0$ & ridge penalty (Def.~\ref{def:sieve})\\ \nrule
$\widehat H_{\mdl}(\omega)$ & Gram estimate (Prop.~\ref{prop:sieve-closed-form})\\ \nrule
$\Gamma_R\triangleq\E_R[\zeta\zeta^\top]$ & sieve Gram matrix\\ \nrule
$\beta_p$ & best sieve coefficients\\ \nrule
$\tau^\star_p\triangleq\zeta^\top\beta_p$ & best sieve approximation\\ \nrule
$n_R\triangleq|D_R^{\mathrm{tune}}|$, $n_O\triangleq|D_O^{\mathrm{tune}}|$ & tuning sample sizes\\ \nrule
$\bar\beta(\omega)$ & mean coefficients (Prop.~\ref{prop:sieve-moments})\\ \nrule
$\bar H_\rho$ & mean Hessian (Prop.~\ref{prop:sieve-moments})\\ \nrule
$\mathtt{bias}(\omega)$ & squared bias (Prop.~\ref{prop:sieve-moments})\\ \nrule
$\mathcal J_{\mdl}(\lambda,\omega)$ & variance (Thm.~\ref{thm:sieve-risk})\\
\end{notationtable}
\end{table}

\subsection{CATE Risks}
\label{sec:cate-loss}

For a prespecified class $\mathcal T\subseteq L^2(P_R^X)$, the CATE risk of $t\in\mathcal T$ is
\begin{align}
\mathcal R_0(t)\triangleq\E_R[\{t(X)-\tau_0(X)\}^2].
\label{eq:cate-risk}
\end{align}

The risk $\mathcal R_0$ requires the unknown $\tau_0$. We therefore define a computable proxy:
\begin{definition}[DRF and RF proxy risks]
\label{def:fusion-losses}
 For learner $\mdl\in\{\mathrm{DRF},\mathrm{RF}\}$, let
\begin{align*}
\kappa_{\mathrm{DRF}}&\triangleq1, & Z_{\mathrm{DRF}}^\lambda&\triangleq Z^\lambda,\\
\kappa_{\mathrm{RF}}&\triangleq\frac{(A-e)^2}{e(1-e)}, & Z_{\mathrm{RF}}^\lambda&\triangleq\frac{Y-m_\lambda}{A-e}.
\end{align*}
The proxy risk of learner $\mdl$ is
\begin{equation}
    \mathcal R_{\mdl}(t;\lambda)\triangleq\E_R[\kappa_{\mdl}\{Z_{\mdl}^\lambda-t(X)\}^2].
\label{eq:rct-loss}
\end{equation}
\noindent With $\mathbb P_S^{\mathrm{tune}}$ the average over $D_S^{\mathrm{tune}}$, $\omega\in\mathbb R$, and a candidate weight $r$, its empirical proxy risk is
\begin{equation}
\widehat{\mathcal R}_{\mdl}(t;\lambda,\omega,r)\triangleq\mathbb P_R^{\mathrm{tune}}[\kappa_{\mdl}\{Z_{\mdl}^\lambda-t(X)\}^2]+\omega\,\mathbb P_O^{\mathrm{tune}}[r(\hg-t)^2]-\omega\,\mathbb P_R^{\mathrm{tune}}[\kappa_{\mdl}(\hg-t)^2].
\label{eq:fusion-loss}
\end{equation}
\end{definition}

DRF regresses the AIPW score $Z^\lambda_{\mathrm{DRF}} = Z^{\lambda}$, as in the DR-learner \citep{kennedy2023dr}. RF is the $R$-learner \citep{nie2021rlearner} with each squared residual divided by $e(1-e)$, since $\{Y-m_\lambda-(A-e)t\}^2=e(1-e)\,\kappa_{\mathrm{RF}}\{Z_{\mathrm{RF}}^\lambda-t\}^2$.

When $r=r_0$, in expectation, each empirical risk equals $\mathcal R_0$ up to an additive constant that does not depend on $t$, so the two have the same minimizer.
\begin{theorem}[Exact CATE targets]
\label{prop:cate}
Fix $\omega\in\mathbb R$, $t\in\mathcal T$, and $r$. Let $Q_{\mdl}(\lambda)\triangleq \mathcal R_{\mdl}(\tau_0;\lambda)$ be the loss at the true CATE. Then for $\mdl\in\{\mathrm{DRF},\mathrm{RF}\}$,
\begin{equation*}
\E[\widehat{\mathcal R}_{\mdl}(t;\lambda,\omega,r)]=Q_{\mdl}(\lambda)+\mathcal R_0(t)+\omega\E_O[(r-r_0)(\hg-t)^2].
\end{equation*}
Here $Q_{\mdl}(\lambda)$ does not depend on $t$, and the last term is zero if $r=r_0$. In particular, if $r=r_0$,
\begin{align}
\arg\min_{t\in\mathcal T}\E[\widehat{\mathcal R}_{\mdl}(t;\lambda,\omega,r_0)]=\arg\min_{t\in\mathcal T}\mathcal R_0(t)
\label{eq:cate-argmin}
\end{align}
for all $(\lambda,\omega)$.
\end{theorem}
\noindent At $r=r_0$, borrowing from the OBS never shifts the target, and $(\lambda,\omega)$ change only the variance of the fitted learner (Secs.~\ref{sec:cate-select} and~\ref{sec:cate-sieve}). At $r\neq r_0$, the last term can shift the minimizer when $\omega>0$. The shift does not depend on $\lambda$, and for linear sieves calibration reduces it to nuisance-sampling error (Sec.~\ref{sec:cate-sieve}).

\noindent The proposed empirical risk $\widehat{\mathcal R}_{\mdl}$ is, in expectation, an orthogonal risk \citep{foster2023orthogonal} with respect to the RCT regression $\hmu_R$.
\begin{proposition}[Orthogonality of the learning equation]
\label{prop:cate-orth}
For $t,s\in\mathcal T$ and any fixed $r$, not necessarily $r_0$, let
\begin{align}
\Psi_{\mdl}(t)[s;\hmu_R,\hmu_O,r]\triangleq-\tfrac12D_t\E[\widehat{\mathcal R}_{\mdl}(t;\lambda,\omega,r)][s]
\label{eq:learning-eq}
\end{align}
be the learning equation of learner $\mdl$ at $t$ in the direction $s$, where $D_t$ is the directional derivative with respect to $t$; the learned CATE solves $\Psi_{\mdl}(t)[s]=0$ for all $s\in\mathcal T$. Then, for every $h:\mathcal X\times\{0,1\}\to\mathbb R$ and every $h_r$ with $\E_O[h_r]=0$, writing $\Delta h\triangleq h(\cdot,1)-h(\cdot,0)$, the derivative of $\Psi_{\mdl}(t)[s]$ at $\varepsilon=0$ is
\begin{enumerate}[label=(\roman*),leftmargin=*,itemsep=0pt,topsep=1pt,parsep=0pt]
\item $0$ along $\hmu_R+\varepsilon h$;
\item $\omega\E_O[(r-r_0)\,\Delta h\,s]$ along $\hmu_O+\varepsilon h$;
\item $\omega\E_O[h_r\,s\,(\hg-t)]$ along $r+\varepsilon h_r$.
\end{enumerate}
Thus $\Psi_{\mdl}$ is orthogonal to $\hmu_R$ for every $r$, to $\hmu_O$ only at $r=r_0$, and to $r$ only if $\hg=t$.
\end{proposition}
\noindent This is the orthogonality of the AIPW score and the $R$ loss to their outcome regressions under a known propensity \citep{chernozhukov2018dml,kennedy2023dr,nie2021rlearner}. At $t=\tau_0$, $\hg=t$ requires no unmeasured confounding in the OBS. In practice, we set $r=\hr$, fitted on the nuisance samples by the classifier estimator $\hr_{\mathrm{CLS}}$ (Sec.~\ref{sec:ratio}) or the calibrated ratio $\hr_{\mathrm{BAL}}$ (Def.~\ref{def:bal}).

\begin{figure*}[tp]
	\centering
	\subfloat[DRF and RF (schematic)]{%
 \begin{tikzpicture}[font=\footnotesize,>=stealth,
 bx/.style={draw,rounded corners=1pt,inner sep=2pt,align=center,fill=white}]
 \useasboundingbox (0,0) rectangle (5.35,4.0);
 \node[blue!70!black,inner sep=0pt] at (2.65,3.8) {in the pseudo-outcome};
 \node[bx] (muR) at (0.38,3.22) {$\hmu_R$};
 \node[bx] (muO) at (0.38,2.38) {$\hmu_O$};
 \node[bx,draw=blue!70!black,text=blue!70!black] (bl) at (1.97,2.8) {blend by $\lambda$};
 \node[bx] (Z) at (4.3,2.8) {RCT loss of\\ $Z^\lambda_{\mdl}-t$};
 \node[bx] (g) at (0.38,1.15) {$\hg$};
 \node[bx] (c) at (1.95,1.15) {OBS loss $-$\\ RCT loss\\ of $\hg-t$};
 \node[draw,circle,inner sep=1pt] (p) at (4.3,1.15) {$+$};
 \node[inner sep=1pt] (th) at (4.3,0.24) {$\hzeta_{\mdl}$};
 \draw[->] (muR) -- (bl); \draw[->] (muO) -- (bl); \draw[->] (bl) -- (Z);
 \draw[->] (muO) -- (g);
 \draw[->] (g) -- (c);
 \draw[->,blue!70!black] (Z) -- (p);
 \draw[->,orange!85!black] (c) -- node[above]{$\times\,\omega$} (p);
 \draw[->] (p) -- node[right]{$\min_{t}$} (th);
 \node[orange!85!black,inner sep=0pt] at (1.95,0.17) {PPI correction of the loss};
 \end{tikzpicture}}\hfill
	\subfloat[Algorithm~\ref{alg:cate} (schematic)]{%
 \begin{tikzpicture}[font=\footnotesize,>=stealth,
 bx/.style={draw,rounded corners=1pt,inner sep=1.5pt,align=center,fill=white},
 lab/.style={gray!60!black,font=\footnotesize\itshape,inner sep=0pt,anchor=west}]
 \useasboundingbox (0,0) rectangle (5.35,4.0);
 \node[lab] at (0.03,3.77) {nuisance};
 \node[lab] at (0.03,2.8) {tuning};
 \node[lab] at (0.03,1.62) {evaluation};
 \draw[densely dashed,gray] (1.53,1.04) rectangle (3.76,3.29);
 \node[bx] (N) at (2.62,3.77) {fit $\hmu_R,\hmu_O,\hg,\hr$};
 \foreach \d in {0.07,0.035}{\node[bx,draw=blue!70!black] at (2.62+\d,2.8-\d) {\phantom{$\hzeta_m\gets$}\\ \phantom{min Eq.~(88)}};}
 \node[bx,draw=blue!70!black] (T) at (2.62,2.8) {$\hzeta_m\gets$\\ min Eq.~\eqref{eq:fusion-loss}};
 \foreach \d in {0.07,0.035}{\node[bx,draw=orange!85!black] at (2.6+\d,1.62-\d) {\phantom{$\widehat{\mathcal R}_{\mathrm{val}}(\hzeta_m;\omega_m)$}\\ \phantom{Eq.~(88)}};}
 \node[bx,draw=orange!85!black] (E) at (2.6,1.62) {$\widehat{\mathcal R}_{\mathrm{val}}(\hzeta_m;\omega_m)$\\ Eq.~\eqref{eq:validation}};
 \node[bx,fill=gray!12] (S) at (4.62,1.62) {keep the\\ best $m$};
 \node[gray!60!black,inner sep=0pt,anchor=west] at (3.81,2.8) {for each $m$};
 \node[bx] (G) at (2.62,0.41) {grid $\mathcal G=\{(\lambda_m,\omega_m)\}_{m=1}^M$,\\ fixed in advance};
 \draw[->] (N) -- (T); \draw[->] (T) -- (E); \draw[->] (E.east) ++(0.07,0) -- (S.west);
 \draw[->] (G.north) -- (2.62,1.04);
 \end{tikzpicture}}\hfill
	\subfloat[Validation guarantee (schematic)]{%
 \begin{tikzpicture}[x=1cm,y=1cm,font=\footnotesize]
 \useasboundingbox (0,0) rectangle (5.35,4.0);
 \begin{scope}[shift={(0.3,0.47)},y=0.62cm,x=0.5cm]
 \foreach \i in {0,...,4}{\foreach \j in {0,...,4}{
 \pgfmathsetmacro{\v}{0.12*(\i-3)^2+0.15*(\j-2)^2+0.1}
 \pgfmathsetmacro{\c}{6+52*\v/1.8}
 \fill[blue!\c!white] (\i,\j) rectangle ++(1,1);}}
 \draw[white] (0,0) grid (5,5);
 \draw[red!75!black,thick,densely dashed] (2,1) rectangle (5,4);
 \node at (3.5,2.5) {$\star$};
 \fill[red!75!black] (2.5,3.5) circle (2pt);
 \draw[black,thick] (0.18,0.15) rectangle (0.82,0.85);
 \node[below right,inner sep=0pt] at (-0.54,-0.13) {RCT only};
 \node[below,inner sep=0pt] at (4.5,-0.13) {$\lambda$};
 \node[left,inner sep=0pt] at (-0.14,4.5) {$\omega$};
 \end{scope}
 \begin{scope}[shift={(2.93,0)},lg/.style={anchor=west,inner sep=0pt}]
 \node at (0.1,3.7) {$\star$}; \node[lg] at (0.26,3.7) {best candidate};
 \fill[red!75!black] (0.1,3.27) circle (2pt); \node[lg] at (0.26,3.27) {selected};
 \draw[red!75!black,thick,densely dashed] (0.01,2.75) rectangle (0.19,2.93); \node[lg] at (0.26,2.84) {within $2\varepsilon_{\mathrm{eval}}$};
 \node[lg] at (0.26,2.46) {of the best};
 \node[lg,gray!60!black] at (0.0,1.95) {shading:};
 \node[lg,gray!60!black] at (0.0,1.58) {CATE risk};
 \end{scope}
 \end{tikzpicture}}

	\caption{CATE fusion with DRF and RF. (a)~Where $\lambda$ and $\omega$ enter the loss (Def.~\ref{def:fusion-losses}). (b)~Algorithm~\ref{alg:cate}. (c)~Thm.~\ref{thm:validate}: shading is the CATE risk; the selected candidate is within $2\varepsilon_{\mathrm{eval}}(\alpha)$ of the best (dashed) when $\hr=r_0$.}
	\label{fig:cate-geometry}
\end{figure*}
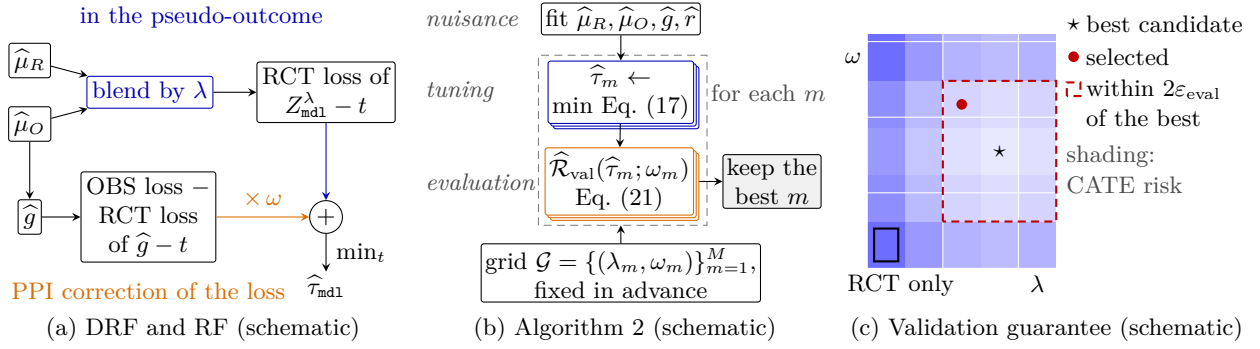
\subsection{Fitting and Selecting Coefficients}
\label{sec:cate-select}
We now select the coefficients. We define a grid $\mathcal G=\{(\lambda_{m},\omega_{m})\}_{m=1}^M\subseteq[0,1]^2$, compute for each $m$ the minimizer of the empirical proxy risk in Eq.~\eqref{eq:fusion-loss}, and choose the $(\lambda_{m},\omega_{m})$ whose minimizer has the smallest estimated CATE risk (Def.~\ref{def:validation}).

We write
\begin{align}
\hzeta_{\mdl}(\cdot;\lambda,\omega)\in\arg\min_{t\in\mathcal T}\widehat{\mathcal R}_{\mdl}(t;\lambda,\omega,\hr)
\label{eq:fitted-learner}
\end{align}
for this minimizer, the fitted fusion learner, with ties broken by a fixed rule (e.g., the smallest $\E_R[t(X)^2]$).

For grid point $m$, write $\hzeta_m\triangleq\hzeta_{\mdl}(\cdot;\lambda_m,\omega_m)$. We pick the $\hzeta_m$ with the smallest validation score, defined below.
\begin{definition}[Validation score on a finite grid]
\label{def:validation}
Let $\mathcal G\subseteq[0,1]^2$ be a prespecified finite grid containing $(0,0)$. For $m=1,\dots,M$,
\begin{equation}
\widehat{\mathcal R}_{\mathrm{val}}(\hzeta_m;\omega_m)\triangleq\mathbb P_{R,n}[\{Z^0-\hzeta_m(X)\}^2]+\omega_m\,\mathbb P_{O,N}[\hr(\hg-\hzeta_m)^2]-\omega_m\,\mathbb P_{R,n}[(\hg-\hzeta_m)^2],
\label{eq:validation}
\end{equation}
and $(\hlam_{\mdl},\homg_{\mdl})\triangleq(\lambda_{\widehat m},\omega_{\widehat m})$ with $\widehat m\in\arg\min_m\widehat{\mathcal R}_{\mathrm{val}}(\hzeta_m;\omega_m)$. The score depends on $\lambda_m$ through $\hzeta_m$. That is, $\widehat{\mathcal R}_{\mathrm{val}}(\hzeta_m;\omega_m)$ is the DRF empirical proxy risk of Eq.~\eqref{eq:fusion-loss} with $\lambda=0$ and $r=\hr$, evaluated at $t=\hzeta_m$ on the evaluation samples.
\end{definition}

Since $\widehat{\mathcal R}_{\mathrm{val}}(\hzeta_m;\omega_m)$ is the DRF proxy risk $\widehat{\mathcal{R}}_{\mathrm{DRF}}(\widehat{\tau}_m;0,\omega_m,\hr)$ evaluated at a candidate fitted on other samples, Thm.~\ref{prop:cate} applies with $\lambda=0$. When $\hr=r_0$, conditional on the fitted objects,
\begin{align}
\E[\widehat{\mathcal R}_{\mathrm{val}}(\hzeta_m;\omega_m)]=Q_{\mathrm{DRF}}(0)+\mathcal R_0(\hzeta_m).
\label{eq:val-mean}
\end{align}
The constant $Q_{\mathrm{DRF}}(0)$ is the same for every $m$, so the score ranks the candidates by CATE risk.

Since $\widehat{\mathcal R}_{\mathrm{val}}$ uses finite samples and $\hr$, its minimizer may miss the best candidate; Thm.~\ref{thm:validate} bounds this loss.
\begin{theorem}[Validation guarantee]
\label{thm:validate}
Suppose $|Z^0|$, $|\hg|$, and all $\hzeta_m$ are bounded by $B$, and $0\le \hr\le\bar r$. Let
\begin{align}
\varrho_{\mdl}(\lambda,\omega)\triangleq\omega\E_O[(\hr-r_0)\{\hg-\hzeta_{\mdl}(\cdot;\lambda,\omega)\}^2]
\label{eq:transport-error}
\end{align}
be the transport error of $\hr$ in Eq.~\eqref{eq:validation}, which is zero when $\hr=r_0$. Conditional on the nuisance and tuning samples, with probability at least $1-\alpha$,
\begin{equation}
\mathcal R_0\{\hzeta_{\mdl}(\cdot;\hlam_{\mdl},\homg_{\mdl})\}\le\min_{(\lambda,\omega)\in\mathcal G}\mathcal R_0\{\hzeta_{\mdl}(\cdot;\lambda,\omega)\}+2\varepsilon_{\mathrm{eval}}(\alpha)+2\max_{\mathcal G}|\varrho_{\mdl}|,
\label{eq:cate-selector}
\end{equation}
where $M\triangleq|\mathcal G|$ and
\begin{align}
\varepsilon_{\mathrm{eval}}(\alpha)\triangleq2B^2\sqrt{2\log(4M/\alpha)}\,(2n^{-1/2}+\bar rN^{-1/2}).
\label{eq:eps-eval}
\end{align}
\end{theorem}
\noindent That is, the risk of the selected learner exceeds the smallest risk on the grid by at most a sampling term of order $n^{-1/2}+N^{-1/2}$ and a term from the error in $\hr$. Since $(0,0)\in\mathcal G$, when $\hr=r_0$ the selected learner is at most $2\varepsilon_{\mathrm{eval}}(\alpha)$ worse than the RCT-only learner.

\subsection{Linear Sieves: Risk and Calibration}
\label{sec:cate-sieve}

We now analyze DRF and RF when $\mathcal T$ is a linear sieve \citep{newey1997series,chen2007sieve}.
% For this class, the fitted coefficients have a closed form (Prop.~\ref{prop:sieve-closed-form}), the CATE risk splits into an approximation error and an estimation error that is quadratic in $(\lambda,\omega)$ (Thm.~\ref{thm:sieve-risk}), and calibrating $\hr$ on finitely many moments removes the ratio error (Prop.~\ref{prop:sieve-balance}). Thm.~\ref{thm:validate} holds for any class but does not quantify these terms.
\begin{definition}[Linear-sieve fusion learner]
\label{def:sieve}
Let $\zeta:\mathcal X\to\mathbb R^p$ be fixed before tuning (e.g., splines, random features, or a neural network\footnotemark), and $\mathcal T_{\zeta}\triangleq\{\zeta^\top\beta:\beta\in\mathbb R^p\}$. For $\rho\ge0$, $\hzeta_{\mdl}(x;\lambda,\omega)\triangleq \zeta(x)^\top\widehat\beta_{\mdl}(\lambda,\omega)$ with
\begin{align}
\widehat\beta_{\mdl}(\lambda,\omega)\in\arg\min_{\beta}\{\widehat{\mathcal R}_{\mdl}(\zeta^\top\beta;\lambda,\omega,\hr)+\rho\|\beta\|_2^2\}.
\label{eq:sieve-fit}
\end{align}
\end{definition}
\footnotetext{A frozen neural representation trained on the nuisance samples: the network is fit on $D_S^{\mathrm{nuis}}$, and its weights are held fixed during tuning.}

 Let $\Gamma_R\triangleq\E_R[\zeta\zeta^\top]$, assumed invertible, and $\beta_p\triangleq\Gamma_R^{-1}\E_R[\zeta\tau_0]$. Then
\begin{equation*}
    \tau^\star_p\triangleq\zeta^\top\beta_p=\arg\min_{t\in\mathcal T_\zeta}\mathcal R_0(t)
\end{equation*}
is the best approximation of $\tau_0$ in the sieve class $\mathcal T_\zeta$ of Def.~\ref{def:sieve}. The estimator $\widehat\beta_{\mdl}$ has a closed form.

\begin{algorithm}[tp]
\small
\caption{DRF and RF}
\label{alg:cate}
\DontPrintSemicolon
\KwIn{$D_S^{\mathrm{nuis}},D_S^{\mathrm{tune}},D_S^{\mathrm{eval}}$ for $S\in\{R,O\}$; class $\mathcal T$; grid $\mathcal G\ni(0,0)$; a density-ratio estimator.}
\KwOut{$(\hlam_{\mdl},\homg_{\mdl})$ and $\hzeta_{\mdl}(\cdot;\hlam_{\mdl},\homg_{\mdl})$.}
Fit $\hmu_R,\hmu_O,\hg$ on $D_S^{\mathrm{nuis}}$, and $\hr$ by Def.~\ref{def:bal}\;
For each $m$, fit $\hzeta_m$ on $D_S^{\mathrm{tune}}$ by minimizing Eq.~\eqref{eq:fusion-loss}\;
For each $m$, compute $\widehat{\mathcal R}_{\mathrm{val}}(\hzeta_m;\omega_m)$ by Eq.~\eqref{eq:validation} on $D_S^{\mathrm{eval}}$\;
Select $(\hlam_{\mdl},\homg_{\mdl})$ by minimizing $\widehat{\mathcal R}_{\mathrm{val}}$ over $\mathcal G$, with deterministic tie-breaking\;
\end{algorithm}

\begin{proposition}[Normal equation]
\label{prop:sieve-closed-form}
Let $\widehat H_{\mdl}(\omega)\triangleq(1-\omega)\mathbb P_R^{\mathrm{tune}}(\kappa_{\mdl}\zeta\,\zeta^\top)+\omega\mathbb P_O^{\mathrm{tune}}(\hr \zeta\,\zeta^\top)$. If $\widehat H_{\mdl}(\omega)+\rho I$ is invertible, $\widehat\beta_{\mdl}(\lambda,\omega)$ in Def.~\ref{def:sieve} is the unique solution of the normal equation, which sets the gradient of its objective with respect to $\beta$ to zero:
\begin{equation}
\{\widehat H_{\mdl}(\omega)+\rho I\}\beta=\mathbb P_R^{\mathrm{tune}}\{\kappa_{\mdl}\zeta(Z^\lambda_{\mdl}-\omega\hg)\}+\omega\,\mathbb P_O^{\mathrm{tune}}(\hr\zeta\hg).
\label{eq:sieve-normal}
\end{equation}
\end{proposition}

Next, we bound how far $\widehat\beta_{\mdl}$ is from $\beta_p$ on average.
\begin{proposition}[Bias of $\widehat\beta_{\mdl}$]
\label{prop:sieve-moments}
Under Assumptions~\ref{ass:trial}--\ref{ass:domain}, fix $(\lambda,\omega,\rho)$ and condition on the fitted nuisances, $\hr$, and $\zeta$, so that $\E$ is over the tuning samples and $o_P$ is over the nuisance samples. Let $\bar H_\rho\triangleq\Gamma_R+\omega\,\mathtt{imb}_{\zeta\zeta^\top}(\hr)+\rho I$ be invertible, and let $\bar\beta(\omega)$ solve Eq.~\eqref{eq:sieve-normal} with each average $\mathbb P_S^{\mathrm{tune}}$ replaced by its mean $\E_S$. Let
\begin{align*}
\psi_{R,\mdl}&\triangleq\kappa_{\mdl}\zeta\{Z^\lambda_{\mdl}-\omega\hg-(1-\omega)\zeta^\top\bar\beta\},\\
\psi_O&\triangleq\omega\,\hr\,\zeta(\hg-\zeta^\top\bar\beta)
\end{align*}
be the contributions of one RCT and one OBS tuning unit to those equations, and $V_{\mdl}\triangleq\Var_R(\psi_{R,\mdl})/n_R+\Var_O(\psi_O)/n_O$. Suppose the remainder $\mathtt{rem}\triangleq\widehat\beta_{\mdl}-\bar\beta-\bar H_\rho^{-1}(\mathbb P_R^{\mathrm{tune}}\psi_{R,\mdl}+\mathbb P_O^{\mathrm{tune}}\psi_O-\rho\bar\beta)$ satisfies $\E\|\mathtt{rem}\|^2_{\Gamma_R}=o_P(n_R^{-1}+n_O^{-1})$, where $\|v\|^2_{\Gamma_R}\triangleq v^\top\Gamma_Rv$. Then $\E[\widehat\beta_{\mdl}]=\bar\beta(\omega)+o_P(n_R^{-1/2}+n_O^{-1/2})$, and
\begin{equation*}
\bar\beta(\omega)-\beta_p=\bar H_\rho^{-1}\big\{\omega\,\E_O[(\hr\!-\!r_0)\zeta(\hg\!-\!\tau^\star_p)]-\rho\beta_p\big\}.
\end{equation*}
Hence the squared bias $\mathtt{bias}(\omega)\triangleq\|\bar\beta(\omega)-\beta_p\|^2_{\Gamma_R}$ obeys the bound
\begin{equation*}
\mathtt{bias}(\omega)^{1/2}\le\|\Gamma_R^{1/2}\bar H_\rho^{-1}\|_{\mathrm{op}}\big[\rho\|\beta_p\|+\omega\{\E_O(\hr-r_0)^2\,\E_O\|\zeta(\hg-\tau^\star_p)\|^2\}^{1/2}\big].
\end{equation*}
\end{proposition}
The $\omega$ term is a product of the ratio error $\hr-r_0$ and the gap $\hg-\tau^\star_p$ of the OBS prediction from the best sieve fit, and vanishes when either factor is zero.

\begin{theorem}[Risk of a sieve candidate]
\label{thm:sieve-risk}
Under the conditions of Prop.~\ref{prop:sieve-moments}, suppose also that $\E[\mathtt{rem}]^\top\Gamma_R\{\bar\beta(\omega)-\beta_p\}=o_P(n_R^{-1}+n_O^{-1})$. Then
\begin{equation}
\E[\mathcal R_0(\hzeta_{\mdl})]=\mathcal R_0(\tau^\star_p)+\mathtt{bias}(\omega)+\mathcal J_{\mdl}(\lambda,\omega)+o_P(n_R^{-1}+n_O^{-1}),
\label{eq:sieve-risk}
\end{equation}
where $\mathtt{bias}(\omega)$ is from Prop.~\ref{prop:sieve-moments}, and
\begin{align}
\mathcal J_{\mdl}(\lambda,\omega)\triangleq\operatorname{tr}(\Gamma_R\bar H_\rho^{-1}V_{\mdl}\bar H_\rho^{-1})
\label{eq:jmdl}
\end{align}
is the $\Gamma_R$-norm variance of $\widehat\beta_{\mdl}$.
\end{theorem}

 Here $\mathcal R_0(\tau^\star_p)$ is the approximation error of the sieve, $\mathtt{bias}$ comes from the ratio imbalances and the ridge penalty, and $\mathcal J_{\mdl}$ is the estimation error. At $\hr=r_0$, both imbalances vanish, so only the ridge part $\mathtt{bias}=\rho^2\|(\Gamma_R+\rho I)^{-1}\beta_p\|^2_{\Gamma_R}$ remains.

Def.~\ref{def:bal} calibrates $\hr$ along a chosen feature vector $f$, and the right $f$ depends on the target. The next result compares the ATE and the sieve CATE.

\begin{proposition}[Calibration for ATE and sieve CATE]
\label{prop:sieve-balance}

Let $\hr$ satisfy Eq.~\eqref{eq:balance} on the nuisance samples up to the tolerances $\widehat\delta_k$. Then, for each balanced feature $f_k$,
\begin{equation*}
\mathtt{imb}_{f_k}(\hr)=(\mathbb P_R^{\mathrm{nuis}}-\E_R)f_k-(\mathbb P_O^{\mathrm{nuis}}-\E_O)(\hr f_k)+e_k,\qquad |e_k|\le\widehat\delta_k,
\end{equation*}
and likewise for their linear combinations.
\begin{enumerate}[label=(\roman*),leftmargin=*,itemsep=0pt,topsep=1pt,parsep=0pt]
\item (ATE) If $f$ contains $\hg$, the AIPWF drift $\omega\,\mathtt{imb}_{\hg}(\hr)$ is this nuisance-sampling error, which Thm.~\ref{thm:wald}(ii) absorbs into $\widehat V_{\mathrm{cal}}$.
\item (Sieve CATE) If $f$ contains $\hg^2$, $\zeta\hg$, and $\zeta_k\zeta_l$ for all $k\le l$, the $\omega$ term of $\bar\beta(\omega)-\beta_p$ in Prop.~\ref{prop:sieve-moments} and $\varrho_{\mdl}$ for every candidate in $\mathcal T_\zeta$ are linear in these errors, and hence of order $n_{\mathrm{nuis}}^{-1/2}+N_{\mathrm{nuis}}^{-1/2}$ under Assumption~\ref{ass:wald}.
\end{enumerate}
\end{proposition}
Hence, with a calibrated $\hr$, the ratio part of $\mathtt{bias}(\omega)$ in Thm.~\ref{thm:sieve-risk} is of order $n_{\mathrm{nuis}}^{-1}+N_{\mathrm{nuis}}^{-1}$ and no longer depends on how well the base ratio approximates $r_0$, and the transport error in Thm.~\ref{thm:validate} is of order $n_{\mathrm{nuis}}^{-1/2}+N_{\mathrm{nuis}}^{-1/2}$. The ATE needs one feature, while the sieve CATE needs $1+p+p(p+1)/2$, because the ratio enters the CATE through a $p$-vector and a $p\times p$ matrix rather than a scalar.  For a general class $\mathcal T$, the ratio enters through $\E_O[(\hr-r_0)s(\hg-t)]$ for every direction $s\in\mathcal T$ (Prop.~\ref{prop:cate-orth}), which no finite $f$ removes. Exact balance on many features is fragile, so each gap need only be within its standard error $\widehat\delta_k$ (Sec.~\ref{sec:ratio}).

\section{Experiments}
\label{sec:experiments}

\subsection{Designs}
\label{sec:exp-designs}
\newlength\dfpw\setlength\dfpw{0.19\textwidth}% panel width
\newlength\dfgap\setlength\dfgap{\dimexpr(\textwidth-5\dfpw)/4\relax}% gap between columns
\newsavebox\dfleft
\newlength\dfH\newlength\dfhL\newlength\dfyS\newlength\dfyF\newlength\dfxS% Fig. 4 ATE/CATE outlines
\newcommand{\lgs}{0.6}% legend scale (legend text matches the panel text)
\newcommand{\pnl}[1]{\includegraphics[width=\dfpw]{#1}}
\newcommand{\lgc}[5]{\includegraphics[scale=\lgs,trim=#2 #3 #4 #5,clip]{#1}}% file, trim left bottom right top

\begin{figure*}[tp]
\sbox\dfleft{\begin{minipage}[b]{\dimexpr4\dfpw+3\dfgap\relax}\offinterlineskip
\pnl{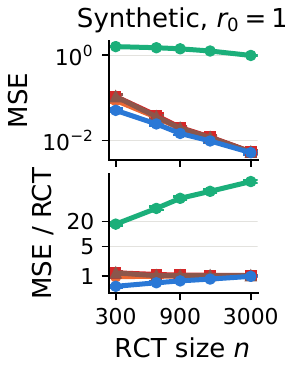}\hfill\pnl{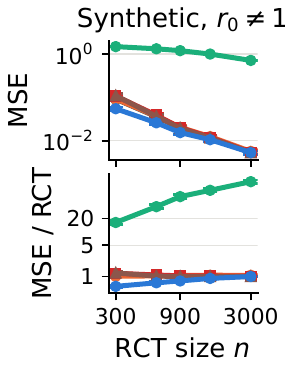}\hfill\pnl{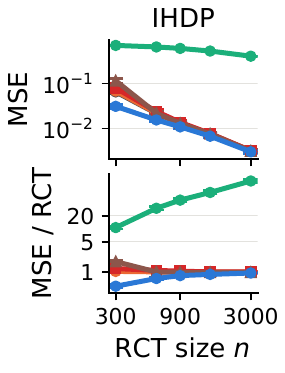}\hfill\pnl{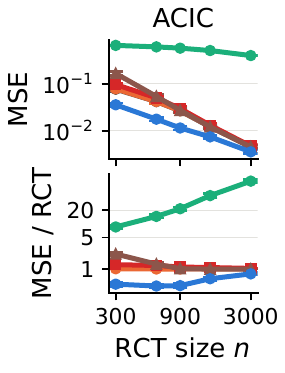}\par\vspace{3pt}
\hspace*{0.03\linewidth}\lgc{s5_legend_cate_full}{0}{0}{378.8}{0}\hfill\lgc{s5_legend_cate_full}{95}{0}{271.8}{0}\hfill\lgc{s5_legend_cate_full}{202.3}{0}{171.8}{0}\hfill\lgc{s5_legend_cate_full}{302.4}{0}{106.3}{0}\hfill\lgc{s5_legend_cate_full}{367.7}{0}{0}{0}\hspace*{0.03\linewidth}\par\vspace{3pt}
\pnl{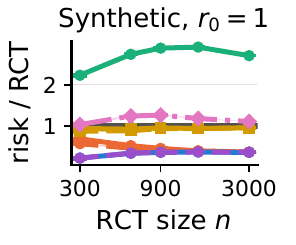}\hfill\pnl{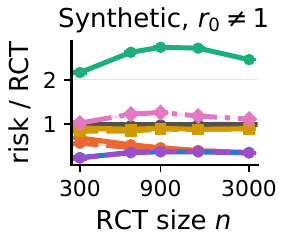}\hfill\pnl{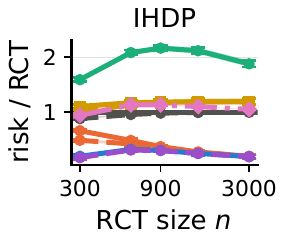}\hfill\pnl{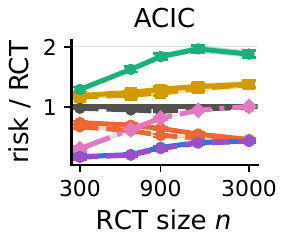}
\end{minipage}}%
\setlength\dfH{\ht\dfleft}\setlength\dfhL{35.2956pt}\setlength\dfhL{\lgs\dfhL}%
\setlength\dfyF{\dimexpr0.827645\dfpw+1.5pt\relax}\setlength\dfyS{\dimexpr\dfyF+3pt+\dfhL\relax}\setlength\dfxS{\dimexpr4\dfpw+3.5\dfgap\relax}%
\noindent\usebox\dfleft\kern\dimexpr\dfgap-0.5pt\relax%
\begin{minipage}[b][\dimexpr\ht\dfleft+\dp\dfleft\relax][s]{\dfpw}\offinterlineskip
\vspace*{0pt}\hspace*{0.12\dfpw}\parbox{0.88\dfpw}{\offinterlineskip
\lgc{s5_legend_ate}{0}{17.7}{204.6}{0}\par
\lgc{s5_legend_ate}{0}{0}{204.6}{17.6}\par
\lgc{s5_legend_ate}{93.46}{17.7}{103.6}{0}\par
\lgc{s5_legend_ate}{93.46}{0}{103.6}{17.6}\par
\lgc{s5_legend_ate}{194.5}{17.7}{0}{0}\par
\lgc{s5_legend_ate}{194.5}{0}{0}{17.6}}\par
\vfill
\pnl{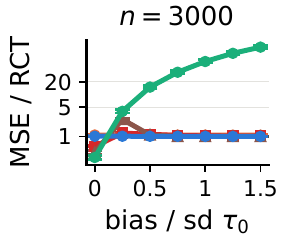}\par\vspace{2.2pt}
\pnl{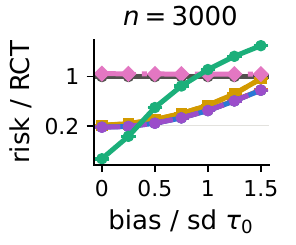}
\end{minipage}%
\begin{tikzpicture}[overlay,line width=0.4pt,xshift={-\linewidth+0.5pt}]%
\draw[black!60] (0pt,\dfH) -- (\linewidth,\dfH) -- (\linewidth,\dfyF+0.75pt) -- (\dfxS+0.75pt,\dfyF+0.75pt) -- (\dfxS+0.75pt,\dfyS+0.75pt) -- (0pt,\dfyS+0.75pt) -- cycle;%
\draw[black!60] (0pt,\dfyS-0.75pt) -- (\dfxS-0.75pt,\dfyS-0.75pt) -- (\dfxS-0.75pt,\dfyF-0.75pt) -- (\linewidth,\dfyF-0.75pt) -- (\linewidth,0pt) -- (0pt,0pt) -- cycle;%
\node[fill=white,inner xsep=1.5pt,inner ysep=0.5pt,anchor=west,font=\sffamily\bfseries\small,text=black!70,yshift=2pt] at (5pt,\dfH) {ATE};%
\node[fill=white,inner xsep=1.5pt,inner ysep=0.5pt,anchor=west,font=\sffamily\bfseries\small,text=black!70,yshift=2pt] at (5pt,\dfyS) {CATE};%
\end{tikzpicture}%
\caption{ATE MSE and its ratio to trial-only AIPW (top); CATE risk ratio to trial-only DR-learner (bottom); 95\% bootstrap intervals. Columns 1--4: versus $n$; synthetic with $r_0=1$, then synthetic, IHDP, and ACIC~2016 under a covariate shift. Column 5: shifted synthetic at $n=3{,}000$, versus confounding bias${}/\mathrm{sd}_R\,\tau_0(X)$.}
\label{fig:results}
\end{figure*}
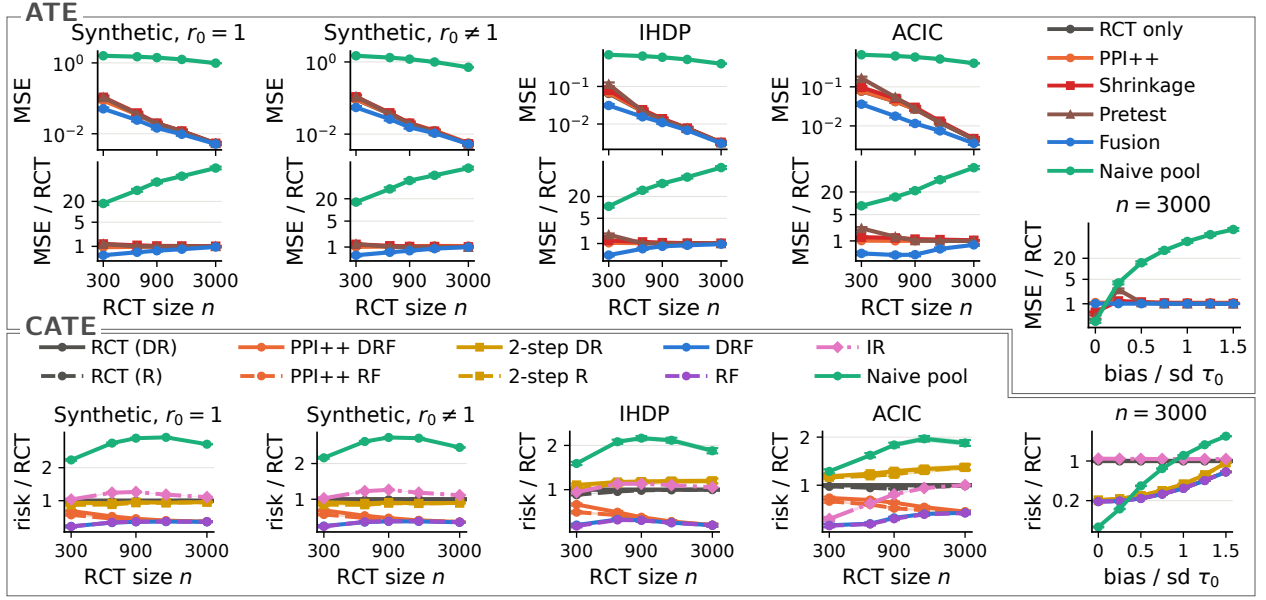

We use one synthetic design and two semi-synthetic benchmarks, IHDP \citep{hill2011bart} and ACIC~2016 \citep{dorie2019acic}. In each design, the trial randomizes the treatment, whereas an unmeasured confounder $U$ affects both the treatment and the outcome in the OBS. The OBS bias is large: in the synthetic design, its root mean square is 1.5 times the standard deviation of $\tau_0(X)$ in the trial. The benchmarks keep their real covariates and noiseless response surfaces, and we simulate the treatments and outcomes. The OBS covariate law either equals the trial law ($r_0=1$) or is tilted away from it, which lowers the effective OBS sample fraction $1/\E_O[r_0^2]$ to 0.70 (weak shift) or 0.40 (strong shift). Appendix~\ref{app:exp-details} gives the full designs.

\subsection{Estimators and Metrics}
\label{sec:exp-estimators}
For the ATE, we compare AIPWF (Fusion) with trial-only AIPW \citep{robins1994estimation}; PPI\textsuperscript{++} \citep{angelopoulos2026ppipp}, which uses the PPI correction alone ($\lambda=0$); Shrinkage \citep[after][]{rosenman2023shrinkage} and Pretest, which combine the trial-only ATE with the ATE of a naive pool of all rows; and the naive pool itself. For the CATE, we compare DRF and RF with the trial-only DR- and R-learners (RCT), their PPI\textsuperscript{++} versions ($\lambda=0$), the 2-step learner \citep[after][]{kallus2018removing}, the integrative $R$-learner (IR) of \citet{wu2022integrative}, and the naive pool. The trial has $n$ between 300 and 3{,}000 units, the OBS has $N=15{,}000$ units, and every split-based method is cross-fitted over three trial blocks. Each setting has 1{,}000 replications. We report the ATE mean squared error (MSE) and the CATE risk $\E_R\{\hat\tau(X)-\tau_0(X)\}^2$ relative to the trial-only AIPW estimator and DR-learner, so a ratio below 1 means that the OBS help. Appendix~\ref{app:exp-details} gives the nuisance models, the tuning rules, and the implementation of each baseline.

\begin{table}[tp]
\caption{ATE mean squared error relative to trial-only AIPW, by trial size $n$ and covariate law of the OBS: common ($r_0{=}1$), weak shift (effective sample fraction 0.70), and strong shift (0.40, as in Fig.~\ref{fig:results}). Our methods are shaded, and the smallest value in each column and trial size is in bold.}
\label{tab:app-ate-ratio}
\begin{center}\footnotesize
\begin{tabular}{@{}llccccccccc@{}}
\toprule
& & \multicolumn{3}{c}{Synthetic} & \multicolumn{3}{c}{IHDP} & \multicolumn{3}{c}{ACIC~2016}\\
\cmidrule(lr){3-5}\cmidrule(lr){6-8}\cmidrule(lr){9-11}
$n$ & Method & $r_0{=}1$ & weak & strong & $r_0{=}1$ & weak & strong & $r_0{=}1$ & weak & strong\\
\midrule
300 & PPI\textsuperscript{++} & 0.95 & 1.01 & 1.01 & 1.00 & 1.04 & 1.05 & 1.00 & 1.02 & 1.02 \\
\rowcolor{gray!12} & Fusion & \textbf{0.55} & \textbf{0.60} & \textbf{0.60} & \textbf{0.46} & \textbf{0.47} & \textbf{0.48} & \textbf{0.45} & \textbf{0.46} & \textbf{0.47} \\
 & Shrinkage & 1.17 & 1.18 & 1.18 & 1.21 & 1.21 & 1.22 & 1.25 & 1.26 & 1.26 \\
 & Pretest & 1.17 & 1.16 & 1.20 & 1.71 & 1.70 & 1.76 & 2.07 & 2.09 & 2.16 \\
 & Naive pool & 17.41 & 16.92 & 16.24 & 11.05 & 10.92 & 10.69 & 9.16 & 8.93 & 8.56 \\
\addlinespace
900 & PPI\textsuperscript{++} & 0.97 & 1.04 & 1.05 & 1.00 & 1.02 & 1.02 & 1.00 & 1.01 & 1.01 \\
\rowcolor{gray!12} & Fusion & \textbf{0.75} & \textbf{0.79} & \textbf{0.80} & \textbf{0.82} & \textbf{0.84} & \textbf{0.84} & \textbf{0.42} & \textbf{0.42} & \textbf{0.43} \\
 & Shrinkage & 1.03 & 1.03 & 1.03 & 1.05 & 1.05 & 1.06 & 1.10 & 1.11 & 1.12 \\
 & Pretest & 1.00 & 1.00 & 1.00 & 1.00 & 1.00 & 1.00 & 1.03 & 1.04 & 1.02 \\
 & Naive pool & 72.74 & 67.82 & 61.71 & 51.00 & 48.97 & 45.79 & 24.50 & 23.25 & 21.62 \\
\addlinespace
3{,}000 & PPI\textsuperscript{++} & 0.99 & 1.06 & 1.06 & 0.97 & 1.04 & 1.00 & 1.00 & 1.02 & 1.01 \\
\rowcolor{gray!12} & Fusion & \textbf{0.96} & \textbf{1.00} & \textbf{1.00} & \textbf{0.92} & \textbf{0.98} & \textbf{0.96} & \textbf{0.77} & \textbf{0.79} & \textbf{0.79} \\
 & Shrinkage & 1.01 & 1.01 & 1.01 & 1.02 & 1.02 & 1.02 & 1.03 & 1.03 & 1.04 \\
 & Pretest & 1.00 & \textbf{1.00} & \textbf{1.00} & 1.00 & 1.00 & 1.00 & 1.00 & 1.00 & 1.00 \\
 & Naive pool & 184.50 & 160.01 & 133.47 & 163.58 & 148.30 & 126.35 & 113.49 & 101.57 & 88.22 \\
\bottomrule
\end{tabular}
\end{center}
\end{table}

\begin{table}[tp]
\caption{CATE risk relative to the trial-only DR-learner; otherwise as Table~\ref{tab:app-ate-ratio}. RCT (R) is the trial-only R-learner, and IR the integrative $R$-learner. Our methods are shaded, and the smallest value in each column and trial size is in bold.}
\label{tab:app-cate-ratio}
\begin{center}\footnotesize
\begin{tabular}{@{}llccccccccc@{}}
\toprule
& & \multicolumn{3}{c}{Synthetic} & \multicolumn{3}{c}{IHDP} & \multicolumn{3}{c}{ACIC~2016}\\
\cmidrule(lr){3-5}\cmidrule(lr){6-8}\cmidrule(lr){9-11}
$n$ & Method & $r_0{=}1$ & weak & strong & $r_0{=}1$ & weak & strong & $r_0{=}1$ & weak & strong\\
\midrule
\rowcolor{gray!12}300 & DRF & \textbf{0.58} & 0.59 & \textbf{0.59} & \textbf{0.56} & 0.57 & \textbf{0.57} & 0.54 & 0.54 & \textbf{0.56} \\
\rowcolor{gray!12} & RF & \textbf{0.58} & \textbf{0.58} & \textbf{0.59} & \textbf{0.56} & \textbf{0.56} & \textbf{0.57} & \textbf{0.53} & \textbf{0.53} & \textbf{0.56} \\
 & PPI\textsuperscript{++} DRF & 0.81 & 0.80 & 0.81 & 0.79 & 0.79 & 0.79 & 0.82 & 0.80 & 0.83 \\
 & PPI\textsuperscript{++} RF & 0.74 & 0.74 & 0.74 & 0.70 & 0.70 & 0.70 & 0.78 & 0.76 & 0.79 \\
 & 2-step DR & 0.96 & 0.95 & 0.94 & 1.06 & 1.07 & 1.08 & 1.06 & 1.10 & 1.14 \\
 & 2-step R & 0.92 & 0.91 & 0.90 & 0.99 & 0.99 & 1.01 & 1.04 & 1.08 & 1.12 \\
 & IR & 1.02 & 1.02 & 1.02 & 0.94 & 0.96 & 0.96 & 0.61 & 0.61 & 0.61 \\
 & RCT (R) & 0.96 & 0.96 & 0.96 & 0.92 & 0.92 & 0.92 & 0.98 & 0.98 & 0.98 \\
 & Naive pool & 2.35 & 2.30 & 2.24 & 1.52 & 1.52 & 1.51 & 1.18 & 1.21 & 1.22 \\
\addlinespace
\rowcolor{gray!12}900 & DRF & \textbf{0.64} & \textbf{0.65} & \textbf{0.65} & \textbf{0.60} & \textbf{0.61} & \textbf{0.62} & \textbf{0.58} & 0.61 & \textbf{0.62} \\
\rowcolor{gray!12} & RF & \textbf{0.64} & \textbf{0.65} & \textbf{0.65} & \textbf{0.60} & \textbf{0.61} & \textbf{0.62} & \textbf{0.58} & \textbf{0.60} & \textbf{0.62} \\
 & PPI\textsuperscript{++} DRF & 0.69 & 0.69 & 0.69 & 0.63 & 0.64 & 0.65 & 0.74 & 0.76 & 0.78 \\
 & PPI\textsuperscript{++} RF & 0.67 & 0.67 & 0.68 & 0.62 & 0.63 & 0.64 & 0.68 & 0.70 & 0.72 \\
 & 2-step DR & 0.97 & 0.95 & 0.94 & 1.11 & 1.12 & 1.14 & 1.10 & 1.16 & 1.22 \\
 & 2-step R & 0.96 & 0.94 & 0.93 & 1.10 & 1.11 & 1.13 & 1.05 & 1.11 & 1.17 \\
 & IR & 1.20 & 1.20 & 1.20 & 1.08 & 1.09 & 1.10 & 0.87 & 0.87 & 0.88 \\
 & RCT (R) & 0.99 & 0.99 & 0.99 & 0.99 & 0.99 & 0.99 & 0.95 & 0.95 & 0.95 \\
 & Naive pool & 3.72 & 3.54 & 3.32 & 2.38 & 2.33 & 2.24 & 1.80 & 1.80 & 1.79 \\
\addlinespace
\rowcolor{gray!12}3{,}000 & DRF & \textbf{0.64} & \textbf{0.64} & \textbf{0.64} & \textbf{0.55} & \textbf{0.56} & \textbf{0.57} & \textbf{0.62} & \textbf{0.65} & \textbf{0.67} \\
\rowcolor{gray!12} & RF & \textbf{0.64} & \textbf{0.64} & \textbf{0.64} & \textbf{0.55} & \textbf{0.56} & \textbf{0.57} & \textbf{0.62} & \textbf{0.65} & \textbf{0.67} \\
 & PPI\textsuperscript{++} DRF & \textbf{0.64} & \textbf{0.64} & \textbf{0.64} & \textbf{0.55} & \textbf{0.56} & \textbf{0.57} & 0.64 & 0.66 & 0.68 \\
 & PPI\textsuperscript{++} RF & \textbf{0.64} & \textbf{0.64} & \textbf{0.64} & \textbf{0.55} & \textbf{0.56} & \textbf{0.57} & 0.63 & 0.66 & 0.68 \\
 & 2-step DR & 0.97 & 0.94 & 0.93 & 1.11 & 1.12 & 1.15 & 1.15 & 1.22 & 1.30 \\
 & 2-step R & 0.97 & 0.94 & 0.93 & 1.11 & 1.12 & 1.15 & 1.14 & 1.22 & 1.29 \\
 & IR & 1.08 & 1.08 & 1.08 & 1.02 & 1.03 & 1.04 & 1.00 & 1.00 & 1.00 \\
 & RCT (R) & 1.00 & 1.00 & 1.00 & 1.00 & 1.00 & 1.00 & 0.99 & 0.99 & 0.99 \\
 & Naive pool & 3.28 & 3.02 & 2.73 & 2.13 & 2.00 & 1.84 & 1.95 & 1.90 & 1.84 \\
\bottomrule
\end{tabular}
\end{center}
\end{table}

\subsection{Main Results}
\label{sec:exp-results}
Fig.~\ref{fig:results} shows each error relative to the trial-only estimator on a log scale, so a curve below 1 improves on the trial alone. Fusion has the smallest MSE in every design, covariate law, and trial size (columns 1--4). It is also robust to the size of the confounding (column 5): its error stays flat as the bias grows. DRF and RF have the smallest risk in every design, covariate law, and trial size. They stay below the trial-only risk at every bias. Tables~\ref{tab:app-ate-ratio} and~\ref{tab:app-cate-ratio} give the numbers.

\paragraph{ATE.} Across the 45 settings (three designs, three covariate laws, five trial sizes), the MSE of fusion is 0.40 to 1.00 of the trial-only MSE (Table~\ref{tab:app-ate-ratio}). The gain is largest for small trials (0.47 to 0.60 at $n=300$ in Fig.~\ref{fig:results}) and shrinks as the trial grows. PPI\textsuperscript{++} stays near the trial-only MSE, at 0.95 to 1.07, because the PPI correction can remove at most the share of effect heterogeneity in the score variance (Remark~\ref{rem:channels}), which is 0.02 to 0.10 in these designs. Shrinkage and Pretest stay at or above the trial-only MSE, and the naive pool is 8.6 to 184 times worse, because its confounding bias does not shrink with $n$. Appendix~\ref{app:prog} compares fusion with two prognostic-score adjustments \citep{schuler2022prognostic}, which put the same OBS prediction into a trial regression: they are slightly better than fusion on IHDP at $n=300$ under a shift but up to 1.5 times worse than the trial alone on the synthetic design, whereas fusion never exceeds the trial-only MSE. The Wald intervals of fusion have coverage 0.94 to 0.96 and are 63\% to 99\% as wide as the trial-only ones (Tables~\ref{tab:ate-inference} and~\ref{tab:ate-inference-b}).

\paragraph{CATE.} DRF and RF have 0.53 to 0.67 of the trial-only risk in all 45 settings and differ from each other by at most 0.01 (Table~\ref{tab:app-cate-ratio}). Their PPI\textsuperscript{++} versions catch up only at $n=3{,}000$. Fitting the trial loss alone on the same sieve, which uses $\hg$ only as a basis function, shows that at small $n$ most of the gain comes from the fusion loss and at large $n$ from $\hg$ as a basis function (Appendix~\ref{sec:exp-ablation}). IR improves on the trial only on ACIC and on IHDP at $n=300$, and 2-step stays at 0.90 to 1.30 in the settings of Fig.~\ref{fig:results}. Both model the OBS bias, so the OBS help them only when the bias follows that model.

\paragraph{Further results.} Appendix~\ref{sec:exp-sweeps} repeats the bias sweep of Fig.~\ref{fig:results} at other trial sizes and varies the OBS size $N$. Appendix~\ref{sec:exp-ablation} separates the two channels $\lambda$ and $\omega$, compares fusion with practitioner's trial-only baselines, and studies a misspecified density ratio, where the calibrated ratio keeps fusion unbiased, and a design with larger effect heterogeneity, where $\omega$ improves on $\lambda$ alone when the covariate laws coincide.

\FloatBarrier
\section{Conclusion}
\label{sec:conclusion}
For the ATE, AIPWF has closed-form optimal weights (Thm.~\ref{thm:mean-var}, Cor.~\ref{cor:oracle}) and valid Wald intervals with a calibrated ratio (Thm.~\ref{thm:wald}). For the CATE, DRF and RF keep the trial's CATE risk as their target (Thm.~\ref{prop:cate}), validation selects the weights with a guarantee relative to the RCT-only learner (Thm.~\ref{thm:validate}), and for linear sieves, calibration on finitely many moments controls the ratio error in the risk (Thm.~\ref{thm:sieve-risk}, Prop.~\ref{prop:sieve-balance}). Fusion has the lowest ATE error, and DRF and RF the lowest CATE risk, among the estimators of Section~\ref{sec:experiments} in every design and trial size. Prognostic-score adjustment is slightly better than fusion on one benchmark but up to 1.5 times worse than the trial alone on another, whereas fusion never exceeds the trial-only error (Appendix~\ref{app:prog}).

\endgroup

%%%%%%%%%%%%%%%%%%%%%%%%%%%%%%%%%%%%%%%%%%%%%%%%%%%%%%%
% References
%%%%%%%%%%%%%%%%%%%%%%%%%%%%%%%%%%%%%%%%%%%%%%%%%%%%%%%
\bibliographystyle{abbrvnat}
\bibliography{reference}

%%%%%%%%%%%%%%%%%%%%%%%%%%%%%%%%%%%%%%%%%%%%%%%%%%%%%%%
% Appendix
%%%%%%%%%%%%%%%%%%%%%%%%%%%%%%%%%%%%%%%%%%%%%%%%%%%%%%%
\clearpage
\appendix

\vtop{
    \centering
    \vspace{0cm}
    \hspace{0cm}
    \begin{tikzpicture}
        \centering
        \node (box){
          \begin{minipage}[t!]{0.9\textwidth}
            \centering
            {\Large Supplement of \Paste{title} }
          \end{minipage}
        };
    \end{tikzpicture}
    \vspace{0.5cm}
}
\suppressfloats[t]
\begingroup
\allowdisplaybreaks
\sloppy
\section{Proofs}
\label{app:proofs}

\begin{table}[tp]
\centering\small
\caption{Section-specific notation. Notation used throughout the paper is in Tables~\ref{tab:notation} and~\ref{tab:notation-cate}.}
\label{tab:notation-sec}
\begin{tabular}{@{}lll@{}}
\toprule
Symbol & Meaning & Where\\
\midrule
\multicolumn{3}{@{}l}{\textit{Sec.~\ref{sec:ate}: ATE}}\\
$A,B,C,D$ & variance coefficients & Thm.~\ref{thm:mean-var}\\
$\lambda^\star,\ \omega^\star$ & oracle coefficients & Thm.~\ref{thm:mean-var}\\
$\widetilde V(\lambda,\omega)$ & scaled variance $n\Var\{\htheta_{r_0}(\lambda,\omega)\}$ & Cor.~\ref{cor:oracle}\\
\ $\epsilon_A,\ \epsilon_B$ & coefficient thresholds & Eq.~\eqref{eq:coef-select}\\
$\hA,\hB,\hC,\hD$ & tuning-sample estimates of $A,\dots,D$ & Sec.~\ref{sec:ate-alg}\\
$\psi_R\triangleq Z^{\hlam}-\homg\hg$, $\psi_O\triangleq\homg\hr\hg$, $\hV$ & RCT and OBS terms of $\htheta$; variance estimate & Algorithm~\ref{alg:honest-ate}\\
$V_n$ & variance of $\htheta$ & Thm.~\ref{thm:wald}\\
$\widehat r_{\mathrm{RZ}},\ \widehat r_{\mathrm{CLS}},\ \widehat r_{\mathrm{BAL}}$ & Riesz, classifier, and balancing ratio estimates & Sec.~\ref{sec:ratio}\\
$f,\ \widehat\delta_{k}$ & balance features and tolerances & Def.~\ref{def:bal}\\
\midrule
\multicolumn{3}{@{}l}{\textit{Sec.~\ref{sec:cate}: CATE}}\\
$\mathcal R_0(t)$ & CATE risk & Sec.~\ref{sec:cate-loss}\\
$\mathcal R_{\mdl},\ \widehat{\mathcal R}_{\mdl}$ & RCT loss and fusion loss of learner $\mdl\in\{\mathrm{DRF},\mathrm{RF}\}$ & Sec.~\ref{sec:cate-loss}, Def.~\ref{def:fusion-losses}\\
$\kappa_{\mdl},\ Z_{\mdl}^\lambda,\ m_\lambda$ & weight, pseudo-outcome, marginal regression & Def.~\ref{def:fusion-losses}\\
$Q_{\mdl}(\lambda)$ & noise floor & Thm.~\ref{prop:cate}\\
$\Psi_{\mdl}$ & population learning equation & Prop.~\ref{prop:cate-orth}\\
$\mathcal G,\ \widehat{\mathcal R}_{\mathrm{val}}$ & candidate grid and validation score & Def.~\ref{def:validation}\\
$\varepsilon_{\mathrm{eval}}(\alpha),\ \varrho_{\mdl}$ & evaluation error and transport error & Thm.~\ref{thm:validate}\\
$\zeta,\ \mathcal T_{\zeta},\ \widehat\beta_{\mdl}$ & sieve features, class, and coefficients & Def.~\ref{def:sieve}\\
$\tau^\star_p,\ \mathcal J_{\mdl}$ & projection of $\tau_0$ and first-order variance & Thm.~\ref{thm:sieve-risk}\\
\bottomrule
\end{tabular}
\end{table}

\paragraph{Orthogonality.}
Sec.~\ref{sec:cate} states which fitted nuisances enter the first-order condition of a loss. The following two definitions fix the language.
\begin{definition}[Gateaux derivative {\citep[Chapter~20]{vandervaart1998asymptotic}}]
\label{def:gateaux}
Let $F$ be a functional of a function-valued argument $\eta$, and let $h$ be a direction such that $\eta+\varepsilon h$ stays in the domain of $F$ for all sufficiently small $|\varepsilon|$. Whenever the limit exists, the Gateaux derivative of $F$ at $\eta$ in the direction $h$ is
\begin{align}
	D_\eta F(\eta)[h]&\triangleq\left.\frac{d}{d\varepsilon}F(\eta+\varepsilon h)\right|_{\varepsilon=0}
	\label{eq:gateaux-derivative}
\end{align}
\end{definition}
\begin{definition}[Learning equation and orthogonality {\citep{chernozhukov2018dml,foster2023orthogonal}}]
\label{def:orthogonality}
Let $L(t;\eta)$ be a population loss with a target argument $t$ in a linear class and a nuisance argument $\eta$, and let $\eta_0$ denote the true nuisance. The learning equation of $L$ is
\begin{align}
\Psi(t;\eta)[s]&\triangleq-\frac12D_t L(t;\eta)[s]
\label{eq:learning-equation-generic}
\end{align}
for target directions $s$ in the class. The loss is orthogonal to $\eta$ if $D_\eta\Psi(t^\star;\eta_0)[h,s]=0$ for every admissible nuisance direction $h$ and every target direction $s$.
\end{definition}
When $L$ is convex in $t$, a function $t^\star$ minimizes $L(\cdot\,;\eta_0)$ over the class exactly when $\Psi(t^\star;\eta_0)[s]=0$ for every $s$ in the class, which is why $\Psi$ is called the learning equation.

\paragraph{Lemma~\ref{lem:aipw}.}
Condition on $X$. Randomization and consistency give
$\E_R[(A/e(X))\{Y-q(X,1)\}\mid X]=\E_R[Y(1)-q(X,1)\mid X]$ and the analogous identity for $A=0$. Substituting into Eq.~\eqref{eq:aipw} cancels $q$ and leaves $\tau_0(X)$. The same argument with $q=\hmu_\lambda$ yields $\E_R[Z^\lambda\mid X]=\tau_0(X)$ and $\E_R[\Delta\mid X]=0$.

\paragraph{Thm.~\ref{thm:mean-var}.}
RCT-validity gives $\E_R[Z^\lambda]=\theta_0$. Exact transport gives $\E_O[r_0\hg]=\E_R[\hg]$. These two identities prove unbiasedness of $\htheta_{r_0}$. Replacing $r_0$ by $\widetilde r$ leaves the drift Eq.~\eqref{eq:drift}. Centering gives
$\htheta_{r_0}-\theta_0=(\mathbb P_{R,n}-P_R)(Z^\lambda-\omega\hg)+(\mathbb P_{O,N}-P_O)(\omega r_0\hg)$.
Independence of the evaluation blocks and the usual variance of an empirical mean give $\Var\{\htheta_{r_0}(\lambda,\omega)\}=\frac1n\Var_R(Z^\lambda-\omega\hg)+\frac{\omega^2}{N}\Var_O(r_0\hg)$; substituting $Z^\lambda=Z^0+\lambda\Delta$ and $\Cov_R(\Delta,\hg)=0$ gives Eq.~\eqref{eq:j}.

\paragraph{Thm.~\ref{thm:mean-var} (minimizer) and Cor.~\ref{cor:oracle}.}
Expand $\Var_R(Z^0+\lambda\Delta-\omega\hg)$ and use $\Cov_R(\Delta,\hg)=0$. The resulting quadratic is separated. The unique critical point when $A,B>0$ is $\lambda^\star=-C/A$, $\omega^\star=D/B$. Completing the square yields the stated minimum. The identity $D=\Cov_R(\tau_0(X),\hg(X))$ follows from $\E_R[Z^0\mid X]=\tau_0(X)$.  Also, $C=\E_R[\Cov_R(Z^0,\Delta\mid X)]+\Cov_R\{\E_R[Z^0\mid X],\E_R[\Delta\mid X]\}$ and the second term is zero. If $\hmu_R$ equals the trial regression, $Z^0-\tau_0(X)=\{A/e(X)\}\varepsilon(1)-\{(1-A)/(1-e(X))\}\varepsilon(0)$ with $\varepsilon(a)\triangleq Y-\mu_R(X,a)$ and $\E_R[\varepsilon(a)\mid A=a,X]=0$, while $\Delta$ is a function of $(A,X)$, so $C=0$.

\paragraph{Prop.~\ref{prop:drift}.}
Under Assumption~\ref{ass:honest-separation}, the evaluation samples are independent of $\hmu_R$ and $\hmu_O$, so $Z^\lambda$ and $\hg$ are fixed functions given the fits, and $\widetilde r$ is fixed. Taking the mean of each sample average in Eq.~\eqref{eq:ate} gives $\E[\htheta_{\widetilde r}(\lambda,\omega)]=\E_R[Z^\lambda]+\omega\{\E_O[\widetilde r\hg]-\E_R[\hg]\}$. Lemma~\ref{lem:aipw} with $q=\hmu_\lambda$ gives $\E_R[Z^\lambda]=\E_R[\tau_0(X)]=\theta_0$, and the braces equal $\mathtt{imb}_{\hg}(\widetilde r)$ by definition, which proves Eq.~\eqref{eq:drift}. The second form of the imbalance, $\E_O[\widetilde rf]-\E_R[f]=\E_O[(\widetilde r-r_0)f]$, is Prop.~\ref{prop:ratio-char}(i) with $h=f$.

\paragraph{Prop.~\ref{prop:ratio-char}.}
(i) Under Assumption~\ref{ass:exact-transport}, $r_0=dP_R^X/dP_O^X$ exists, and the Radon--Nikodym theorem gives $\E_O[r_0\mathbf 1_B]=P_R^X(B)=\E_R[\mathbf 1_B]$ for every measurable $B\subseteq\mathcal X$. Linearity extends this identity to simple functions, and monotone convergence extends it to every nonnegative measurable $h$. For $h\in L^1(P_R^X)$, applying it to $h^+$ and $h^-$ shows that both parts are finite, so $r_0h\in L^1(P_O^X)$ and $\E_O[r_0h]=\E_R[h]$. For uniqueness, let $w\in L^1(P_O^X)$ satisfy $\E_O[wh]=\E_R[h]$ for every bounded $h$. With $B\triangleq\{w>r_0\}$ and $h=\mathbf 1_B$, the hypothesis and the identity just shown give $\E_O[(w-r_0)\mathbf 1_B]=P_R^X(B)-P_R^X(B)=0$. The integrand is nonnegative and is positive on $B$, so $P_O^X(B)=0$. The same argument with $B=\{w<r_0\}$ gives $w=r_0$ $P_O^X$-a.s.

(ii) Let $a\in L^2(P_O^X)$. By part~(i) applied to $|a|$ and the Cauchy--Schwarz inequality, $\E_R|a|=\E_O[r_0|a|]\le\{\E_O[r_0^2]\,\E_O[a^2]\}^{1/2}<\infty$, so part~(i) applies to $h=a$ and gives $\E_R[a]=\E_O[r_0a]$. Hence $\E_O[a^2]-2\E_R[a]=\E_O[(a-r_0)^2]-\E_O[r_0^2]$. The last term does not depend on $a$, and the first is zero only when $a=r_0$ $P_O^X$-a.s., so $r_0$ is the unique minimizer over $L^2(P_O^X)$.

\paragraph{Thm.~\ref{thm:wald}.}
Condition on the fitted and selected objects. Unbiased sample variances on the two independent evaluation blocks prove $\E[\hV]=\Var(\htheta_r)$ for any fixed $r$. The studentized limit is a two-sample triangular-array CLT under the $(2+\eta)$-moment (Lyapunov) condition of Assumption~\ref{ass:wald}, plus Slutsky using $\hV/V_n\to_p 1$. Variance unbiasedness is not a CLT.

\paragraph{Estimated ratio.}
Write $\hr\hg=r_0\hg+(\hr-r_0)\hg$ on the OBS tuning sample to see that $\hr$ changes $\hB$ only. On evaluation, $\htheta_{\hr}-\htheta_{r_0}=\homg\,\mathbb P_{O,N}\{(\hr-r_0)\hg\}$. The conditional mean is $\theta_0+\homg\{\E_O[\hr\hg]-\E_R[\hg]\}=\theta_0+\homg\,\mathtt{imb}_{\hg}(\hr)$, and $\E_R[\hg]=\E_O[r_0\hg]$ gives $\mathtt{imb}_{\hg}(\hr)=\E_O[(\hr-r_0)\hg]$.

\noindent\emph{Calibration noise (Thm.~\ref{thm:wald}(ii)).} Under Assumption~\ref{ass:wald}, calibration with $\hg$ among the features gives $\mathbb P_O^{\mathrm{nuis}}(\hr\hg)=\mathbb P_R^{\mathrm{nuis}}(\hg)$, so $\mathtt{imb}_{\hg}(\hr)=(\mathbb P_R^{\mathrm{nuis}}-\E_R)\hg-(\mathbb P_O^{\mathrm{nuis}}-\E_O)(\hr\hg)$. Since $\hg$ is fitted on $D_O^{\mathrm{nuis}}$, the first term is, given $D_O^{\mathrm{nuis}}$, an average of $n_{\mathrm{nuis}}$ independent centered terms, and $\hg\to\bar g$ replaces it by $(\mathbb P_R^{\mathrm{nuis}}-\E_R)\bar g$ up to $o_p(n_{\mathrm{nuis}}^{-1/2})$. The second term is linearized by assumption. Because $\mathtt{imb}_{\hg}(\hr)=O_p(n^{-1/2})$, $(\homg-\bar\omega)\mathtt{imb}_{\hg}(\hr)=o_p(n^{-1/2})$. Hence $\htheta-\theta_0$ equals, up to $o_p(n^{-1/2})$, a sum of four centered averages over the independent samples $D_R^{\mathrm{eval}}$, $D_O^{\mathrm{eval}}$, $D_R^{\mathrm{nuis}}$, and $D_O^{\mathrm{nuis}}$, with total variance $V_n+V_{\mathrm{cal}}$ to first order. The Lyapunov central limit theorem and Slutsky's lemma give the coverage. Under Assumption~\ref{ass:common-marginal} with $\hr\equiv1$, $\mathtt{imb}_{\hg}(\hr)=0$ exactly and $\widehat V_{\mathrm{cal}}=0$.

\paragraph{Thm.~\ref{prop:cate}.}
The cross term in $\E_R[\{(Z^\lambda-\tau_0)+(\tau_0-t)\}^2]$ vanishes by $\E_R[Z^\lambda-\tau_0\mid X]=0$, so $\mathcal R_{\mathrm{DRF}}=Q_{\mathrm{DRF}}(\lambda)+\mathcal R_0(t)$. For the $R$ loss, write $Y-m_\lambda-(A-e)t=U+(A-e)(\tau_0-t)$ with $U\triangleq Y-m_\lambda-(A-e)\tau_0$; randomization gives $\E_R[U(A-e)\mid X]=0$. For the $\omega$ terms, $\E_R[\kappa_{\mdl}\mid X]=1$ and $\E_O[r_0h]=\E_R[h]$ give $\E\{\mathbb P_O^{\mathrm{tune}}[r(\hg-t)^2]-\mathbb P_R^{\mathrm{tune}}[\kappa_{\mdl}(\hg-t)^2]\}=\E_O[(r-r_0)(\hg-t)^2]$, which is zero at $r=r_0$.

\paragraph{Prop.~\ref{prop:cate-orth}.}
Fix the fitted regressions and $r$. Under Assumption~\ref{ass:honest-separation}, the tuning averages in Eq.~\eqref{eq:fusion-loss} are independent of the fits, so $\E[\widehat{\mathcal R}_{\mdl}(t;\lambda,\omega,r)]=\E_R[\kappa_{\mdl}\{Z^\lambda_{\mdl}-t\}^2]+\omega\E_O[r(\hg-t)^2]-\omega\E_R[\kappa_{\mdl}(\hg-t)^2]$. Along $t+\varepsilon s$ this is a quadratic in $\varepsilon$, and Definition~\ref{def:gateaux} gives
\begin{align*}
\Psi_{\mdl}(t)[s]={}&\E_R[\kappa_{\mdl}\{Z^\lambda_{\mdl}-t\}s]+\omega\E_O[r(\hg-t)s]-\omega\E_R[\kappa_{\mdl}(\hg-t)s].
\end{align*}
Each perturbation below changes $\Psi_{\mdl}(t)[s]$ affinely in $\varepsilon$, so its derivative at $\varepsilon=0$ is the coefficient of $\varepsilon$. Since $e(X)=P_R(A=1\mid X)$, we have $\E_R[A-e\mid X]=0$ and $\E_R[(A-e)^2\mid X]=e(1-e)$, so $\E_R[\kappa_{\mdl}\mid X]=1$ for both learners.

(i) Along $\hmu_R+\varepsilon h$, only $Z^\lambda_{\mdl}$ changes, through $\hmu_\lambda+\varepsilon(1-\lambda)h$. For DRF, the score in Eq.~\eqref{eq:aipw} is affine in $q$, and the coefficient of $\varepsilon$ in $Z^\lambda$ is $(1-\lambda)[h(X,1)\{1-A/e\}-h(X,0)\{1-(1-A)/(1-e)\}]$, whose conditional mean given $X$ is zero. For RF, $\kappa_{\mathrm{RF}}Z^\lambda_{\mathrm{RF}}=(A-e)(Y-m_\lambda)/\{e(1-e)\}$, and $m_\lambda$ changes by $\varepsilon(1-\lambda)\{e\,h(X,1)+(1-e)h(X,0)\}$, so the coefficient of $\varepsilon$ is $-(1-\lambda)(A-e)\{e\,h(X,1)+(1-e)h(X,0)\}/\{e(1-e)\}$, whose conditional mean given $X$ is also zero. Since $s$ is a function of $X$, the derivative is $0$ for every $r$.

(ii) Along $\hmu_O+\varepsilon h$, $\hmu_\lambda$ changes by $\varepsilon\lambda h$ and $\hg$ by $\varepsilon\Delta h$. The first change contributes zero by the argument of part~(i) with $\lambda$ in place of $1-\lambda$. The second contributes $\omega\E_O[r\,\Delta h\,s]-\omega\E_R[\kappa_{\mdl}\Delta h\,s]=\omega\{\E_O[r\,\Delta h\,s]-\E_R[\Delta h\,s]\}$, because $\E_R[\kappa_{\mdl}\mid X]=1$. Prop.~\ref{prop:ratio-char}(i) with $h$ replaced by $\Delta h\,s$ gives $\E_R[\Delta h\,s]=\E_O[r_0\Delta h\,s]$, so the derivative is $\omega\E_O[(r-r_0)\Delta h\,s]$.

(iii) Along $r+\varepsilon h_r$, only the OBS term changes, and the derivative is $\omega\E_O[h_r\,s(\hg-t)]$.

Hence the derivative in (i) is zero for every $r$, the derivative in (ii) is zero for all $h$ and $s$ when $r=r_0$, and the derivative in (iii) is zero for all $h_r$ when $\hg=t$ $P_O^X$-a.s. For the converse statements, let $\omega\neq0$, and let $\mathcal T$ contain the constant function $1$ and a function that is not $P_O^X$-a.s.\ constant. If $r\neq r_0$ on a set of positive $P_O^X$-probability, then $h(x,1)=\mathrm{sign}\{r(x)-r_0(x)\}$, $h(x,0)=0$, and $s=1$ give $\omega\E_O|r-r_0|\neq0$ in (ii). In (iii), take $h_r=\mathbf 1_B-P_O^X(B)$ for measurable $B\subseteq\mathcal X$. A zero derivative for every such $h_r$ gives $\E_O[\{W-\E_O W\}\mathbf 1_B]=0$ for every $B$, with $W\triangleq s(\hg-t)$, so $W$ is $P_O^X$-a.s.\ constant. With $s=1$, $\hg-t=c$ for a constant $c$, and with the non-constant $s$, $c\,s$ is constant only if $c=0$. Thus orthogonality to $r$ forces $\hg=t$ $P_O^X$-a.s.

\paragraph{Thm.~\ref{thm:validate}.}
Condition on fitted objects. The validation score has mean $Q_{\mathrm{DRF}}(0)+\mathcal R_0(\hzeta_{\mdl})+\omega\E_O[(\hr-r_0)(\hg-\hzeta_{\mdl})^2]$. Hoeffding on the $M$ bounded RCT and OBS averages, with a union bound, yields the uniform deviation $\varepsilon_{\mathrm{eval}}(\alpha)$. For $\omega\in[0,1]$, each RCT summand $\{Z^0-\hzeta_{\mdl}\}^2-\omega G$ lies in an interval of length $8B^2$ and each OBS summand $\omega rG$ in an interval of length $4\bar rB^2$. A two-sided Hoeffding bound with failure probability $\alpha/(2M)$ for each of the $2M$ averages gives the RCT term $2B^2\sqrt{2\log(4M/\alpha)}\,2n^{-1/2}$ and the OBS term $2B^2\sqrt{2\log(4M/\alpha)}\,\bar rN^{-1/2}$. The selected pair then lies within $2\varepsilon_{\mathrm{eval}}(\alpha)$ plus twice the largest transport error of the best grid risk, which is at most the $(0,0)$ risk.

\paragraph{Details for Sec.~\ref{sec:cate-sieve}.}
\emph{Best sieve approximation.} Since $\E_R[\zeta(\tau_0-\zeta^\top\beta_p)]=0$, $\mathcal R_0(\zeta^\top\beta)=\E_R[\{\zeta^\top(\beta-\beta_p)+(\tau^\star_p-\tau_0)\}^2]=\|\beta-\beta_p\|^2_{\Gamma_R}+\mathcal R_0(\tau^\star_p)$, which is minimized at $\beta=\beta_p$. Let $\Gamma_R\triangleq\E_R[\zeta\,\zeta^\top]\succ0$ and $\beta_p\triangleq\Gamma_R^{-1}\E_R[\zeta\tau(X)]$, so $\tau^\star_p=\zeta^\top\beta_p$.  Since $\widehat{\mathcal R}_{\mdl}(\zeta^\top\beta;\lambda,\omega,\hr)$ is quadratic in $\beta$ with Hessian $2\widehat H_{\mdl}(\omega)\succeq0$, every minimizer solves the normal equations $\{\widehat H_{\mdl}(\omega)+\rho I\}\beta=\mathbb P_R^{\mathrm{tune}}\{\kappa_{\mdl}\zeta(Z_{\mdl}^\lambda-\omega\hg)\}+\omega\mathbb P_O^{\mathrm{tune}}(\hr\zeta\hg)$, whose right side is affine in $\lambda$; this proves Prop.~\ref{prop:sieve-closed-form}.  \emph{Prop.~\ref{prop:sieve-moments}.} Because $\E_R[\kappa_{\mdl}\mid X]=1$ and $\E_R[\kappa_{\mdl}Z^\lambda_{\mdl}\mid X]=\tau_0(X)$, Assumption~\ref{ass:honest-separation} gives $\E[\widehat H_{\mdl}(\omega)]+\rho I=\bar H_\rho$, and the right side of Eq.~\eqref{eq:sieve-normal} has mean $\Gamma_R\beta_p+\omega\,\mathtt{imb}_{\zeta\hg}(\hr)$. Hence $\bar\beta=\bar H_\rho^{-1}\{\Gamma_R\beta_p+\omega\,\mathtt{imb}_{\zeta\hg}(\hr)\}$, and subtracting $\{\widehat H_{\mdl}(\omega)+\rho I\}\bar\beta$ from both sides of Eq.~\eqref{eq:sieve-normal} gives $\{\widehat H_{\mdl}(\omega)+\rho I\}(\widehat\beta_{\mdl}-\bar\beta)=\bar\psi_{\mdl}$, where $\bar\psi_{\mdl}\triangleq\mathbb P_R^{\mathrm{tune}}\psi_{R,\mdl}+\mathbb P_O^{\mathrm{tune}}\psi_O-\rho\bar\beta$ has mean zero and variance $V_{\mdl}$. Thus $\widehat\beta_{\mdl}-\bar\beta=\bar H_\rho^{-1}\bar\psi_{\mdl}+\mathtt{rem}$, so $\E[\widehat\beta_{\mdl}]=\bar\beta+\E[\mathtt{rem}]$ with $\|\E[\mathtt{rem}]\|_{\Gamma_R}\le(\E\|\mathtt{rem}\|^2_{\Gamma_R})^{1/2}=o_P(n_R^{-1/2}+n_O^{-1/2})$. Subtracting $\bar H_\rho\beta_p$ from $\Gamma_R\beta_p+\omega\,\mathtt{imb}_{\zeta\hg}(\hr)$ gives $\bar\beta-\beta_p=\bar H_\rho^{-1}[\omega\{\mathtt{imb}_{\zeta\hg}(\hr)-\mathtt{imb}_{\zeta\zeta^\top}(\hr)\beta_p\}-\rho\beta_p]$. Since $\mathtt{imb}_f(\hr)=\E_O[(\hr-r_0)f]$ under Assumption~\ref{ass:domain} and $\zeta^\top\beta_p=\tau^\star_p$, the braces equal $\E_O[(\hr-r_0)\zeta(\hg-\tau^\star_p)]$, which gives the display for $\bar\beta(\omega)-\beta_p$. For the bound, $\mathtt{bias}(\omega)^{1/2}=\|\Gamma_R^{1/2}(\bar\beta-\beta_p)\|\le\|\Gamma_R^{1/2}\bar H_\rho^{-1}\|_{\mathrm{op}}\|\omega\,\E_O[(\hr-r_0)\zeta(\hg-\tau^\star_p)]-\rho\beta_p\|$, and Jensen's and the Cauchy--Schwarz inequalities give $\|\E_O[(\hr-r_0)\zeta(\hg-\tau^\star_p)]\|\le\{\E_O(\hr-r_0)^2\,\E_O\|\zeta(\hg-\tau^\star_p)\|^2\}^{1/2}$, which proves the proposition. \emph{Thm.~\ref{thm:sieve-risk}.} By the identity $\mathcal R_0(\zeta^\top\beta)=\mathcal R_0(\tau^\star_p)+\|\beta-\beta_p\|^2_{\Gamma_R}$ shown above, $\mathcal R_0(\hzeta_{\mdl})=\mathcal R_0(\tau^\star_p)+\|\widehat\beta_{\mdl}-\beta_p\|^2_{\Gamma_R}$. Write $\widehat\beta_{\mdl}-\beta_p=(\widehat\beta_{\mdl}-\bar\beta)+(\bar\beta-\beta_p)$. Since $\bar H_\rho^{-1}\bar\psi_{\mdl}$ has mean zero and covariance $\bar H_\rho^{-1}V_{\mdl}\bar H_\rho^{-1}$, the Cauchy--Schwarz inequality gives $\E\|\widehat\beta_{\mdl}-\bar\beta\|^2_{\Gamma_R}=\mathcal J_{\mdl}+o_P(n_R^{-1}+n_O^{-1})$. The cross term is $2\E[\mathtt{rem}]^\top\Gamma_R(\bar\beta-\beta_p)=o_P(n_R^{-1}+n_O^{-1})$ by assumption, which proves the theorem. The two conditions on $\mathtt{rem}$ are assumptions of Prop.~\ref{prop:sieve-moments} and Thm.~\ref{thm:sieve-risk}; they hold under the usual fixed-$p$ expansion of $\widehat\beta_{\mdl}$, which we do not verify here.

\emph{Prop.~\ref{prop:sieve-balance}.} By the subgradient condition in Sec.~\ref{sec:ratio}, $\mathbb P_O^{\mathrm{nuis}}(\hr f_k)-\mathbb P_R^{\mathrm{nuis}}(f_k)=e_k$ with $|e_k|\le\widehat\delta_k$. Subtracting this from $\mathtt{imb}_{f_k}(\hr)=\E_O[\hr f_k]-\E_R[f_k]$ gives the display, and linearity of $g\mapsto\mathtt{imb}_g(\hr)$ extends it to linear combinations of the features. Part~(i) takes $g=\hg$; Thm.~\ref{thm:wald}(ii) treats the resulting drift as noise with variance $\widehat V_{\mathrm{cal}}$. For part~(ii), the $\omega$ term of Prop.~\ref{prop:sieve-moments} is $\mathtt{imb}_{\zeta\hg}(\hr)-\mathtt{imb}_{\zeta\,\zeta^\top}(\hr)\beta_p$, and for a candidate $\hzeta=\zeta^\top\beta$ in $\mathcal T_\zeta$, $(\hg-\hzeta)^2=\hg^2-2\beta^\top\zeta\hg+\beta^\top\zeta\,\zeta^\top\beta$, so $\varrho_{\mdl}=\omega\,\mathtt{imb}_{(\hg-\hzeta)^2}(\hr)$. Both are fixed linear combinations of the balanced imbalances given the tuning samples. Each nuisance-sample average in the display is $O_P(n_{\mathrm{nuis}}^{-1/2}+N_{\mathrm{nuis}}^{-1/2})$ under the conditions of Thm.~\ref{thm:wald}(ii), and so is $\widehat\delta_k$.

\section{Additional Simulation Results}
\label{app:sim}
% Panel and legend macros of Sec. 5 (re-declared here because the moved figures use them).
\makeatletter
\@ifundefined{dfpw}{\newlength\dfpw}{}\setlength\dfpw{0.19\textwidth}%
\providecommand{\lgs}{0.6}%
\providecommand{\pnl}[1]{\includegraphics[width=\dfpw]{#1}}%
\providecommand{\lgc}[5]{\includegraphics[scale=\lgs,trim=#2 #3 #4 #5,clip]{#1}}%
\makeatother
Appendix~\ref{app:exp-details} gives the full experimental setup, Appendix~\ref{sec:exp-sweeps} the bias and observational-sample-size sweeps, Appendix~\ref{sec:exp-ablation} the ablations and further designs, and Appendix~\ref{app:prog} the comparison with prognostic-score adjustment. Appendix~\ref{app:further} gives the remaining results.

\subsection{Experimental Details}
\label{app:exp-details}

A latent $U$ confounds the OBS in a synthetic design and two semi-synthetic benchmarks, IHDP \citep{hill2011bart} and ACIC~2016 \citep{dorie2019acic}; the bias has root mean square $1.5\,\mathrm{sd}_R\{\tau_0(X)\}$ in the synthetic design. The benchmarks keep only their real covariates and noiseless surfaces $\mu_0,\mu_1$, and both samples draw rows of this table. The trial randomizes $A$, and the OBS assigns $A$ through $X$ and $U$, which also enters $Y=\mu_A(X)+U+\varepsilon$. The OBS covariate law equals the trial law ($r_0=1$) or tilts it by $\exp\{a\,h(X)\}$ for a standardized direction $h$ (the first principal component in the benchmarks), so $r_0\propto\exp(-a\,h)$.

\paragraph{Synthetic design.} The covariates are $X=lU+0.9\,\epsilon_X\in\mathbb R^{10}$ with $U\sim N(0,1)$ and $\epsilon_X\sim N(0,I_{10})$, where $l$ loads $U$ on the first two coordinates with weights $\pm0.5$, so that $\Var(U\mid X)=0.62$. The outcome is $Y=m(X)+2U+A\tau(X)+0.7\varepsilon$ with $\varepsilon\sim N(0,1)$, where $m$ is a fixed combination of linear, sine, quadratic, and pairwise-product terms and $\tau$ is a nonlinear function with $\E_R\tau(X)=0.8$ and $\Var_R\tau(X)=1.47$. The trial sets $A\sim\mathrm{Bernoulli}(0.35)$. The OBS sets $A\sim\mathrm{Bernoulli}[\mathrm{clip}\{\mathrm{expit}(\eta(X)+\nu(X)U),0.05,0.95\}]$ for a fixed index $\eta$, with $\nu(x)=\nu_0\{1+\eta_{\mathrm{std}}(x)\}$ and $\eta_{\mathrm{std}}$ the index standardized over the trial covariates, so the confounding is strongest for units that are likely to be treated. The bias of the OBS contrast is $\delta(x)=2\{\E_O[U\mid A=1,X=x]-\E_O[U\mid A=0,X=x]\}$, computed by Gauss--Hermite quadrature, and $\nu_0$ sets $\{\E_R\delta(X)^2\}^{1/2}=1.5\,\mathrm{sd}_R\{\tau_0(X)\}$. The bias sweep changes only $\nu_0$ so that this root mean square is $0,0.25,\dots,1.25$ times $\mathrm{sd}_R\{\tau_0(X)\}$.

\paragraph{Benchmarks.} IHDP \citep{hill2011bart} has 747 units, 25 covariates, and 100 realizations of its response surfaces in the release that accompanies \citet{shalit2017estimating}. ACIC~2016 \citep{dorie2019acic} has 4{,}802 units and 58 covariates, three of them categorical and one-hot encoded, and we use its settings 1--10. We discard the treatments and outcomes recorded in both data sets and keep only the covariates and the noiseless surfaces. Both samples draw rows of the covariate table with replacement. The benchmarks are thus finite-population designs. An OBS sample of 15{,}000 rows contains each IHDP row about 20 times on average and each ACIC row about 3 times, and a trial of $n=3{,}000$ contains each IHDP row about 4 times. The CATE risk averages over the rows of the table (Metrics), so it is computed at covariate values that also appear in the OBS sample, and the repeated rows may make the OBS regression more accurate there than it would be at new covariate values. Replication $k$ uses IHDP realization $k \bmod 100$ and ACIC setting $(k\bmod 10)+1$. The published noiseless surfaces $\mu_0,\mu_1$ are scaled so that $\Var_R\mu_A(X)=1$, and $Y=\mu_A(X)+U+\varepsilon$ with $U,\varepsilon\sim N(0,1)$ independent of $X$. The trial draws rows uniformly and sets $A\sim\mathrm{Bernoulli}(0.35)$. The OBS treatment follows the rule of the synthetic design with $\eta$ linear in three standardized covariates, and $\nu_0$ sets $|\E_R\delta(X)|=0.50\,\mathrm{sd}_R(Y)$, where $\delta(x)=\E_O[U\mid A=1,X=x]-\E_O[U\mid A=0,X=x]$.

\paragraph{Covariate shift.} A shift changes the law of $X$ only, not that of $(U,A,Y)$ given $X$. In the synthetic design, the OBS covariates have mean $a\,d$ instead of zero for a fixed direction $d$, which is the exponential tilt $\exp(a\,d^\top\Sigma^{-1}x)$ with $\Sigma=\Cov(X)$. In the benchmarks, the OBS draws row $i$ with probability proportional to $\exp(a h_i)$, where $h$ is the standardized first principal component of the standardized covariates. Since the trial draws rows uniformly, the density ratio at row $i$ is $r_0(x_i)=\overline{e^{ah}}/e^{a h_i}$, where $\overline{e^{ah}}$ is the average of $e^{a h_j}$ over the rows, and $1/\E_O[r_0^2]=1/(\overline{e^{ah}}\,\overline{e^{-ah}})$. In the synthetic design, with $h(x)=d^\top\Sigma^{-1}x$ and $d^\top\Sigma^{-1}d=1$, $r_0(x)=\exp\{a^2/2-a\,h(x)\}$ and $1/\E_O[r_0^2]=e^{-a^2}$. The constant $a$ sets the effective sample fraction $1/\E_O[r_0^2]$ to $0.40$ (the shift of Fig.~\ref{fig:results}) or $0.70$ (the weak shift of Tables~\ref{tab:app-ate-ratio} and~\ref{tab:app-cate-ratio}).

\paragraph{Samples and estimators.} A trial of size $n$ is split into equal nuisance, tuning, and evaluation samples, and the OBS has $N=15{,}000$ rows split $60/20/20$; Fig.~\ref{fig:obs} varies $N$. Every split-based method is cross-fitted: the three trial roles rotate over three equal blocks, each method runs once per rotation, and the three ATE estimates or CATE predictions are averaged. The naive pool instead fits one regression on all trial and OBS rows, and Shrinkage and Pretest combine it with the cross-fitted trial-only estimate. The OBS keeps one split because it is large. Here $n$ and $N$ are the total trial and OBS sizes, so the evaluation sizes in each rotation, written $n$ and $N$ in Secs.~\ref{sec:ate} and~\ref{sec:cate}, are $n/3$ and $0.2N$. Thm.~\ref{thm:wald} applies to each rotation. The Wald interval of the averaged ATE uses a first-order standard error of the average: each trial unit contributes its score $Z^{\hlam}-\homg\hg$ from the rotation that evaluates its block and, with an estimated ratio, the calibration term $\homg\hg(X)$ from the rotation whose nuisance sample contains it, and the OBS adds the variances of its rotation-averaged evaluation scores and calibration terms $\homg\hr\hg$. The same OBS random numbers are used under every covariate law. Both outcome regressions are ridge regressions (penalty 5) of $Y$ on $(f,A,Af)$, where $f$ standardizes $X$, $X^2$, $\sin X$, and the pairwise products of the first six covariates, and their predictions are clipped to the 0.5\% to 99.5\% range of each arm on the fitting sample. $Z_1,\dots,Z_5$ are the first five principal components of the standardized trial nuisance covariates, scaled to unit variance. When $r_0=1$, we set $\hr\equiv1$ (Assumption~\ref{ass:domain}(i)). Under a shift, the base ratio is the classifier ratio of Sec.~\ref{sec:ratio} with a logistic regression on $(Z_1,\dots,Z_5)$, calibrated by Def.~\ref{def:bal}: for the ATE it balances $\hg$ exactly, and for the CATE it balances $\hg^2$, $\zeta\hg$, and $\zeta_k\zeta_l$ ($k\le l$) within their standard errors $\widehat\delta_k$ (Prop.~\ref{prop:sieve-balance}). AIPWF uses $\epsilon_A=\epsilon_B=10^{-3}\widehat{\Var}_R(Z^0)$ in Eq.~\eqref{eq:coef-select}, and PPI\textsuperscript{++} fixes $\lambda=0$ and uses $\homg=\mathrm{Proj}_{[0,1]}(\hD/\hB)$, the rule of \citet{angelopoulos2026ppipp} with our projection, without $\hB_{\mathrm{cal}}$. Shrinkage weights the naive pool by $\mathrm{se}_R^2/(\mathrm{se}_R^2+\mathrm{gap}^2)$, where $\mathrm{gap}$ is the difference of the naive-pool and trial-only ATEs and $\mathrm{se}_R$ is the trial-only standard error, and Pretest uses the naive pool when the gap is within $1.96\,\mathrm{se}_R$. Shrinkage is a single-estimate rule in the spirit of \citet{rosenman2023shrinkage}, not their Stein-type estimator, which shrinks the stratum estimates jointly. DRF and RF use the sieve $\zeta=(1,Z_1,\dots,Z_5,\hg)$, ridge $\rho=0.01$, and the grid $\{0,0.5,1\}^2$, with ties broken toward smaller $(\lambda,\omega)$; their PPI\textsuperscript{++} versions use the grid points with $\lambda=0$. The trial-only learners use $\hmu_R$ and the sieve $(1,Z_1,\dots,Z_5)$ without $\hg$. 2-step fits $\hg+(1,Z_1,\dots,Z_5)^\top\beta$, with $\beta$ a ridge regression of the trial pseudo-outcome minus $\hg$ \citep{kallus2018removing}. IR \citep{wu2022integrative} minimizes one $R$-loss over the trial and OBS rows, in which the OBS rows add a bias function $c(x)$ to $\tau(x)$. It takes $\tau$ in $(1,Z_1,\dots,Z_5)$ and $c$ in $(1,Z^O_1,\dots,Z^O_5,\hg)$, where $Z^O_k$ are the principal components of the OBS nuisance covariates, and uses an OBS propensity from logistic regression. A ridge penalty replaces their SCAD penalty, and the penalty ratio $\lambda_\tau/\lambda_c\in\{0,0.5,1,1.5\}$ and $\lambda_c\in\{10^{-3},10^{-2},\dots,10^{2}\}$ are chosen by the $R$-loss of both sources on the evaluation samples instead of cross-validation. The naive pool fits the same outcome regression to all trial and OBS rows, with no source indicator or weight and without clipping.

\paragraph{Metrics.} Each cell has 1{,}000 replications, seeded by the base seed 20261005, the design, the trial size, and the replication index, and no replication failed except 2 of the 10{,}000 misspecified-shift replications (IHDP, $n=600$ and $3{,}000$), whose ratio calibration did not converge and which are left out. The ATE error is the estimate minus $\theta_0$, and the CATE risk is $\E_R\{\hat\tau(X)-\tau_0(X)\}^2$ over 50{,}000 trial covariate draws (synthetic design) or all rows (benchmarks). The figures and Tables~\ref{tab:app-ate-ratio} and~\ref{tab:app-cate-ratio} report ratios of mean errors to the trial-only AIPW estimator or DR-learner; the figures add paired percentile 95\% intervals from 2{,}000 bootstrap resamples of the replications.

\subsection{Bias and Observational Sample Size}
\label{sec:exp-sweeps}

\begin{figure}[tp]
\centering
\begin{minipage}[c]{\dfpw}\pnl{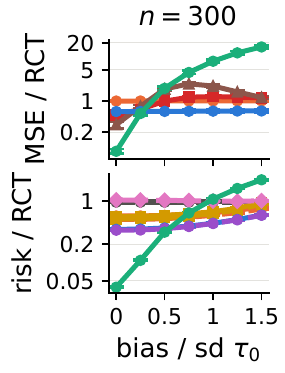}\end{minipage}\hfill
\begin{minipage}[c]{\dfpw}\pnl{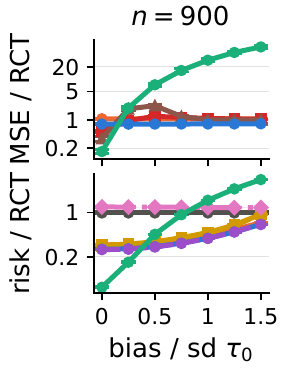}\end{minipage}\hfill
\begin{minipage}[c]{\dfpw}\pnl{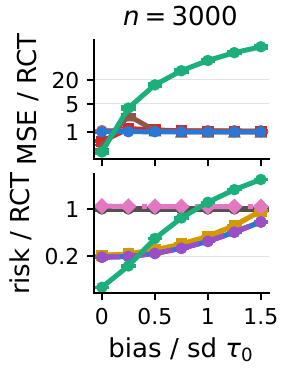}\end{minipage}\hfill
\begin{minipage}[c]{\dfpw}\raggedright\offinterlineskip
\hspace*{0.1\dfpw}\lgc{s5_legend_ate}{0}{0}{204.6}{0}\par\vspace{2pt}
\hspace*{0.1\dfpw}\lgc{s5_legend_ate}{93.46}{0}{103.6}{0}\par\vspace{2pt}
\hspace*{0.1\dfpw}\lgc{s5_legend_ate}{194.5}{0}{0}{0}
\end{minipage}\hfill
\begin{minipage}[c]{\dfpw}\raggedright\offinterlineskip
\lgc{s5_legend_cate_full}{0}{0}{378.8}{0}\par\vspace{1pt}
\lgc{s5_legend_cate_full}{95}{0}{271.8}{0}\par\vspace{1pt}
\lgc{s5_legend_cate_full}{202.3}{0}{171.8}{0}\par\vspace{1pt}
\lgc{s5_legend_cate_full}{302.4}{0}{106.3}{0}\par\vspace{1pt}
\lgc{s5_legend_cate_full}{367.7}{0}{0}{0}
\end{minipage}
\caption{Bias sweep on the synthetic design under the strong shift at $n=300$, $900$, and $3{,}000$, against the confounding bias in units of $\mathrm{sd}_R\,\tau_0(X)$. Top: ATE MSE relative to trial-only AIPW; bottom: CATE risk relative to the trial-only DR-learner. 95\% bootstrap intervals.}
\label{fig:sweep-n}
\end{figure}

\begin{figure}[tp]
\centering
\includegraphics[scale=\lgs]{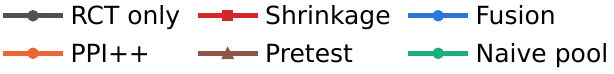}\hfill\includegraphics[scale=\lgs]{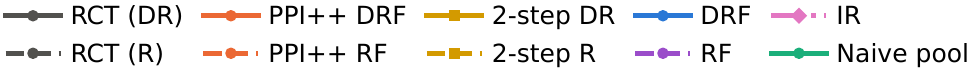}\par\vspace{3pt}
\pnl{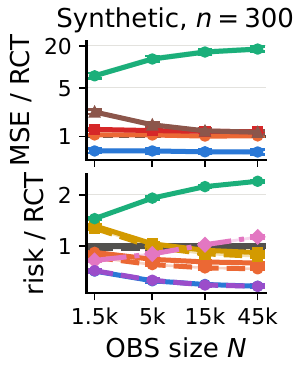}\hfill\pnl{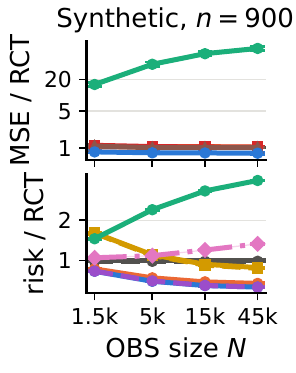}\hfill\pnl{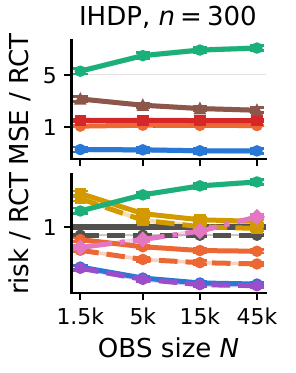}\hfill\pnl{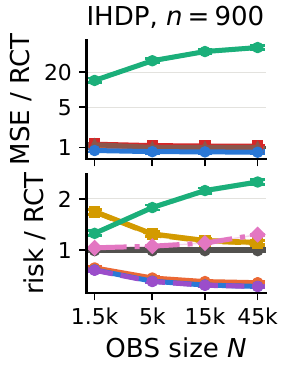}
\caption{Effect of the observational sample size $N$ (split 60/20/20) under the covariate shift, for the synthetic design and IHDP at $n=300$ and $900$. Top: ATE MSE relative to RCT only; bottom: CATE risk relative to the trial-only DR-learner. 95\% bootstrap intervals.}
\label{fig:obs}
\end{figure}

 Fig.~\ref{fig:sweep-n} repeats the bias sweep at $n=300$ and $900$. The ATE ratio of fusion is flat in the bias at every $n$, and at $n=3{,}000$ the CATE ratio of IR stays between 1.08 and 1.10 at every bias. The naive pool, Shrinkage, and Pretest beat fusion when the OBS are nearly unconfounded. At $n=3{,}000$ the naive pool falls behind fusion at a bias between 0 and 0.25 (ATE) or between 0.25 and 0.5 (CATE) standard deviations. This threshold decreases as $n$ grows: for the CATE, it lies between 0.5 and 0.75 standard deviations at $n=300$ and between 0.25 and 0.5 at $n=3{,}000$. In Fig.~\ref{fig:obs}, the ATE ratio of fusion changes little with $N$, and its CATE ratio falls as $N$ grows. Fig.~\ref{fig:abs} gives the absolute errors.

\subsection{Ablations and Further Designs}
\label{sec:exp-ablation}

\paragraph{Patterns in Fig.~\ref{fig:results}.} The ATE gap of fusion below 1 is widest at small $n$ and narrows by $n=3{,}000$, so the OBS help the ATE most when the trial is small; the CATE ratios of DRF and RF stay between 0.53 and 0.67 at every $n$. PPI\textsuperscript{++} stays near 1 for the ATE and approaches DRF and RF for the CATE only at $n=3{,}000$, so the gain of fusion at small $n$ comes from $\lambda$. For the ATE, this is the cap in Remark~\ref{rem:channels}: the PPI correction can remove at most the share of effect heterogeneity in the score variance, which is small in these designs: at $r_0=1$, $\Var_R\{\tau_0(X)\}/\Var_R(Z^0)$ is 0.02 to 0.10 on the tuning samples. For the CATE, the trial-only learners use the sieve $(1,Z_1,\dots,Z_5)$ without $\hg$, whereas DRF, RF, and their PPI\textsuperscript{++} versions use $(1,Z_1,\dots,Z_5,\hg)$, so the ratios to the trial-only learners combine the fusion loss with the gain from $\hg$ as a basis function. The trial loss on the same sieve separates the two (Table~\ref{tab:app-samesieve}): at large $n$ the gap comes from $\hg$ as a basis function, and at small $n$ mostly from the fusion loss. IR, in contrast, reaches or exceeds the trial-only risk by $n=3{,}000$ and is below it only on ACIC and on IHDP at $n=300$. IR models the confounding function in a linear class, so the OBS rows inform $\tau$ only as far as the OBS bias is close to that class. Its CATE class $(1,Z_1,\dots,Z_5)$ is the sieve of the trial-only learners, which is consistent with its ratio approaching or exceeding 1 as $n$ grows. DRF and RF do not model the confounding function, keep the trial's target under any OBS confounding (Thm.~\ref{prop:cate}), and use the OBS to reduce variance. The naive pool rises with $n$, because its confounding bias stays fixed while the trial-only error shrinks. Shrinkage and Pretest never fall below 1 in columns 1--4.

\begin{table}[tp]
\caption{The two channels of fusion. ATE MSE relative to trial-only AIPW of $\lambda$ only ($\omega=0$), PPI\textsuperscript{++} ($\lambda=0$, $\homg=\mathrm{Proj}_{[0,1]}(\hD/\hB)$), fusion with $\homg$ of Eq.~\eqref{eq:coef-select} or with $\homg=\mathrm{Proj}_{[0,1]}(\hD/\hB)$.}
\label{tab:app-channels}
\begin{center}\footnotesize
\begin{tabular}{@{}llccccccccc@{}}
\toprule
& & \multicolumn{3}{c}{Synthetic} & \multicolumn{3}{c}{IHDP} & \multicolumn{3}{c}{ACIC~2016}\\
\cmidrule(lr){3-5}\cmidrule(lr){6-8}\cmidrule(lr){9-11}
$n$ & Method & $r_0{=}1$ & weak & strong & $r_0{=}1$ & weak & strong & $r_0{=}1$ & weak & strong\\
\midrule
300 & $\lambda$ only & 0.60 & 0.60 & 0.60 & 0.47 & 0.46 & 0.46 & 0.46 & 0.46 & 0.46 \\
 & PPI\textsuperscript{++} & 0.95 & 1.01 & 1.01 & 1.00 & 1.04 & 1.05 & 1.00 & 1.02 & 1.02 \\
 & Fusion & 0.55 & 0.60 & 0.60 & 0.46 & 0.47 & 0.48 & 0.45 & 0.46 & 0.47 \\
 & Fusion, $\homg=\hD/\hB$ & 0.55 & 0.62 & 0.62 & 0.46 & 0.49 & 0.50 & 0.45 & 0.47 & 0.48 \\
\addlinespace
900 & $\lambda$ only & 0.78 & 0.78 & 0.78 & 0.83 & 0.84 & 0.83 & 0.42 & 0.42 & 0.43 \\
 & PPI\textsuperscript{++} & 0.97 & 1.04 & 1.05 & 1.00 & 1.02 & 1.02 & 1.00 & 1.01 & 1.01 \\
 & Fusion & 0.75 & 0.79 & 0.80 & 0.82 & 0.84 & 0.84 & 0.42 & 0.42 & 0.43 \\
 & Fusion, $\homg=\hD/\hB$ & 0.75 & 0.82 & 0.83 & 0.82 & 0.86 & 0.85 & 0.42 & 0.43 & 0.44 \\
\addlinespace
3{,}000 & $\lambda$ only & 0.97 & 0.97 & 0.97 & 0.95 & 0.96 & 0.97 & 0.78 & 0.77 & 0.78 \\
 & PPI\textsuperscript{++} & 0.99 & 1.06 & 1.06 & 0.97 & 1.04 & 1.00 & 1.00 & 1.02 & 1.01 \\
 & Fusion & 0.96 & 1.00 & 1.00 & 0.92 & 0.98 & 0.96 & 0.77 & 0.79 & 0.79 \\
 & Fusion, $\homg=\hD/\hB$ & 0.96 & 1.04 & 1.04 & 0.92 & 1.01 & 0.97 & 0.77 & 0.80 & 0.80 \\
\bottomrule
\end{tabular}
\end{center}
\end{table}

\begin{table}[tp]
\caption{Ratios to the practitioner's trial-only estimators (3-fold cross-fitting, nuisances fitted on two thirds of the trial, no tuning sample), marked (2/3). The rotated trial-only estimators of Tables~\ref{tab:app-ate-ratio} and~\ref{tab:app-cate-ratio}, whose nuisances use one third, are marked (1/3).}
\label{tab:app-practitioner}
\begin{center}\footnotesize
\begin{tabular}{@{}llccccccccc@{}}
\toprule
& & \multicolumn{3}{c}{Synthetic} & \multicolumn{3}{c}{IHDP} & \multicolumn{3}{c}{ACIC~2016}\\
\cmidrule(lr){3-5}\cmidrule(lr){6-8}\cmidrule(lr){9-11}
$n$ & Method & $r_0{=}1$ & weak & strong & $r_0{=}1$ & weak & strong & $r_0{=}1$ & weak & strong\\
\midrule
300 & Fusion / trial-only (2/3) & 0.70 & 0.76 & 0.76 & 0.63 & 0.64 & 0.65 & 0.40 & 0.41 & 0.41 \\
 & Trial-only (1/3) / (2/3) & 1.27 & 1.27 & 1.27 & 1.37 & 1.37 & 1.37 & 0.89 & 0.89 & 0.89 \\
 & DRF / DR (2/3) & 0.67 & 0.67 & 0.68 & 0.67 & 0.68 & 0.69 & 0.56 & 0.57 & 0.59 \\
 & RF / DR (2/3) & 0.66 & 0.67 & 0.67 & 0.66 & 0.67 & 0.68 & 0.56 & 0.56 & 0.58 \\
 & RF / R (2/3) & 0.69 & 0.70 & 0.70 & 0.70 & 0.70 & 0.71 & 0.58 & 0.58 & 0.61 \\
 & DR (1/3) / DR (2/3) & 1.14 & 1.14 & 1.14 & 1.20 & 1.20 & 1.20 & 1.05 & 1.05 & 1.05 \\
\addlinespace
900 & Fusion / trial-only (2/3) & 0.84 & 0.89 & 0.90 & 0.92 & 0.95 & 0.94 & 0.64 & 0.64 & 0.66 \\
 & Trial-only (1/3) / (2/3) & 1.12 & 1.12 & 1.12 & 1.12 & 1.12 & 1.12 & 1.52 & 1.52 & 1.52 \\
 & DRF / DR (2/3) & 0.63 & 0.64 & 0.64 & 0.60 & 0.61 & 0.62 & 0.65 & 0.67 & 0.69 \\
 & RF / DR (2/3) & 0.63 & 0.64 & 0.64 & 0.60 & 0.61 & 0.62 & 0.64 & 0.66 & 0.68 \\
 & RF / R (2/3) & 0.64 & 0.64 & 0.64 & 0.60 & 0.61 & 0.62 & 0.67 & 0.69 & 0.71 \\
 & DR (1/3) / DR (2/3) & 0.99 & 0.99 & 0.99 & 1.00 & 1.00 & 1.00 & 1.11 & 1.11 & 1.11 \\
\addlinespace
3{,}000 & Fusion / trial-only (2/3) & 0.97 & 1.00 & 1.01 & 0.95 & 1.01 & 0.99 & 0.90 & 0.92 & 0.92 \\
 & Trial-only (1/3) / (2/3) & 1.01 & 1.01 & 1.01 & 1.03 & 1.03 & 1.03 & 1.17 & 1.17 & 1.17 \\
 & DRF / DR (2/3) & 0.61 & 0.61 & 0.61 & 0.54 & 0.55 & 0.57 & 0.63 & 0.66 & 0.68 \\
 & RF / DR (2/3) & 0.61 & 0.61 & 0.61 & 0.54 & 0.55 & 0.57 & 0.63 & 0.66 & 0.68 \\
 & RF / R (2/3) & 0.61 & 0.61 & 0.61 & 0.54 & 0.55 & 0.57 & 0.63 & 0.66 & 0.68 \\
 & DR (1/3) / DR (2/3) & 0.95 & 0.95 & 0.95 & 0.99 & 0.99 & 0.99 & 1.01 & 1.01 & 1.01 \\
\bottomrule
\end{tabular}
\end{center}
\end{table}

\begin{table}[tp]
\caption{CATE with the trial loss alone on the sieve with $\hg$: its risk relative to the trial-only DR-learner, and the risk of DRF and RF relative to it.}
\label{tab:app-samesieve}
\begin{center}\footnotesize
\begin{tabular}{@{}llccccccccc@{}}
\toprule
& & \multicolumn{3}{c}{Synthetic} & \multicolumn{3}{c}{IHDP} & \multicolumn{3}{c}{ACIC~2016}\\
\cmidrule(lr){3-5}\cmidrule(lr){6-8}\cmidrule(lr){9-11}
$n$ & Method & $r_0{=}1$ & weak & strong & $r_0{=}1$ & weak & strong & $r_0{=}1$ & weak & strong\\
\midrule
300 & Trial loss, sieve with $\hat g$ (DR) & 0.84 & 0.83 & 0.83 & 0.84 & 0.84 & 0.85 & 0.90 & 0.92 & 0.93 \\
 & Trial loss, sieve with $\hat g$ (R) & 0.79 & 0.78 & 0.78 & 0.75 & 0.76 & 0.76 & 0.88 & 0.90 & 0.91 \\
 & DRF / same sieve (DR) & 0.69 & 0.70 & 0.72 & 0.67 & 0.68 & 0.68 & 0.60 & 0.59 & 0.60 \\
 & RF / same sieve (R) & 0.74 & 0.74 & 0.76 & 0.74 & 0.74 & 0.74 & 0.61 & 0.59 & 0.61 \\
\addlinespace
900 & Trial loss, sieve with $\hat g$ (DR) & 0.68 & 0.67 & 0.67 & 0.63 & 0.64 & 0.65 & 0.76 & 0.78 & 0.79 \\
 & Trial loss, sieve with $\hat g$ (R) & 0.67 & 0.66 & 0.65 & 0.62 & 0.63 & 0.63 & 0.70 & 0.72 & 0.73 \\
 & DRF / same sieve (DR) & 0.94 & 0.96 & 0.97 & 0.95 & 0.96 & 0.96 & 0.77 & 0.78 & 0.79 \\
 & RF / same sieve (R) & 0.96 & 0.98 & 1.00 & 0.97 & 0.97 & 0.98 & 0.82 & 0.84 & 0.84 \\
\addlinespace
3{,}000 & Trial loss, sieve with $\hat g$ (DR) & 0.64 & 0.63 & 0.62 & 0.55 & 0.56 & 0.57 & 0.64 & 0.66 & 0.67 \\
 & Trial loss, sieve with $\hat g$ (R) & 0.64 & 0.63 & 0.62 & 0.55 & 0.56 & 0.57 & 0.63 & 0.65 & 0.67 \\
 & DRF / same sieve (DR) & 1.00 & 1.02 & 1.02 & 1.00 & 1.01 & 1.01 & 0.98 & 0.99 & 1.00 \\
 & RF / same sieve (R) & 1.01 & 1.02 & 1.03 & 1.00 & 1.01 & 1.01 & 0.99 & 1.00 & 1.01 \\
\bottomrule
\end{tabular}
\end{center}
\end{table}

\begin{table}[tp]
\caption{Misspecified shift (tilt $\exp\{a(h^2-1)/\sqrt2\}$, effective sample fraction 0.40): fusion with the calibrated ratio $\hr_{\mathrm{BAL}}$ and with the uncalibrated logistic ratio $\hr_{\mathrm{CLS}}$ and $\homg=\mathrm{Proj}_{[0,1]}(\hD/\hB)$. MSE ratio: relative to RCT only.}
\label{tab:app-misspec}
\begin{center}\footnotesize
\begin{tabular}{@{}lllrrrrr@{}}
\toprule
Design & $n$ & Method & Bias & SD & RMSE & Cov. & MSE ratio \\
\midrule
Synthetic & 300 & RCT only & 0.007 & 0.302 & 0.302 & 0.95 & 1.00 \\
 & & Fusion & 0.002 & 0.232 & 0.232 & 0.94 & 0.59 \\
 & & Fusion, uncalibrated $\hat r_{\mathrm{CLS}}$ & 0.123 & 0.233 & 0.264 & 0.90 & 0.76 \\
\addlinespace
 & 900 & RCT only & -0.007 & 0.139 & 0.139 & 0.95 & 1.00 \\
 & & Fusion & -0.005 & 0.123 & 0.123 & 0.96 & 0.78 \\
 & & Fusion, uncalibrated $\hat r_{\mathrm{CLS}}$ & 0.105 & 0.124 & 0.163 & 0.87 & 1.36 \\
\addlinespace
 & 3{,}000 & RCT only & -0.002 & 0.073 & 0.073 & 0.95 & 1.00 \\
 & & Fusion & -0.002 & 0.072 & 0.072 & 0.95 & 0.99 \\
 & & Fusion, uncalibrated $\hat r_{\mathrm{CLS}}$ & 0.080 & 0.074 & 0.109 & 0.76 & 2.24 \\
\midrule
IHDP & 300 & RCT only & -0.007 & 0.255 & 0.255 & 0.96 & 1.00 \\
 & & Fusion & -0.002 & 0.177 & 0.176 & 0.95 & 0.48 \\
 & & Fusion, uncalibrated $\hat r_{\mathrm{CLS}}$ & -0.032 & 0.182 & 0.184 & 0.94 & 0.52 \\
\addlinespace
 & 900 & RCT only & -0.003 & 0.114 & 0.114 & 0.96 & 1.00 \\
 & & Fusion & -0.003 & 0.105 & 0.105 & 0.95 & 0.84 \\
 & & Fusion, uncalibrated $\hat r_{\mathrm{CLS}}$ & -0.030 & 0.118 & 0.121 & 0.91 & 1.13 \\
\addlinespace
 & 3{,}000 & RCT only & 0.000 & 0.056 & 0.056 & 0.96 & 1.00 \\
 & & Fusion & 0.001 & 0.055 & 0.055 & 0.96 & 0.97 \\
 & & Fusion, uncalibrated $\hat r_{\mathrm{CLS}}$ & -0.019 & 0.068 & 0.071 & 0.88 & 1.58 \\
\bottomrule
\end{tabular}
\end{center}
\end{table}

\begin{table}[tp]
\caption{Synthetic design with large effect heterogeneity (outcome noise 0.2, $\Var_R\tau_0(X)=5.88$, confounding bias $0.75\,\mathrm{sd}_R\tau_0(X)$), under $r_0=1$ and the strong shift. MSE ratio: relative to RCT only. Prog.\ score and Cov.~+~prog.\ score are the prognostic-score adjustments of Sec.~\ref{sec:exp-estimators}.}
\label{tab:app-hethi}
\begin{center}\footnotesize
\begin{tabular}{@{}lllrrrrr@{}}
\toprule
Law & $n$ & Method & Bias & SD & RMSE & Cov. & MSE ratio \\
\midrule
$r_0{=}1$ & 300 & RCT only & 0.008 & 0.307 & 0.307 & 0.96 & 1.00 \\
 & & PPI\textsuperscript{++} & 0.007 & 0.281 & 0.281 & 0.96 & 0.84 \\
 & & $\lambda$ only & 0.013 & 0.237 & 0.238 & 0.95 & 0.60 \\
 & & Fusion & 0.012 & 0.205 & 0.206 & 0.95 & 0.45 \\
 & & Fusion, $\homg=\hD/\hB$ & 0.012 & 0.205 & 0.206 & 0.95 & 0.45 \\
 &  & Prog.\ score & 0.012 & 0.360 & 0.360 & 0.95 & 1.38 \\
 &  & Cov.~+~prog.\ score & 0.009 & 0.262 & 0.262 & 0.94 & 0.73 \\
\addlinespace
 & 900 & RCT only & 0.000 & 0.149 & 0.149 & 0.95 & 1.00 \\
 & & PPI\textsuperscript{++} & -0.003 & 0.136 & 0.136 & 0.95 & 0.83 \\
 & & $\lambda$ only & 0.000 & 0.138 & 0.138 & 0.95 & 0.86 \\
 & & Fusion & -0.003 & 0.122 & 0.122 & 0.95 & 0.67 \\
 & & Fusion, $\homg=\hD/\hB$ & -0.003 & 0.122 & 0.122 & 0.95 & 0.67 \\
 &  & Prog.\ score & -0.004 & 0.202 & 0.202 & 0.95 & 1.84 \\
 &  & Cov.~+~prog.\ score & -0.004 & 0.151 & 0.151 & 0.94 & 1.03 \\
\addlinespace
 & 3{,}000 & RCT only & 0.002 & 0.074 & 0.074 & 0.95 & 1.00 \\
 & & PPI\textsuperscript{++} & 0.003 & 0.072 & 0.072 & 0.95 & 0.93 \\
 & & $\lambda$ only & 0.002 & 0.073 & 0.073 & 0.96 & 0.96 \\
 & & Fusion & 0.003 & 0.070 & 0.070 & 0.95 & 0.89 \\
 & & Fusion, $\homg=\hD/\hB$ & 0.003 & 0.070 & 0.070 & 0.95 & 0.89 \\
 &  & Prog.\ score & 0.001 & 0.111 & 0.111 & 0.96 & 2.24 \\
 &  & Cov.~+~prog.\ score & 0.002 & 0.081 & 0.081 & 0.96 & 1.20 \\
\midrule
strong & 300 & RCT only & 0.008 & 0.307 & 0.307 & 0.96 & 1.00 \\
 & & PPI\textsuperscript{++} & 0.008 & 0.317 & 0.317 & 0.96 & 1.07 \\
 & & $\lambda$ only & 0.013 & 0.238 & 0.238 & 0.96 & 0.60 \\
 & & Fusion & 0.014 & 0.241 & 0.241 & 0.95 & 0.62 \\
 & & Fusion, $\homg=\hD/\hB$ & 0.013 & 0.250 & 0.250 & 0.95 & 0.66 \\
 &  & Prog.\ score & 0.012 & 0.361 & 0.361 & 0.95 & 1.38 \\
 &  & Cov.~+~prog.\ score & 0.009 & 0.263 & 0.263 & 0.95 & 0.73 \\
\addlinespace
 & 900 & RCT only & 0.000 & 0.149 & 0.149 & 0.95 & 1.00 \\
 & & PPI\textsuperscript{++} & -0.004 & 0.156 & 0.156 & 0.96 & 1.10 \\
 & & $\lambda$ only & 0.000 & 0.138 & 0.138 & 0.95 & 0.86 \\
 & & Fusion & -0.001 & 0.140 & 0.140 & 0.95 & 0.89 \\
 & & Fusion, $\homg=\hD/\hB$ & -0.003 & 0.147 & 0.147 & 0.95 & 0.97 \\
 &  & Prog.\ score & -0.004 & 0.202 & 0.202 & 0.96 & 1.85 \\
 &  & Cov.~+~prog.\ score & -0.004 & 0.151 & 0.151 & 0.94 & 1.03 \\
\addlinespace
 & 3{,}000 & RCT only & 0.002 & 0.074 & 0.074 & 0.95 & 1.00 \\
 & & PPI\textsuperscript{++} & 0.003 & 0.085 & 0.085 & 0.95 & 1.30 \\
 & & $\lambda$ only & 0.002 & 0.073 & 0.073 & 0.96 & 0.96 \\
 & & Fusion & 0.003 & 0.077 & 0.077 & 0.95 & 1.06 \\
 & & Fusion, $\homg=\hD/\hB$ & 0.003 & 0.083 & 0.083 & 0.95 & 1.26 \\
 &  & Prog.\ score & 0.001 & 0.112 & 0.111 & 0.96 & 2.24 \\
 &  & Cov.~+~prog.\ score & 0.002 & 0.081 & 0.081 & 0.96 & 1.19 \\
\bottomrule
\end{tabular}
\end{center}
\end{table}

\paragraph{Two channels.} Table~\ref{tab:app-channels} separates $\lambda$ and $\omega$. At $r_0=1$, $\hB_{\mathrm{cal}}=0$, and fusion is at or below $\lambda$ only in 14 of 15 cells, for example 0.55 against 0.60 on the synthetic design at $n=300$. Under a shift, $\lambda$ only is slightly below fusion in 28 of 30 cells, by at most 0.03, so once the ratio is estimated the PPI correction does not help in these designs. The rule $\homg=\mathrm{Proj}_{[0,1]}(\hD/\hB)$, which ignores $\hB_{\mathrm{cal}}$, does worse: Eq.~\eqref{eq:coef-select} lowers the MSE of fusion by 1\% to 4\% in all 30 shift cells, and the old rule exceeds the trial-only MSE in three cells at $n=3{,}000$ (1.01 to 1.04).

\paragraph{Practitioner's baselines.} A practitioner without a tuning sample would use 3-fold cross-fitted trial-only AIPW and DR- and R-learners with the nuisances fitted on two thirds of the trial. This AIPW has lower MSE than the rotated trial-only AIPW of Table~\ref{tab:app-ate-ratio} in 42 of 45 cells (Table~\ref{tab:app-practitioner}). Fusion is still below it in 42 of 45 cells; the exceptions are three shift cells at $n=3{,}000$, at 1.00 to 1.01. DRF and RF are below the practitioner's DR- and R-learners in all 45 cells, at 0.54 to 0.71 of their risk.

\paragraph{Same sieve.} Table~\ref{tab:app-samesieve} fits the trial loss alone on the sieve $(1,Z_1,\dots,Z_5,\hg)$, the grid point $(\lambda,\omega)=(0,0)$ of Algorithm~\ref{alg:cate}. It uses $\hg$, so it is not a trial-only learner. It has 0.75 to 0.93 of the trial-only DR risk at $n=300$ and 0.55 to 0.67 at $n=3{,}000$. DRF and RF have 0.59 to 0.76 of its risk at $n=300$ and 0.98 to 1.03 at $n=3{,}000$. Thus at small $n$ most of the gain of DRF and RF over the trial-only learners comes from the fusion loss, and at large $n$ it comes from $\hg$ as a basis function.

\paragraph{Misspecified ratio.} Table~\ref{tab:app-misspec} uses a shift that the base ratio cannot represent. On the synthetic design and IHDP, the OBS tilts the trial law by $\exp\{a(h^2-1)/\sqrt2\}$ with $a>0$ and effective sample fraction 0.40, so the OBS spread more widely along $h$, while the logistic classifier on $(Z_1,\dots,Z_5)$ is log-linear in $h$. With the uncalibrated $\hr_{\mathrm{CLS}}$, for which $\homg=\mathrm{Proj}_{[0,1]}(\hD/\hB)$ and the standard error omits $\widehat V_{\mathrm{cal}}$, fusion is biased (0.08 to 0.12 on the synthetic design and 0.02 to 0.03 on IHDP), its coverage falls to 0.76 to 0.94, and its MSE exceeds the trial-only MSE at $n\ge900$ (1.13 to 2.24). With $\hr_{\mathrm{BAL}}$, the absolute bias is at most 0.005, the coverage is 0.94 to 0.96, and the MSE is 0.48 to 0.99 of the trial-only MSE. DRF and RF have 0.52 to 0.62 of the trial-only CATE risk here.

\paragraph{Large heterogeneity.} Table~\ref{tab:app-hethi} changes the synthetic design so that $\omega$ can matter: the outcome noise is 0.2 instead of 0.7, $\Var_R\tau_0(X)=5.88$ instead of 1.47, and $\nu_0$ is unchanged, so the confounding bias is $0.75\,\mathrm{sd}_R\tau_0(X)$. The heterogeneity share $\Var_R\{\tau_0(X)\}/\Var_R(Z^0)$ rises to 0.20 to 0.33. At $r_0=1$, PPI\textsuperscript{++} has 0.83 to 0.93 of the trial-only MSE, and fusion improves on $\lambda$ only at every $n$, from 0.60 to 0.45 at $n=300$ and from 0.96 to 0.89 at $n=3{,}000$. Under the strong shift, PPI\textsuperscript{++} is above 1 (1.07 to 1.30), and fusion is above $\lambda$ only (0.62 against 0.60 at $n=300$, 1.06 against 0.96 at $n=3{,}000$). Eq.~\eqref{eq:coef-select} reduces this loss relative to the rule $\homg=\mathrm{Proj}_{[0,1]}(\hD/\hB)$ (0.66 and 1.26) but does not remove it. The coverage of fusion is 0.95 to 0.96 throughout. The prognostic-score adjustments do worse here: Prog.\ score has 1.38 to 2.24 of the trial-only MSE, and Cov.~+~prog.\ score has 0.73 at $n=300$ and 1.03 to 1.20 for $n\ge900$, against 0.45 to 1.06 for fusion.

\subsection{Comparison with Prognostic-Score Adjustment}
\label{app:prog}

\paragraph{Estimators.} Prog.\ score is the least-squares fit of $Y$ on $(1,A,s,As)$ over the whole trial, where $s$ is $\hmu_O(X,0)$ centered on the trial, and Cov.~+~prog.\ score is the least-squares fit of $Y$ on $(1,A,X,\hmu_O(X,0))$. Both use the $\hmu_O$ of fusion, report the coefficient of $A$ with a sandwich standard error, and keep the trial's target under any OBS confounding because the trial is randomized. They are the linear prognostic-score adjustment of \citet{schuler2022prognostic}; \citet{liao2025prognostic} and \citet{hojbjerre2026prognostic} put the same score into efficient and GLM plug-in estimators. All of them target the ATE; none of them estimates the CATE.

\paragraph{Results.} Table~\ref{tab:app-prog} and Fig.~\ref{fig:prog} compare them with fusion on the 45 settings of Sec.~\ref{sec:experiments}. On IHDP they match fusion, and Prog.\ score is slightly better at $n=300$ under a shift (0.45 against 0.47 and 0.48). On ACIC, fusion is at or below both adjustments in 14 of 15 settings. On the synthetic design, Prog.\ score exceeds the trial-only MSE for $n\ge600$ (up to 1.50), and Cov.~+~prog.\ score exceeds it at $n=3{,}000$. Over the 45 settings, fusion is at or below both adjustments in 35, and the paired 95\% bootstrap interval of the MSE ratio of fusion to the better adjustment is below 1 in 16 settings and above 1 in 2. Fusion never exceeds the trial-only MSE. The coverage of the 95\% Wald intervals is 0.94 to 0.96 for fusion and Prog.\ score and 0.91 to 0.97 for Cov.~+~prog.\ score. Under large effect heterogeneity (Table~\ref{tab:app-hethi}), Prog.\ score has 1.38 to 2.24 of the trial-only MSE and Cov.~+~prog.\ score 0.73 to 1.20, against 0.45 to 1.06 for fusion. For the CATE, the closest analogue is the trial loss on the sieve with $\hg$ (Table~\ref{tab:app-samesieve}), which uses the OBS prediction only as a basis function.

\begin{table}[tp]
\caption{ATE mean squared error of fusion and of the two prognostic-score adjustments relative to trial-only AIPW, on the designs and covariate laws of Table~\ref{tab:app-ate-ratio}; 1{,}000 replications with the same trial and OBS draws for every method. Fusion is shaded, and the smallest value in each column and trial size is in bold.}
\label{tab:app-prog}
\begin{center}\footnotesize
\begin{tabular}{@{}llccccccccc@{}}
\toprule
& & \multicolumn{3}{c}{Synthetic} & \multicolumn{3}{c}{IHDP} & \multicolumn{3}{c}{ACIC~2016}\\
\cmidrule(lr){3-5}\cmidrule(lr){6-8}\cmidrule(lr){9-11}
$n$ & Method & $r_0{=}1$ & weak & strong & $r_0{=}1$ & weak & strong & $r_0{=}1$ & weak & strong\\
\midrule
\rowcolor{gray!12}300 & Fusion & \textbf{0.55} & \textbf{0.60} & \textbf{0.60} & \textbf{0.46} & 0.47 & 0.48 & \textbf{0.45} & \textbf{0.46} & \textbf{0.47} \\
 & Prog.\ score & 1.00 & 1.00 & 1.00 & \textbf{0.46} & \textbf{0.45} & \textbf{0.45} & 0.48 & 0.48 & 0.49 \\
 & Cov.~+~prog.\ score & 0.63 & 0.63 & 0.63 & 0.48 & 0.48 & 0.48 & 0.61 & 0.61 & 0.61 \\
\addlinespace
\rowcolor{gray!12}900 & Fusion & \textbf{0.75} & \textbf{0.79} & \textbf{0.80} & \textbf{0.82} & 0.84 & 0.84 & \textbf{0.42} & \textbf{0.42} & \textbf{0.43} \\
 & Prog.\ score & 1.31 & 1.31 & 1.31 & \textbf{0.82} & \textbf{0.82} & \textbf{0.82} & 0.45 & 0.45 & 0.45 \\
 & Cov.~+~prog.\ score & 0.80 & 0.80 & \textbf{0.80} & 0.85 & 0.85 & 0.84 & 0.45 & 0.45 & 0.45 \\
\addlinespace
\rowcolor{gray!12}3{,}000 & Fusion & \textbf{0.96} & \textbf{1.00} & \textbf{1.00} & \textbf{0.92} & 0.98 & \textbf{0.96} & \textbf{0.77} & \textbf{0.79} & 0.79 \\
 & Prog.\ score & 1.50 & 1.50 & 1.50 & 0.96 & \textbf{0.96} & \textbf{0.96} & 0.83 & 0.83 & 0.82 \\
 & Cov.~+~prog.\ score & 1.03 & 1.03 & 1.02 & 0.96 & \textbf{0.96} & \textbf{0.96} & 0.78 & \textbf{0.79} & \textbf{0.78} \\
\bottomrule
\end{tabular}
\end{center}
\end{table}

\begin{figure}[tp]
\centering
\includegraphics[scale=0.82]{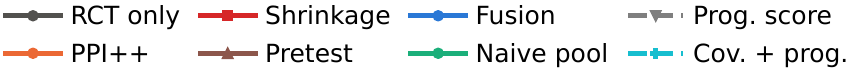}\\[2pt]
\includegraphics[width=\linewidth]{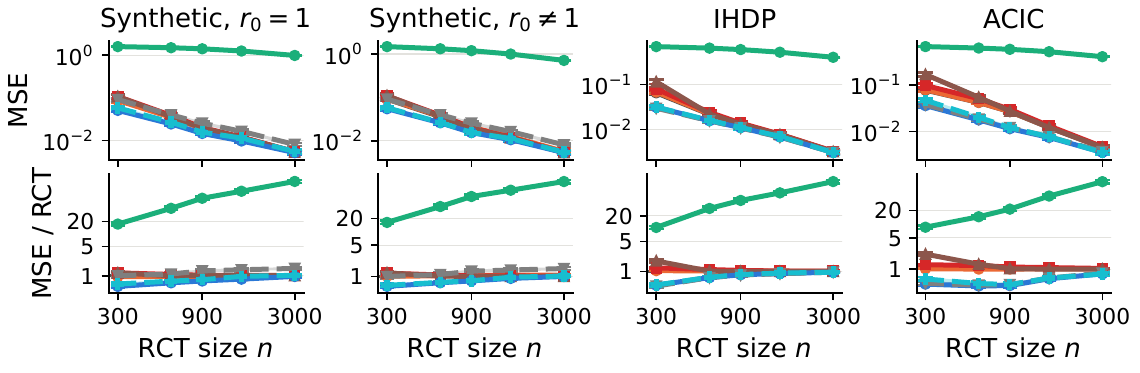}
\caption{Fig.~\ref{fig:results}, columns 1--4 (ATE), with the two prognostic-score adjustments added (dashed). Top: ATE MSE; bottom: its ratio to trial-only AIPW. 95\% bootstrap intervals.}
\label{fig:prog}
\end{figure}

\subsection{Inference and Absolute Errors}
\label{app:further}
This appendix gives the bias, standard deviation, and Wald intervals of the ATE estimates (Tables~\ref{tab:ate-inference} and~\ref{tab:ate-inference-b}) and the absolute errors behind Fig.~\ref{fig:results} (Fig.~\ref{fig:abs}).

\begin{table}[tp]
\caption{ATE over 1{,}000 replications on the synthetic design: bias, standard deviation, and root mean squared error, mean estimated standard error (SE), coverage, and mean width of the 95\% Wald interval. Shrinkage, Pretest, and the naive pool have no interval.}
\label{tab:ate-inference}
\begin{center}\footnotesize
\begin{tabular}{@{}llrrrrrrr@{}}
\toprule
Design & $n$ & Method & Bias & SD & RMSE & SE & Cov. & Width \\
\midrule
Synthetic, $r_0{=}1$ & 300 & RCT only & 0.007 & 0.302 & 0.302 & 0.298 & 0.95 & 1.169 \\
 &  & PPI++ & 0.007 & 0.294 & 0.294 & 0.294 & 0.95 & 1.152 \\
 &  & Fusion & 0.003 & 0.225 & 0.225 & 0.219 & 0.94 & 0.857 \\
 &  & Shrinkage & 0.077 & 0.318 & 0.327 & -- & -- & -- \\
 &  & Pretest & 0.014 & 0.326 & 0.326 & -- & -- & -- \\
 &  & Naive pool & 1.255 & 0.117 & 1.260 & -- & -- & -- \\
\addlinespace
 & 900 & RCT only & -0.007 & 0.139 & 0.139 & 0.143 & 0.95 & 0.562 \\
 &  & PPI++ & -0.009 & 0.137 & 0.137 & 0.140 & 0.95 & 0.550 \\
 &  & Fusion & -0.009 & 0.120 & 0.121 & 0.126 & 0.96 & 0.493 \\
 &  & Shrinkage & 0.011 & 0.141 & 0.141 & -- & -- & -- \\
 &  & Pretest & -0.007 & 0.139 & 0.139 & -- & -- & -- \\
 &  & Naive pool & 1.185 & 0.067 & 1.187 & -- & -- & -- \\
\addlinespace
 & 3000 & RCT only & -0.002 & 0.073 & 0.073 & 0.072 & 0.95 & 0.283 \\
 &  & PPI++ & -0.002 & 0.073 & 0.073 & 0.071 & 0.95 & 0.279 \\
 &  & Fusion & -0.002 & 0.072 & 0.072 & 0.070 & 0.95 & 0.273 \\
 &  & Shrinkage & 0.003 & 0.073 & 0.073 & -- & -- & -- \\
 &  & Pretest & -0.002 & 0.073 & 0.073 & -- & -- & -- \\
 &  & Naive pool & 0.991 & 0.039 & 0.992 & -- & -- & -- \\
\midrule
Synthetic, $r_0{\neq}1$ & 300 & RCT only & 0.007 & 0.302 & 0.302 & 0.298 & 0.95 & 1.169 \\
 &  & PPI++ & 0.002 & 0.304 & 0.304 & 0.303 & 0.95 & 1.186 \\
 &  & Fusion & 0.002 & 0.234 & 0.234 & 0.226 & 0.94 & 0.886 \\
 &  & Shrinkage & 0.080 & 0.319 & 0.328 & -- & -- & -- \\
 &  & Pretest & 0.016 & 0.331 & 0.331 & -- & -- & -- \\
 &  & Naive pool & 1.211 & 0.117 & 1.217 & -- & -- & -- \\
\addlinespace
 & 900 & RCT only & -0.007 & 0.139 & 0.139 & 0.143 & 0.95 & 0.562 \\
 &  & PPI++ & -0.007 & 0.143 & 0.143 & 0.146 & 0.96 & 0.574 \\
 &  & Fusion & -0.005 & 0.124 & 0.124 & 0.130 & 0.95 & 0.510 \\
 &  & Shrinkage & 0.012 & 0.141 & 0.142 & -- & -- & -- \\
 &  & Pretest & -0.007 & 0.139 & 0.139 & -- & -- & -- \\
 &  & Naive pool & 1.091 & 0.068 & 1.093 & -- & -- & -- \\
\addlinespace
 & 3000 & RCT only & -0.002 & 0.073 & 0.073 & 0.072 & 0.95 & 0.283 \\
 &  & PPI++ & -0.003 & 0.075 & 0.075 & 0.074 & 0.95 & 0.291 \\
 &  & Fusion & -0.002 & 0.073 & 0.073 & 0.071 & 0.95 & 0.280 \\
 &  & Shrinkage & 0.004 & 0.073 & 0.073 & -- & -- & -- \\
 &  & Pretest & -0.002 & 0.073 & 0.073 & -- & -- & -- \\
 &  & Naive pool & 0.843 & 0.043 & 0.844 & -- & -- & -- \\
\bottomrule
\end{tabular}

\end{center}
\end{table}

\begin{table}[tp]
\caption{As Table~\ref{tab:ate-inference}, for IHDP and ACIC~2016.}
\label{tab:ate-inference-b}
\begin{center}\footnotesize
\begin{tabular}{@{}llrrrrrrr@{}}
\toprule
Design & $n$ & Method & Bias & SD & RMSE & SE & Cov. & Width \\
\midrule
IHDP & 300 & RCT only & -0.007 & 0.255 & 0.255 & 0.250 & 0.96 & 0.981 \\
 &  & PPI++ & -0.008 & 0.261 & 0.261 & 0.253 & 0.96 & 0.991 \\
 &  & Fusion & -0.002 & 0.177 & 0.177 & 0.182 & 0.95 & 0.713 \\
 &  & Shrinkage & 0.066 & 0.274 & 0.281 & -- & -- & -- \\
 &  & Pretest & 0.034 & 0.337 & 0.339 & -- & -- & -- \\
 &  & Naive pool & 0.831 & 0.073 & 0.835 & -- & -- & -- \\
\addlinespace
 & 900 & RCT only & -0.003 & 0.114 & 0.114 & 0.116 & 0.96 & 0.456 \\
 &  & PPI++ & -0.003 & 0.116 & 0.116 & 0.118 & 0.96 & 0.463 \\
 &  & Fusion & -0.002 & 0.105 & 0.105 & 0.104 & 0.95 & 0.409 \\
 &  & Shrinkage & 0.015 & 0.117 & 0.118 & -- & -- & -- \\
 &  & Pretest & -0.003 & 0.114 & 0.114 & -- & -- & -- \\
 &  & Naive pool & 0.772 & 0.047 & 0.773 & -- & -- & -- \\
\addlinespace
 & 3000 & RCT only & 0.000 & 0.056 & 0.056 & 0.058 & 0.96 & 0.228 \\
 &  & PPI++ & 0.000 & 0.057 & 0.057 & 0.059 & 0.96 & 0.232 \\
 &  & Fusion & 0.000 & 0.055 & 0.055 & 0.057 & 0.96 & 0.224 \\
 &  & Shrinkage & 0.006 & 0.057 & 0.057 & -- & -- & -- \\
 &  & Pretest & 0.000 & 0.056 & 0.056 & -- & -- & -- \\
 &  & Naive pool & 0.633 & 0.038 & 0.635 & -- & -- & -- \\
\midrule
ACIC & 300 & RCT only & 0.003 & 0.278 & 0.278 & 0.287 & 0.96 & 1.124 \\
 &  & PPI++ & -0.001 & 0.281 & 0.281 & 0.289 & 0.95 & 1.132 \\
 &  & Fusion & -0.009 & 0.190 & 0.190 & 0.187 & 0.94 & 0.732 \\
 &  & Shrinkage & 0.095 & 0.297 & 0.312 & -- & -- & -- \\
 &  & Pretest & 0.081 & 0.401 & 0.408 & -- & -- & -- \\
 &  & Naive pool & 0.811 & 0.063 & 0.813 & -- & -- & -- \\
\addlinespace
 & 900 & RCT only & -0.005 & 0.163 & 0.163 & 0.160 & 0.94 & 0.629 \\
 &  & PPI++ & -0.006 & 0.164 & 0.164 & 0.161 & 0.94 & 0.632 \\
 &  & Fusion & -0.003 & 0.107 & 0.107 & 0.107 & 0.95 & 0.421 \\
 &  & Shrinkage & 0.028 & 0.170 & 0.173 & -- & -- & -- \\
 &  & Pretest & -0.005 & 0.165 & 0.165 & -- & -- & -- \\
 &  & Naive pool & 0.759 & 0.044 & 0.760 & -- & -- & -- \\
\addlinespace
 & 3000 & RCT only & 0.003 & 0.067 & 0.067 & 0.066 & 0.94 & 0.259 \\
 &  & PPI++ & 0.004 & 0.068 & 0.068 & 0.067 & 0.95 & 0.261 \\
 &  & Fusion & 0.003 & 0.060 & 0.060 & 0.059 & 0.95 & 0.230 \\
 &  & Shrinkage & 0.010 & 0.068 & 0.069 & -- & -- & -- \\
 &  & Pretest & 0.003 & 0.067 & 0.067 & -- & -- & -- \\
 &  & Naive pool & 0.633 & 0.031 & 0.634 & -- & -- & -- \\
\bottomrule
\end{tabular}

\end{center}
\end{table}

\begin{figure}[tp]
\centering
\includegraphics[scale=0.82]{s5_legend_ate}\\[2pt]
\includegraphics[width=\linewidth]{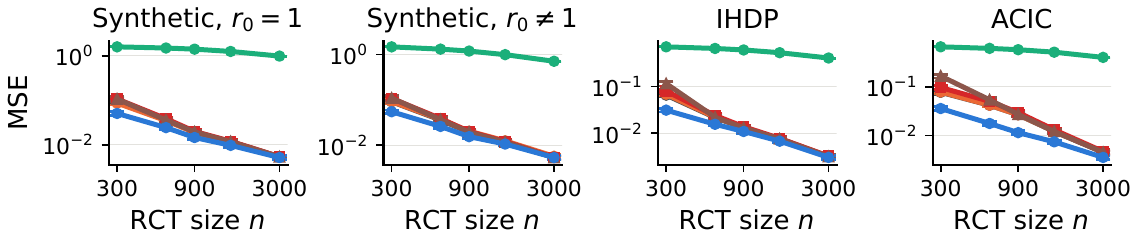}\\[4pt]
\includegraphics[scale=0.82]{s5_legend_cate_full}\\[2pt]
\includegraphics[width=\linewidth]{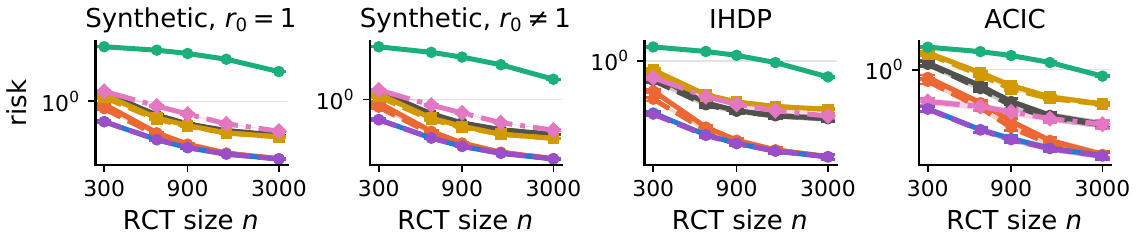}
\caption{Absolute errors for the designs of Fig.~\ref{fig:results}, columns 1--4. Top: ATE MSE; bottom: CATE risk. 95\% bootstrap intervals.}
\label{fig:abs}
\end{figure}

\endgroup

\end{document}